\documentclass{article}

\usepackage{iclr2025_conference_arxiv,times}

\usepackage{amsmath,amsfonts,bm}

\def\eqref#1{equation~\ref{#1}}

\def\1{\bm{1}}

\DeclareMathAlphabet{\mathsfit}{\encodingdefault}{\sfdefault}{m}{sl}
\SetMathAlphabet{\mathsfit}{bold}{\encodingdefault}{\sfdefault}{bx}{n}

\usepackage{hyperref}
\usepackage{url}

\usepackage{cleveref}
\usepackage{booktabs}
\usepackage{amsmath}
\usepackage{graphicx}
\usepackage{tikz}
\usepackage{makecell}
\usepackage{enumitem}
\usepackage{color, colortbl}
\usepackage{tikz}
\usepackage{subcaption}
\usepackage{wrapfig}
\usepackage{multirow}
\usepackage{listings}
\usepackage{algorithm}
\usepackage{algorithmic}
\usepackage{stfloats}
\usepackage{xspace}
\usepackage{booktabs}
\usepackage{xcolor}
\usepackage{afterpage}
\usepackage{setspace}
\usepackage{tocloft}
\usepackage{appendix}

\definecolor{tabhighlight}{HTML}{e5e5e5}
\definecolor{t2i}{HTML}{F0D695}
\definecolor{control}{HTML}{C9B1D3}
\definecolor{semantic}{HTML}{D26F70}
\definecolor{element}{HTML}{98B567}
\definecolor{repaint}{HTML}{82ACD1}

\crefname{section}{Sec.}{Secs.}
\Crefname{section}{Section}{Sections}
\crefname{table}{Tab.}{Tabs.}
\Crefname{table}{Table}{Tables}
\crefname{figure}{Fig.}{Figs.}
\Crefname{figure}{Figure}{Figures}
\crefname{equation}{Eq.}{Eqs.}
\Crefname{equation}{Equation}{Equations}
\crefname{algorithm}{Alg.}{Algs.}
\Crefname{algorithm}{Algorithm}{Algorithms}

\newcommand{\method}{\text{ACE}\xspace}

\title{
\raisebox{-0.05cm}{\includegraphics[scale=0.35]{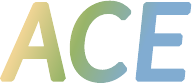}}: All-round Creator and Editor Following Instructions via Diffusion Transformer 
}

\author{%
\xspace \quad \xspace\xspace \quad  
Zhen Han\thanks{Equal Contribution. Order is determined by random dice rolling. \\ \hspace*{1em} $\dagger$ Project leader and corresponding author.} \quad
Zeyinzi Jiang$^*$ \quad
Yulin Pan$^*$ \quad
Jingfeng Zhang$^*$ \quad
Chaojie Mao$^{*\dagger}$ \quad 
\\[3pt]
\xspace \quad \xspace \quad \xspace \quad 
\xspace \quad \xspace \quad \xspace \quad
\xspace \quad \xspace\xspace\xspace\xspace \quad 
\textbf{Chenwei Xie} \quad
\textbf{Yu Liu} \quad
\textbf{Jingren Zhou} \quad 
\\[10pt]
\xspace \quad \xspace \quad \xspace \quad
\xspace \quad \xspace \quad \xspace \quad
\xspace \quad \xspace \quad \xspace \quad 
\xspace \quad \xspace \quad \xspace \quad 
\xspace\xspace \quad 
Tongyi Lab \quad 
}

\iclrfinalcopy 

\begin{document}

\maketitle

\begin{abstract}
%
Diffusion models have emerged as a powerful generative technology and have been found to be applicable in various scenarios. 
Most existing foundational diffusion models are primarily designed for text-guided visual generation and do not support multi-modal conditions, which are essential for many visual editing tasks. 
This limitation prevents these foundational diffusion models from serving as a unified model in the field of visual generation, like GPT-4 in the natural language processing field.
In this work, we propose \textbf{ACE}, an \textbf{A}ll-round \textbf{C}reator and \textbf{E}ditor, which achieves comparable performance compared to those expert models in a wide range of visual generation tasks. 
To achieve this goal, we first introduce a unified condition format termed Long-context Condition Unit (LCU), and propose a novel Transformer-based diffusion model that uses LCU as input, aiming for joint training across various generation and editing tasks.
Furthermore, we propose an efficient data collection approach to address the issue of the absence of available training data.
It involves acquiring pairwise images with synthesis-based or clustering-based pipelines and supplying these pairs with accurate textual instructions by leveraging a fine-tuned multi-modal large language model.
To comprehensively evaluate the performance of our model, we establish a benchmark of manually annotated pairs data across a variety of visual generation tasks.
The extensive experimental results demonstrate the superiority of our model in visual generation fields.
Thanks to the all-in-one capabilities of our model, we can easily build a multi-modal chat system that responds to any interactive request for image creation using a single model to serve as the backend, avoiding the cumbersome pipeline typically employed in visual agents.
%
Code and models will be available on the project page: \url{https://ali-vilab.github.io/ace-page/}.
\end{abstract}

\clearpage
\tableofcontents 
\clearpage

\section{Introduction}\label{sec:intro}

\begin{figure*}[t]
    \scriptsize
    \centering
    \includegraphics[width=\linewidth]{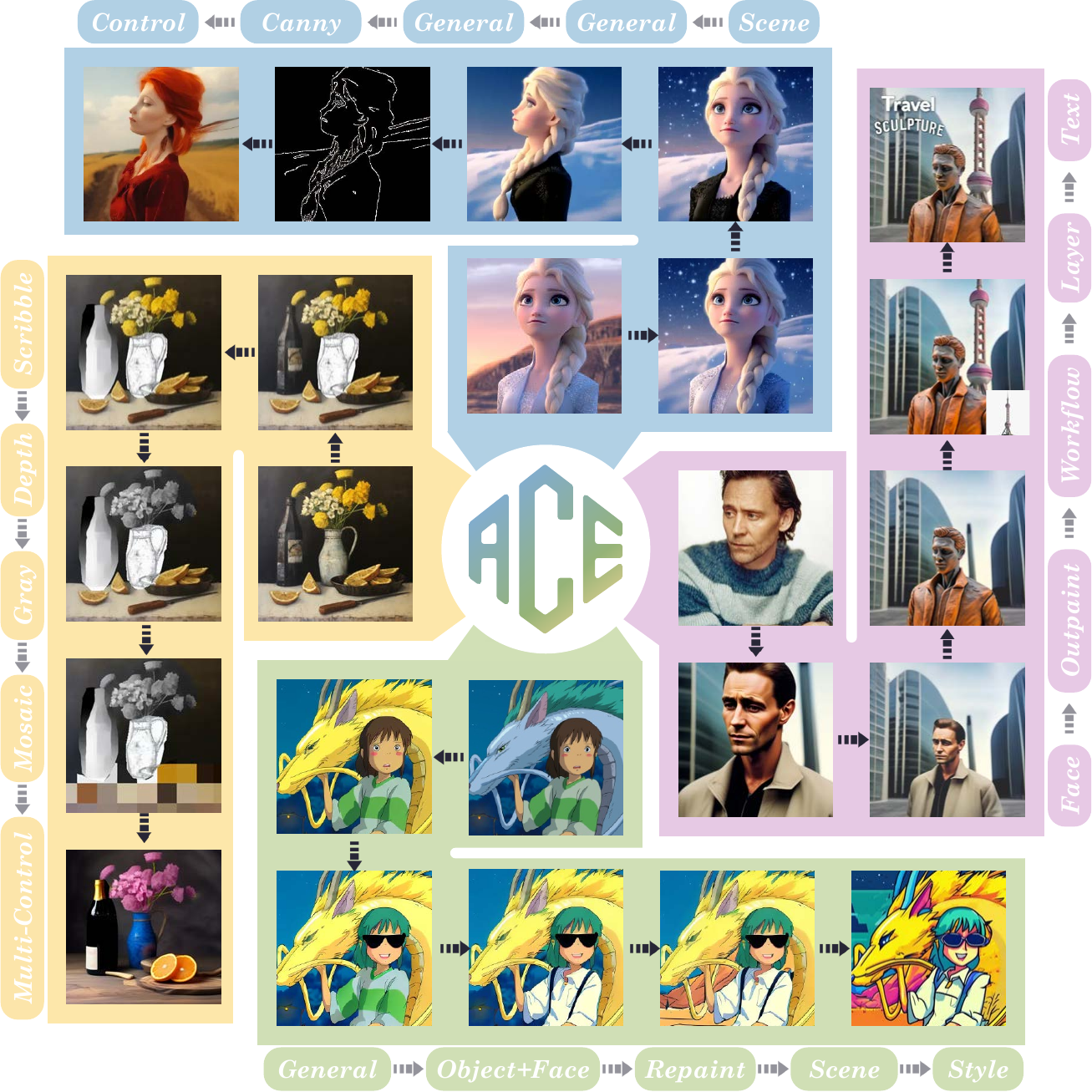}
    \caption{\textbf{Multi-turn image editing results of \method}. \method supports a wide range of image generation and editing tasks through natural language instructions, allowing complex and precise editing requests to be easily accomplished through multi-turn interactions.}
    \vspace{-15pt}
    \label{fig:teaser}
\end{figure*}

In recent years, foundational generative models have made groundbreaking progress in natural language processing (NLP)~\citep{palm2,claude,claude2,instructgpt}.
Conversational language models like ChatGPT~\citep{gpt3,gpt4} offer a unified framework for addressing various NLP tasks through a prompt-guided approach.
By employing a unified input-output structure, these models can achieve dynamic multi-turn interactions with users. Furthermore, by harnessing the knowledge of historical dialogues~\citep{claude3,gpt4o}, they possess the capacity to comprehend intricate queries with greater nuance and depth.
However, such unified architecture has not been fully explored in visual generation field. Existing foundational models of visual generation typically create images or videos from pure text, which is not compatible with most visual generation tasks, such as controllable image generation~\citep{controlnet, scedit} or image editing~\citep{ip2p}.
Thereby, specific visual generation tasks still require tailored tuning based on these foundational models, which is inflexible and inefficient.
For this reason, the visual generative model has not yet become a powerful and unified productivity tool in various application scenarios like large language models (LLMs)~\citep{phi3,llama,qwen,qwen2}. 

One major challenge of building an all-in-one visual generation model lies in the diversity of multi-modal input formats and the variety of supported generation tasks. 
To address this, we design a unified framework using a Diffusion Transformer generation model that accommodates a wide range of inputs and tasks, empowering it to serve as an \textbf{A}ll-round \textbf{C}reator and \textbf{E}ditor, which we refer to as \textbf{\method}.
First, we analyze the condition inputs of most visual generation tasks, and define Condition Unit (CU), which establishes a unified input paradigm consisting of core elements such as image, mask, and textual instruction. 
Second, for those CUs containing multiple images, we introduce Image Indicator Embedding to ensure the order of the images mentioned in instruction matches image sequence within the CUs. Besides, we imply 3d position embedding instead of 2d spatial-level position embedding on the image sequence, allowing for better exploring the relationships among conditional images.
Third, we concatenate the current CU with historical information from previous generation rounds to construct the Long-context Condition Unit (LCU). By leveraging this chain of generation information, we expect the model to better understand the user's request and create the desired image.
As depicted in ~\cref{fig:teaser}, \method supports a range of generating and editing capabilities, allowing it to accomplish complex and precise generation tasks through multi-turn instructions.

To address the issue of the absence of available training data for various visual generation tasks, we establish a meticulous data collection and processing workflow to collect high-quality structured CU data at a scale of 0.7 billion.
For visual conditions, we collect image pairs by synthesizing images from source images or by pairing images from large-scale databases. The former utilizes powerful open-source models to edit images to meet specific requirements, such as changing styles ~\citep{stylebooth} or adding objects ~\citep{largen}, while the latter involves clustering and grouping images from extensive databases to provide sufficient real data, thereby minimizing the risk of overfitting to the synthesized data distribution.
For textual instructions, we first manually construct instructions for diverse tasks by building templates or requesting LLMs, then optimize the instruction construction process by training an end-to-end instruction-labeling multi-modal large language model (MLLM)~\citep{internvl}, thereby enriching the diversity of the text instructions.

Our ACE provides more comprehensive coverage of tasks on a single model compared to previous approaches. Therefore, to thoroughly evaluate the performance of our generation model, we construct an evaluation benchmark that encompasses the main tasks. 
This benchmark incorporates inputs sourced from both the real world and model-generated data, supporting global and local editing tasks. It is larger in scale and broader in scope compared to previous benchmarks ~\citep{emuedit, magicbrush}.
We conduct a user study to subjectively assess the quality of images generated by our method and the adherence to instructions, revealing that our approach generally aligns more closely with human perception across the majority of tasks.
We summarize our main contributions as follows:

\begin{itemize}[left=0pt]
\item We propose \textbf{\method}, a unified foundational model framework that supports a wide range of visual generation tasks. To our knowledge, this is the most comprehensive diffusion generation model to date in terms of task coverage.
\item By defining the CU for unifying multi-modal inputs across different tasks and incorporating long-context CU, we introduce historical contextual information into visual generation tasks, paving the way for ChatGPT-like dialog systems in visual generation.
\item We design specific data construction pipelines for various tasks to enhance the quality and efficiency of data collection, and we ensure the richness of multi-modal data through MLLM fine-tuning for automated instruction labeling.
\item We establish a more comprehensive evaluation benchmark compared to previous ones, covering the most known visual generation tasks. Evaluation results indicate that ACE demonstrates notable competitiveness in specialized models while also exhibiting strong generalization capabilities across a broader range of open tasks.
\end{itemize}
\section{All-Round Creator and Editor}\label{sec:method}

\method is an image creation and editing model based on the Diffusion Transformer that follows textual instructions.
It establishes a unified framework that covers a wide range of tasks through the definition of standard input paradigm and strategy for aligning multi-modal information. 
With this exquisite design, the model is capable of handling various single tasks, multi-turn tasks, and long-context tasks with historical information.

\begin{figure*}[t]
    \scriptsize
    \centering
    \includegraphics[width=\linewidth]{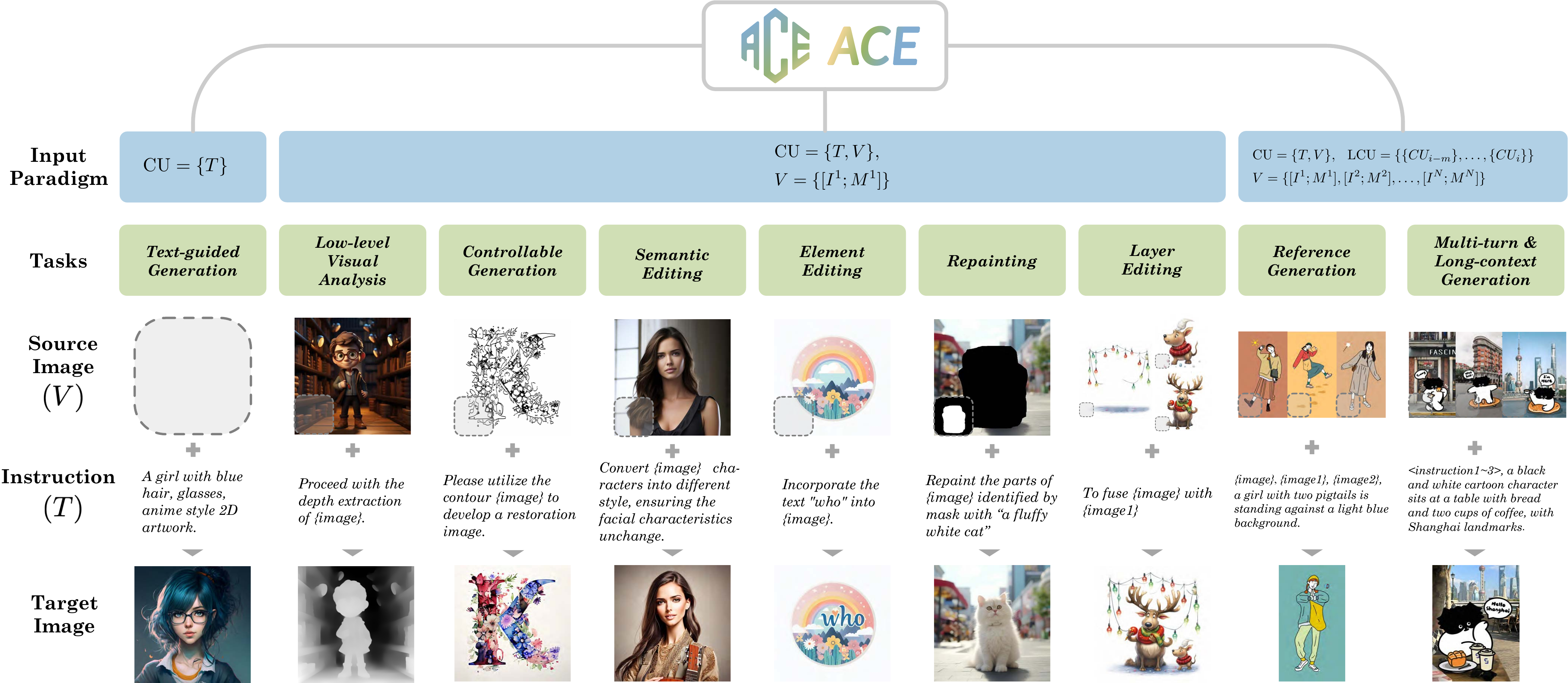}
    \caption{\textbf{The overview of all generation and editing task types supported by \method}. These tasks are categorized into 8 basic types, multi-turn and long-context generation based on different input conditions (in green) and are formulated using the proposed input paradigm as 3 formats (in blue).}
    \vspace{-10pt}
    \label{fig:tasks}
\end{figure*}

\subsection{Problem Definition}

\subsubsection{Tasks}

When it comes to generation and editing, the input condition information varies significantly depending on the specific task types. This encompasses a diverse range of forms, including textual instructions, conditioning images in controllable generation, masks used in region editing, and images in guided generation, among others.
We analyze and categorize these conditions from textual and visual modalities respectively: 
\textbf{(i) Textual modality}: we refer to all types of textual conditions as instructions and categorize them into \textbf{Generating-based Instructions} and \textbf{Editing-based Instructions}, depending on whether they describe the content of the generated image directly or the difference from the input visual cues;
\textbf{(ii) Visual modality}: we categorize all generation tasks into 8 basic types, as shown in ~\cref{fig:tasks}.
\begin{itemize}[leftmargin=10pt]
\setlength{\parsep}{0pt}
\setlength{\parskip}{0pt}
\item \textbf{Text-guided Generation}. It only uses generating-based text prompt as a condition to create images, and none of the visual cues are adopted.
\item \textbf{Low-level Visual Analysis}. It extracts low-level visual features from input images, such as edge maps or segmentation maps. One source image and editing-based instruction are required in the task to accomplish creation.
\item \textbf{Controllable Generation}. It is the inverse task of Low-level Visual Analysis, which creates vivid images based on given conditions, \textit{e.g.}, edge map, contour image, doodle image, scribble image, depth map, segmentation map, low-resolution image, \textit{etc}.
\item \textbf{Semantic Editing}. It aims to modify some semantic attributes of an input image by providing editing instructions, such as altering the style of an image or modifying the facial attributes of a character.
\item \textbf{Element Editing}. It focuses on adding, deleting, or replacing a specific subject in the image while keeping other elements unchanged.
\item \textbf{Repainting}. It erases and repaints partial image content of input image indicated by given mask and instruction.
\item \textbf{Layer Editing}. It decomposes an input image into different layers, each of which contains a subject or background, or reversely fuses different layers.
\item \textbf{Reference Generation}. It generates an image based on one or more reference images, analyzing the common elements among them and presenting these elements in the generated image.
\end{itemize}

By leveraging the generation tasks of these fundamental units, we can combine them to create \textbf{multi-turn scenarios}. Furthermore, utilizing the historical information from every round makes it possible to tackle \textbf{long-context visual generation} tasks.

\subsubsection{Input Paradigm}

A significant obstacle to implementing different types of generation and editing task requests within one framework lies in the diverse input condition formats of tasks. 
To address this issue, we design a unified input paradigm defined as \textbf{C}onditional \textbf{U}nit (\textbf{CU}) that fits as many tasks as possible.
The CUs composed of a textual instruction $T$ that describes the generation requirements, along with visual information $V$, where $V$ consists of a set of images $I$ that can be defined as $I=\emptyset$ (if there are no source image) or $I=\{I^{1}, I^{2}, \dots, I^{N}\}$ (if there are source images) and corresponding masks $M=\{M^{1}, M^{2}, \dots, M^{N}\}$. When there is no specific mask, $M$ is set to a blank image. The overall formulation of the CU is as follows:
\begin{equation}
    \text{CU} = \{T, V\}, \quad V=\{[I^{1}; M^{1}], [I^{2}; M^{2}], \dots, [I^{N}; M^{N}]\},
\end{equation}
where a channel-wise connection operation is performed between corresponding $I$ and $M$, $N$ represents the total number of visual information inputs for this task.

Furthermore, to better address the demands of complex long-context generation and editing, historical information can be optionally integrated into CU, which is formulated as: 
\begin{equation}
    \text{LCU}_{i} = \{\{T_{i-m}, T_{i-m+1}, \dots, T_{i}\}, \{V_{i-m}, V_{i-m+1}, \dots, V_{i}\}\}
\end{equation}
where $m$ denotes the maximum number of rounds of historical knowledge introduced in the current request. 
$\text{LCU}_{i}$ is a \textbf{L}ong-context \textbf{C}ondition \textbf{U}nit used to generate desired content for the $i$-th request.

\begin{figure*}[t]
    \scriptsize
    \centering
    \includegraphics[width=\linewidth]{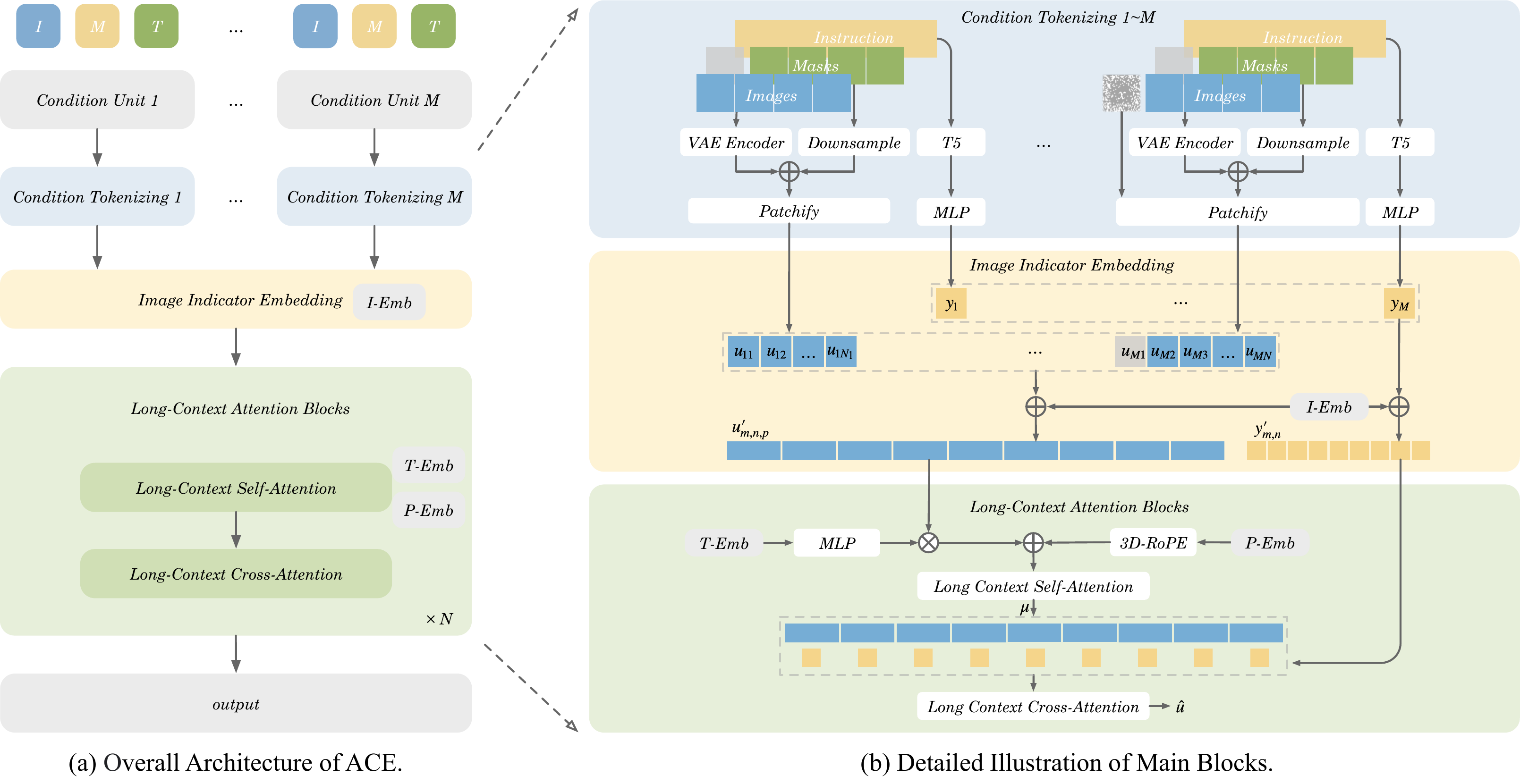}
    \caption{\textbf{The illustration of \method framework.} Condition Tokenizing module tokenizes each input CU, concatenating them to obtain the visual token sequence and the text token sequence. The Image Indicator Embedding module employs pre-defined textual tokens to indicate the image order in textual instructions and distinguish various input images. The Long-context Attention Block ensures effective communication and integration of long-context sequences.}
    \vspace{-10pt}
    \label{fig:method}
\end{figure*}


\subsection{Architecture}

In this section, we introduce a unified visual generation framework that can perform all visual generation tasks within a single model, and incorporate long-context conditions to enhance comprehension.
As illustrated in \cref{fig:method}a, the overall framework is built based on a Diffusion Transformer model~\citep{transformer,dit}, and integrated with three novel components to achieve unified generation: Condition Tokenizing, Image Indicator Embedding, and Long-context Attention Block.
We will provide a detailed description of them below.

\noindent\textbf{Condition Tokenizing}.
Considering an LCU that comprises $M$ CUs, the model involves three entry points for each CU: a language model (T5)~\citep{t5} to encode textual instructions, a Variational Autoencoder (VAE)~\citep{vae} to compress reference image to latent representation, and a down-sampling module to resize mask to the shape of corresponding latent image. 
The latent image and its mask (an all-one mask if no mask is provided) are concatenated along the channel dimension. These image-mask pairs are then patchified into 1-dimensional visual token sequences $u_{m,n,p}$, where $m$, $n$ are indexes for CUs and visual information Vs in each CU, while $p$ denotes the spatial index in patchified latent images. Similarly, textual instructions are encoded into 1-dimensional token sequences $y_{m}$. 
After processing within each CU, we separately concatenate all visual token sequences and all textual token sequences to form a long-context sequence.

\noindent\textbf{Image Indicator Embedding}.
As illustrated as \cref{fig:method}b, to indicate the image order in textual instructions and distinguish various input images, we encode some pre-defined textual tokens ``\{image\}, \{image1\}, ..., \{imageN\}" into T5 embeddings as Image Indicator Embeddings ($I$-Emb). These indicator embeddings are added to the corresponding image embedding sequence and text embedding sequence, which is formulated as:
\begin{equation}
y'_{m,n}=y_m + I\text{-Emb}_{m,n},
\end{equation}
\begin{equation}
u'_{m,n,p}=u_{m,n,p} + I\text{-Emb}_{m,n}.
\end{equation}
In this way, image indicator tokens in textual instructions and the corresponding images are implicitly associated.

\noindent\textbf{Long-context Attention Block}.
Given the long-context visual sequence, we first modulate it with the time step embedding ($T$-Emb), then incorporate a 3D Rotational Positional Encodings (RoPE)~\citep{rope} to differentiate between different spatial- and frame-level image embeddings.
During the Long Context Self-Attention, all image embeddings of each CU at each spatial location, are equivalently and comprehensively interact with each other by $\mu=Attn(u',u')$.
Next, unlike the cross-attention layer of the conventional Diffusion Transformer model, where each visual token attends to all of the textual tokens, we implement cross-attention operation with each condition unit. That means image tokens in $m$-th CU will only attend to the textual tokens from the same CU. This can be formulated as:
\begin{equation}
\hat{u}_{m,n}=Attn(\mu_{m,n},y'_{m,n}).
\end{equation}
This ensures that, within the cross-attention layer, the text embeddings and image embeddings align on a frame-by-frame basis.
\section{Datasets}\label{sec:datas}

\begin{figure*}[t]
    \scriptsize
    \centering
    \includegraphics[width=\linewidth]{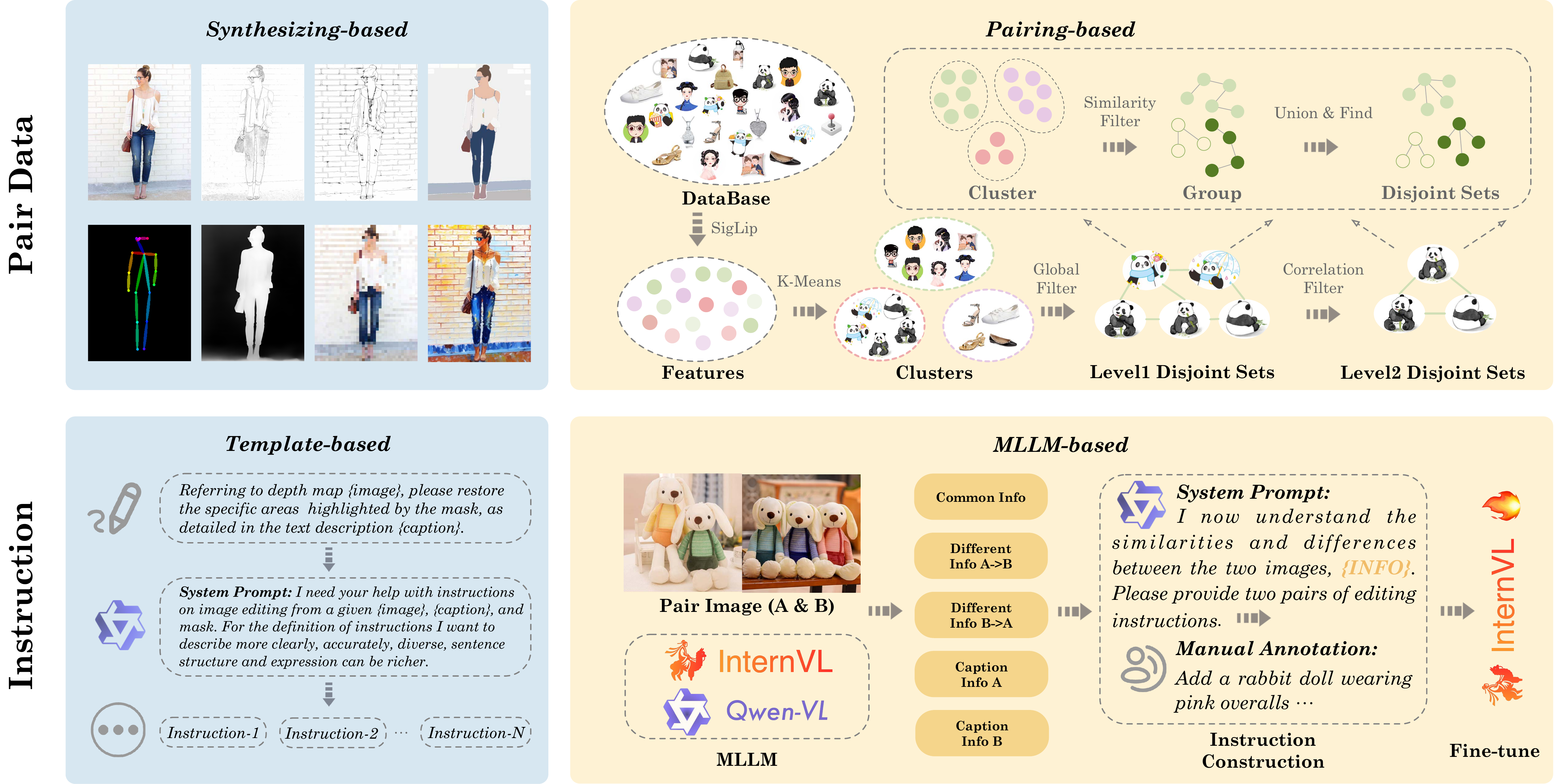}
    \caption{\textbf{The pipeline of dataset construction and instructions labeling.} In data construction, two methods are utilized: synthesizing using open-source expert models and mining from large-scale data. For instruction labeling, we combined templating with MLLM labeling, further training the Instruction Captioner to achieve large-scale instruction labeling.}
    \vspace{-5pt}
    \label{fig:datasets}
\end{figure*}

\subsection{Pair Data Collection} \label{sec:data_pair}
 
A critical challenge of training foundational visual generation model lies in how to acquire pairwise images for various tasks.
In this section, we introduce two ways to efficiently build high-quality datasets for most of the generation and editing tasks:
\textbf{(i) Synthesizing from source image}: 
thanks to the rapid development in the field of visual generation, there have been many of powerful open-source models designed to solve one specific problem. Leveraging these powerful single-point technologies, we could synthesis plenty of image pairs for lots of generation and editing tasks, such as controllable generation, style editing, object editing, and so on.
\textbf{(ii) Pairing from massive databases}: 
though the synthesis-based method is efficient and straightforward in acquiring pairwise data. However, It still possesses two drawbacks. 
First, some editing problems have not been fully explored, and there are no powerful open-source models available for these tasks. Second, using only synthetic data can easily cause over-fitting and reduce the quality of generated images. Therefore, it is essential to provide sufficient real data to address the aforementioned drawbacks.
We propose a hierarchically aggregating pipeline for pairing content-related images from massive databases to build pairs of data for training, as illustrated in ~\cref{fig:datasets}.
We first extract semantic features using SigLIP~\citep{siglip} from large-scale datasets (\textit{e.g.}, LAION-5B~\citep{laion5b}, OpenImages~\citep{openimage}, and our private datasets). 
Then leveraging K-means clustering technology, coarse-grained clustering is implemented to divide all images into tens of thousands of clusters.
Within each cluster, we implement a two-turn union-find algorithm to achieve fine-grained image aggregation. The first turn is based on the SigLIP feature and the second turn uses a similarity score tailored for specific tasks. For instance, we calculate the face similarity score for the facial editing task and the object consistency score for the general editing task.
Finally, we collect all possible pairs from each disjoint set and implement cleaning strategies to filter high-quality pairs. 
Benefiting from these two automatic pipelines, we construct a large-scale training dataset that consists of nearly 0.7 billion image pairs, covering 8 basic types of tasks, multi-turn and long-context generation. We depict its distribution in \cref{fig:train_dist} and provide a detailed description of the specific data construction methods for each task, please refer to ~\cref{sec:datasets_detail}.

\subsection{Instructions}\label{sec:data_instr}

In addition to collecting image pairs, it is essential to label clear natural language instructions that indicate how to transform one image into another. 
Compared to the caption generation commonly used in text-to-image task, instruction labeling is generally more challenging, as it requires analyzing not only the semantics of individual images, but also the discrepancies across multiple images.
We employ both \textbf{Template-based} and \textbf{MLLM-based} methods to tackle this challenge.
Template-based method constructs instruction templates for specific vision tasks by leveraging human knowledge priors.
However, the instructions generated by this method lack diversity, which can lead to significant overfitting problems.
MLLM-based method generates unique instructions for each given editing pair, leveraging off-the-shelf MLLMs. 
Nonetheless, current MLLMs exhibit limitations in producing precise instructions for editing tasks involving non-natural images, such as depth-controlled image generation and image segmentation. 
Thus, we combine these two methods and design an effective strategy to mitigate the aforementioned drawbacks.
For tasks that contain non-natural images, we utilize a template-based method to generate instruction templates. These templates are then combined with the generated captions to produce the final instructions. 
To address the issue of insufficient diversity, we employ LLMs to reformulate instructions multiple times, and tune prompts to ensure that each rewritten version is distinct from all preceding instructions.
For tasks that contain natural images, we employ an MLLM to predict the differences and commonalities between the images in the input pair. 
Then an LLM is used to generate instructions focusing on semantic distinctions according to the analysis of the differences and commonalities.
Further, the collected instructions generated by these two methods undergo human annotation and correction. 
The revised instructions are used for fine-tuning an open-source MLLM, enabling it to predict instructions for any given image pair.
Specifically, we collect a dataset of approximately 800,000 curated instructions and train an \textbf{Instruction Captioner} by fine-tuning the InternVL2-26B~\citep{internvl}.
Once trained, the Instruction Captioner is able to take any two images as input and generates the instruction for transforming the source image to the target image.
It can also be further extended to the processing of cluster data, by entering a set of images, obtaining the similarity description among images within the cluster, and the differences between each pair within the cluster. The above process is illustrated in ~\cref{fig:datasets}.
\section{Experiments}\label{sec:exp}

\begin{figure*}[!ht]
    \scriptsize
    \centering
    \includegraphics[width=1.0\linewidth]{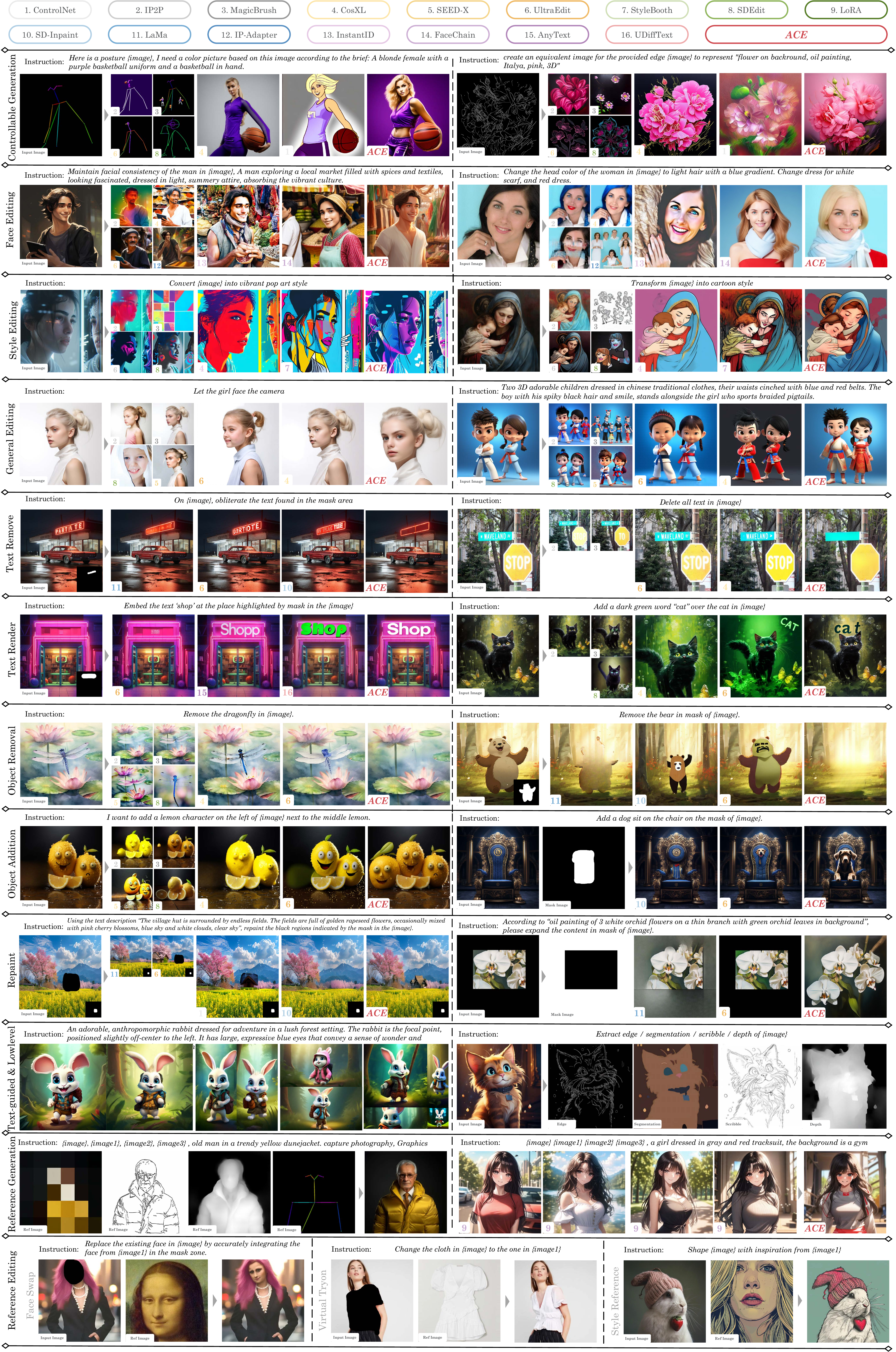}
    \caption{\textbf{Comparison and visualization of \method performance with expert models} in different tasks. \method demonstrates adaptability to multi-task and achieves superior performance.}
    \label{fig:qual}
    \vspace{-10pt}
\end{figure*}


\begin{table}[t]
\scriptsize
\centering
\setlength{\tabcolsep}{9pt}
\caption{\textbf{Results on the MagicBrush benchmark}. LC denotes long-context generation with history.}
\label{tab:assess_mb}
{
\begin{tabular}{clccccc}
\toprule
\textbf{Settings} & \textbf{Methods} & L1$\downarrow$ & L2$\downarrow$ & CLIP-I$\uparrow$ & DINO$\uparrow$ & CLIP-T$\uparrow$ \\ 
\midrule
\multirow{13}{*}{\rotatebox{90}{\textbf{Single-turn}}} & \multicolumn{6}{c}{\textit{\textbf{Global Description-guided}}} \\ \cmidrule{2-7} 
              & SD-SDEdit~\citep{sdedit}        & 0.1014 & 0.0278 & 0.8526 & 0.7726 & \underline{0.2777} \\
              & Null Text Inversion~\citep{nulltextinversion} & \underline{0.0749} & \underline{0.0197} & 0.8827 & 0.8206 & 0.2737\\  
              & GLIDE~\citep{glide}            & 3.4973 & 115.8347 & \bf 0.9487 & \bf 0.9206 & 0.2249\\
              & Blended Diffusion~\citep{blenddiff} & 3.5631 & 119.2813 & 0.9291 & 0.8644 & 0.2622 \\
              & \cellcolor{tabhighlight}\textbf{\method} (Ours)  & \cellcolor{tabhighlight} \bf 0.0505 & \cellcolor{tabhighlight} \bf 0.0160 & \cellcolor{tabhighlight} \underline{0.9436} & \cellcolor{tabhighlight} \underline{0.9184} & \cellcolor{tabhighlight} \bf 0.2833\\ 
              \cmidrule{2-7}
              & \multicolumn{6}{c}{\textit{\textbf{Instruction-guided}}} \\ \cmidrule{2-7} 
              & HIVE~\citep{hive}             & 0.1092 & 0.0380 & 0.8519 & 0.7500 & - \\
              & InstructPix2Pix~\citep{ip2p}  & 0.1122 & 0.0371 & 0.8524 & 0.7428 & 0.2764\\
              & MagicBrush~\citep{magicbrush}       & 0.0625 & 0.0203 & \underline{0.9332} & \underline{0.8987} & \underline{0.2781}\\
              & UltraEdit~\citep{ultraedit}     & \underline{0.0575} & \underline{0.0172} & 0.9307 & 0.8982 & -\\ 
              \rowcolor{tabhighlight} \cellcolor{white} & \textbf{\method} (Ours)     & \bf 0.0507 & \bf 0.0165 & \bf 0.9453 & \bf 0.9215 & \bf 0.2841\\ 
\midrule
\multirow{18}{*}{\rotatebox{90}{\textbf{Multi-turn}}} & \multicolumn{6}{c}{\textit{\textbf{Global Description-guided}}} \\ \cmidrule{2-7}
              & SD-SDEdit~\citep{sdedit}        & 0.1616 & 0.0602 & 0.7933 & 0.6212 & 0.2694 \\
              & Null Text Inversion~\citep{nulltextinversion} & 0.1057 & 0.0335 & 0.8468 & 0.7529 & 0.2710 \\
              & GLIDE~\citep{glide}             & 11.7487 & 1079.5997 & 0.9094 & 0.8494 & 0.2252 \\
              & Blended Diffusion~\citep{blenddiff} & 14.5439 & 1510.2271 & 0.8782 & 0.7690 & 0.2619 \\
              \rowcolor{tabhighlight} \cellcolor{white} & \textbf{\method} (Ours)  & \underline{0.0778} & \underline{0.0290} & \underline{0.9124} & \underline{0.8611} & \bf 0.2843\\
              &\cellcolor{tabhighlight}\textbf{\method} (Ours w/ LC) & \cellcolor{tabhighlight} \bf 0.0768 & \cellcolor{tabhighlight} \bf 0.0285 & \cellcolor{tabhighlight} \bf 0.9136 & \cellcolor{tabhighlight} \bf 0.8635 & \cellcolor{tabhighlight} \underline{0.2819}\\ 
              \cmidrule{2-7}
              & \multicolumn{6}{c}{\textit{\textbf{Instruction-guided}}} \\ \cmidrule{2-7} 
              & HIVE~\citep{hive}                   & 0.1521 & 0.0557 & 0.8004 & 0.6463 & 0.2673 \\
              & InstructPix2Pix~\citep{ip2p}  & 0.1584 & 0.0598 & 0.7924 & 0.6177 & 0.2726 \\
              & MagicBrush~\citep{magicbrush}             & 0.0964 & 0.0353 & 0.8924 & 0.8273 & 0.2754 \\
              &  UltraEdit~\citep{ultraedit}         & \bf 0.0745 & \bf 0.0236 & 0.9045 & 0.8505 &  - \\
              \rowcolor{tabhighlight} \cellcolor{white} & \textbf{\method} (Ours)  & 0.0773 & 0.0293 & \underline{0.9128} & \underline{0.8661} & \bf 0.2855\\
              \rowcolor{tabhighlight} \cellcolor{white} & \textbf{\method} (Ours w/ LC)    & \underline{0.0761} & \underline{0.0284} & \bf 0.9140 & \bf 0.8668 & \underline{0.2809}\\
\bottomrule
\end{tabular}
}
\vspace{-10pt}
\end{table}


\subsection{Benchmarks and Metrics}
\noindent\textbf{Existing Benchmarks}. 
We first evaluate on the commonly used benchmark MagicBrush~\citep{magicbrush}.
It contains an overall 1,053 edit turns and 535 edit sessions for single-turn and multi-turn image editing respectively. 
It compares the output images with groundtruth images and the provided target text descriptions. 
Following the setting proposed in the MagicBrush benchmark, we calculate the L1 distance, L2 distance, CLIP~\citep{clip} similarity, DINO~\citep{groundingdino} similarity between the generated image and groundtruth image, and CLIP similarity between the generated image and textual prompt.
We also evaluate the Emu Edit benchmark~\citep{emuedit}, please see ~\cref{sec:exp_more} for details.

\noindent\textbf{ACE Benchmark}.
To thoroughly evaluate the performance of various visual generation tasks, we build a benchmark dataset that covers all types of tasks the aforementioned. ACE benchmark consists of both real and generated images. The real images are primarily sourced from the MS-COCO~\citep{coco} dataset and the generated images are created by Midjourney~\citep{midjourney}, using prompts obtained from JourneyDB~\citep{journeydb}. 
For each task type, we manually craft instructions and masks to closely resemble actual user input patterns, reaching a total of 12,000 entries. 
The detailed statistics of \method benchmark can be found in ~\cref{fig:benchmark_dist}.
We evaluate image quality and prompt following scores through a user study. 
The image quality score assesses the aesthetic quality of the generated images, while the prompt following score measures how well the images align with the provided textual instructions.

\subsection{Qualitative Evaluation}
In our qualitative evaluation, we present a comparison of our method with SOTA approaches across various tasks, including ControlNet~\citep{controlnet}, InstructPix2Pix~\citep{ip2p}, MagicBrush~\citep{magicbrush}, CosXL~\citep{cosxl}, SEED-X Edit~\citep{seededit}, UltraEdit~\citep{ultraedit}, StyleBooth~\citep{stylebooth}, SDEdit~\citep{sdedit}, LoRA~\citep{lora}, SD-Inpaint~\citep{sdinp}, LaMa~\citep{lama}, IP-Adapter~\citep{ipadapter}, InstantID~\citep{instantid}, FaceChain~\citep{facechain}, AnyText~\citep{anytext}, UDiffText~\citep{udifftext}.
In ~\cref{fig:qual}, we present qualitative comparisons between our single \method model and 16 other methods across 12 subtasks. Overall, our method not only addresses a diverse range of tasks but also performs superior compared to task-specific methods.
Additionally, we also show some extra tasks that the comparison methods do not perform well in the last three lines.
Please see ~\cref{sec:vis_more}, for more examples of qualitative evaluation.

\begin{table}[t]
\caption{
\textbf{User study results on \method benchmark}. For each method in every supported task, we evaluate both prompt following and image quality, reporting the two scores in a single cell, separated by a ``/''.
``-'' means this task does not exist or is not supported by the current method.
}
\centering
\scriptsize
\setlength{\tabcolsep}{0.9pt}
\begin{tabular}{l|c|cccc|ccc|cccc|cc}

\toprule
& \multicolumn{1}{c|}{\textbf{Txt2img}} 
& \multicolumn{4}{c|}{\textbf{Controllable}} 
& \multicolumn{3}{c|}{\textbf{Semantic}} 
& \multicolumn{4}{c|}{\textbf{Element}} 
& \multicolumn{2}{c}{\textbf{Repainting}} \\
& \rotatebox{90}{\raisebox{0.5pt}{\tikz\fill[t2i] (0,0) circle (.5ex);} Txt2img}
 & \rotatebox{90}{\raisebox{0.5pt}{\tikz\fill[control] (0,0) circle (.5ex);} Canny}
 & \rotatebox{90}{\raisebox{0.5pt}{\tikz\fill[control] (0,0) circle (.5ex);} Depth}
 & \rotatebox{90}{\raisebox{0.5pt}{\tikz\fill[control] (0,0) circle (.5ex);} Scribble}
 & \rotatebox{90}{\raisebox{0.5pt}{\tikz\fill[control] (0,0) circle (.5ex);} Pose}
 & \rotatebox{90}{\raisebox{0.5pt}{\tikz\fill[semantic] (0,0) circle (.5ex);} Face}
 & \rotatebox{90}{\raisebox{0.5pt}{\tikz\fill[semantic] (0,0) circle (.5ex);} Style}
 & \rotatebox{90}{\raisebox{0.5pt}{\tikz\fill[semantic] (0,0) circle (.5ex);} General}
 & \rotatebox{90}{\raisebox{0.5pt}{\tikz\fill[element] (0,0) circle (.5ex);} Add Text}
 & \rotatebox{90}{\raisebox{0.5pt}{\tikz\fill[element] (0,0) circle (.5ex);} Rm Text}
 & \rotatebox{90}{\raisebox{0.5pt}{\tikz\fill[element] (0,0) circle (.5ex);} Add Obj.}
 & \rotatebox{90}{\raisebox{0.5pt}{\tikz\fill[element] (0,0) circle (.5ex);} Rm Obj.}
 & \rotatebox{90}{\raisebox{0.5pt}{\tikz\fill[repaint] (0,0) circle (.5ex);} Inpaint}
 & \rotatebox{90}{\raisebox{0.5pt}{\tikz\fill[repaint] (0,0) circle (.5ex);} Outpaint} \\
\midrule
\multicolumn{15}{c}{\textbf{\emph{Global Editing}}} \\

\midrule 
SD1.5~\citep{sd15} & 3.3/2.2 & - & - & - & - & - & - & - & - & - & - & - & - & -  \\ 
SDXL~\citep{sdxl} & \textbf{4.1}/\textbf{2.8} & - & - & - & - & - & - & - & - & - & - & - & - & -  \\ 
CtrlNet~\citep{controlnet} & - & 2.5/2.0 & 3.8/2.4 & 1.9/2.0 & 2.9/1.9 & - & - & - & - & - & - & - & - & -  \\ 
StyleBooth~\citep{stylebooth} & - & - & - & - & - & - & \textbf{3.3}/\underline{2.6} & - & - & - & - & - & - & -  \\ 
IP-Adapter~\citep{ipadapter} & - & - & - & - & - & 2.0/2.2 & - & 1.7/2.5 & - & - & - & - & - & -  \\ 
InstantID~\citep{instantid} & - & - & - & - & - & 2.5/2.7 & - & - & - & - & - & - & - & -  \\ 
FaceChain~\citep{facechain} & - & - & - & - & - & 2.0/\underline{3.0} & - & - & - & - & - & - & - & -  \\ 
SDEdit~\citep{sdedit} & - & 1.4/1.9 & 1.3/1.8 & 1.1/1.6 & 1.2/1.4 & 1.3/2.1 & 1.1/1.7 & 1.5/2.1 & 1.1/2.2 & 1.1/1.7 & 1.5/2.1 & 1.1/2.0 & - & -  \\ 
IP2P~\citep{ip2p} & - & 1.9/2.0 & 1.7/2.0 & 1.5/2.3 & 1.4/1.4 & 2.3/2.4 & 2.4/2.5 & 2.2/2.4 & 1.1/2.6 & 1.3/2.6 & 2.0/2.4 & 1.5/2.4 & - & -  \\ 
MB~\citep{magicbrush} & - & 1.3/1.8 & 1.3/1.7 & 1.3/1.9 & 1.1/1.3 & 2.4/2.3 & 1.4/2.0 & 2.2/2.3 & 1.5/2.4 & \underline{2.2}/2.5 & \textbf{3.1}/2.2 & 2.1/2.4 & - & -  \\ 
SEED-X~\citep{seedx} & - & 1.6/2.1 & 1.7/2.0 & 1.7/2.2 & 1.5/1.5 & 2.0/2.7 & 2.2/2.5 & 2.1/\underline{2.7} & 1.3/2.6 & 2.1/2.6 & 1.9/\textbf{2.6} & \underline{2.5}/2.4 & - & -  \\ 
CosXL~\citep{cosxl} & - & \underline{4.1}/\textbf{2.9} & \underline{4.1}/\textbf{2.8} & \underline{2.6}/\textbf{2.9} & \underline{3.7}/\underline{2.1} & \textbf{2.9}/\textbf{3.1} & \underline{3.2}/\textbf{3.0} & \textbf{3.2}/\textbf{2.9} & 1.4/\textbf{2.7} & 1.0/\textbf{2.9} & \underline{2.8}/\underline{2.5} & 1.1/\textbf{3.1} & - & -  \\ 
UltraEdit~\citep{ultraedit} & - & 1.7/2.2 & 1.2/1.8 & 1.3/2.3 & 1.1/1.3 & 2.3/2.5 & 2.1/2.4 & \underline{2.6}/2.5 & \underline{1.7}/2.6 & 1.1/2.7 & 2.7/2.3 & 1.5/\underline{2.6} & - & -  \\ 
\rowcolor{tabhighlight} \textbf{\method} (Ours) & \underline{3.7}/\underline{2.5} & \textbf{4.6}/\underline{2.7} & \textbf{4.5}/\textbf{2.8} & \textbf{4.8}/\textbf{2.9} & \textbf{4.1}/\textbf{2.3} & \underline{2.8}/2.8 & 2.4/\underline{2.6} & 2.1/2.5 & \textbf{2.8}/\textbf{2.7} & \textbf{4.4}/\textbf{2.9} & 2.6/2.4 & \textbf{3.9}/2.5 & - & - \\ 

\midrule
\multicolumn{15}{c}{\textbf{\emph{Local Editing}}} \\
\midrule
LaMa~\citep{lama} & - & - & - & - & - & - & - & - & - & \underline{3.6}/2.8 & - & \textbf{4.5}/\textbf{2.8} & 1.6/2.3 & 3.0/\underline{2.4}  \\ 
SDInpaint~\citep{sdinp} & - & - & - & - & - & - & - & - & - & 2.6/2.6 & 1.6/\textbf{2.7} & 2.2/\underline{2.5} & \underline{3.6}/\underline{2.6} & -  \\ 
CtrlNet~\citep{controlnet} & - & - & - & - & - & - & - & - & - & 2.9/2.7 & 1.9/\underline{2.5} & 2.6/2.2 & 3.0/2.1 & \underline{3.2}/2.1  \\ 
AnyText~\citep{anytext} & - & - & - & - & - & - & - & - & 3.5/2.7 & - & - & - & - & -  \\ 
UDiffText~\citep{udifftext} & - & - & - & - & - & - & - & - & \underline{3.6}/2.7 & - & - & - & - & -  \\ 
UltraEdit~\citep{ultraedit} & - & \underline{1.4}/\underline{1.9} & \underline{1.2}/\underline{1.8} & \underline{1.2}/\underline{2.0} & - & - & - & - & 1.1/\underline{2.8} & 1.2/\textbf{2.9} & \underline{2.9}/\underline{2.5} & 1.4/\underline{2.5} & 1.1/1.7 & 1.1/2.1  \\ 
\rowcolor{tabhighlight} \textbf{\method} (Ours) & - & \textbf{4.8}/\textbf{2.6} & \textbf{4.3}/\textbf{2.5} & \textbf{4.8}/\textbf{2.6} & - & - & - & - & \textbf{4.5}/\textbf{2.9} & \textbf{4.5}/\textbf{2.9} & \textbf{3.7}/\underline{2.5} & \underline{4.3}/\underline{2.5} & \textbf{4.4}/\textbf{2.7} & \textbf{4.6}/\textbf{2.8} \\ 
\bottomrule
\end{tabular}
\label{tab:user_study}
\vspace{-10pt}
\end{table}


\subsection{Quantitative Evaluation}
\noindent\textbf{Evaluation on Existing Benchmarks}.
We first compare our method with baselines on the MagicBrush benchmark. Results are present on \cref{tab:assess_mb}. 
For single-turn image editing, \method significantly outperforms other methods under an instruction-guided setting while demonstrating comparable performance under a description-guided setting.
For each setting of multi-turn image editing, we first employ the same inference way as MagicBrush, performing independent and continuous edits on a single image. 
The results show that our approach has significant advantages.
Furthermore, we construct a long sequence using the historical information from each editing round, achieving a certain improvement in performance compared to not using it. This also demonstrates the effectiveness of LCU and architecture design.

\noindent\textbf{Evaluation on \method Benchmark}.
We conduct a comprehensive human evaluation using our benchmark to assess the performance of generated images, employing image scoring as the evaluation metric.
Specifically, we score each image considering two aspects: prompt following and image quality. The prompt following metric measures the image compliance with text instructions or text descriptions, and is categorized into five levels. The image quality metric encompasses various aspects such as generated color, details, layout, and visual appeal, and is scored on a scale from 1 to 5. 
Considering the broad capabilities of our method, we compare it with several common approaches and some experts designed for specific tasks. We engaged 5 professional designers as evaluators to carry out these assessments. For each task, the data is evenly distributed among the evaluators in an anonymous manner, and scores are aggregated for analysis.

As shown in ~\cref{tab:user_study}, we compare our approach across multiple global editing tasks and local editing tasks. The prompt following score and image quality score are presented together, separated by a “/” pattern. The bold numbers represent the best, and the underlined numbers indicate the second best.
Our method achieves the highest prompt following scores in 7 of 12 global editing tasks and 8 of 10 local editing tasks, which demonstrates that \method fully understands the intention of the instruction and is able to correctly generate an image that meets the instruction.
Furthermore, \method achieves the best image quality scores in 5 of 10 global editing tasks and 7 of 10 local editing tasks. These results indicate that \method excels at generating high aesthetic images across various image editing tasks. 
Nonetheless, our method performs unsatisfactorily in certain tasks, such as general editing and style editing. 
One possible reason is that images generated by methods using larger models, such as those producing 1024-resolution images based on the SDXL model, are more preferred by evaluators compared to those produced by our model, which has a size of 0.6B parameters and an output resolution of around 512.

\section{Conclusion}\label{sec:conc}

We propose ACE, a versatile foundational generative model that excels at creating images, and following instructions across a wide range of generative tasks. 
Users can specify their generation intentions through customized text prompts and image inputs. Furthermore, we advance the exploration of capabilities within interactive dialogue scenarios, marking a significant step forward in the processing of long contextual historical information in the field of visual generation. 
Our work aims to provide a comprehensive generative model for the public and professional designers, serving as a productivity enhancement tool to foster innovation and creativity.

\textbf{Acknowledgments.} \label{sec:ack}
We sincerely appreciate the contributions of many colleagues for their insightful discussions, valuable suggestions, and constructive feedback, including: 
Haiming Zhao, Yuntao Hong, You Wu, Jixuan Chen, Yuwei Wang, and Sheng Yao for their data contributions, and Lianghua Huang, Kai Zhu, and Yutong Feng for their discussions, suggestions, and the sharing of resources.

\clearpage
\bibliography{iclr2025_conference}
\bibliographystyle{iclr2025_conference}

\clearpage
\appendix

\section{Related Work}\label{sec:related}
Visual generation, which takes multi-modal conditions (\textit{e.g.}, textual instruction and reference image) as input to generate creative image, has emerged as a popular research trend in recent years.
As the basic task, text-guided image generation has undergone a significant development, marked by remarkable advancements in recent years. Many approaches ~\citep{glide,imagen,dalle2,ldm,sdxl,dalle3,midjourney,wanx,ernie_vilg,pixart,sd3,kolors,hunyuan_dit,flux} 
have been proposed and achieved impressive results in terms of both image quality and semantic fidelity.
By incorporating low-level visual features as input, ~\citet{composer} and ~\citet{controlnet} pave the way for the initial forms of multi-modal controllable generation.
Recently, some approaches 
~\citep{t2i-adapter,uni_controlnet,unicontrol}
have tried to use multiple visual features as conditions, facilitating the multi-modal controllable generation. 
By integrating fine-tuning technologies such as ~\citet{dreambooth,lora}, these approaches have further enabled the customization of diverse controllable generation applications. 
Another popular trend is image editing technology  ~\citep{ipadapter,stylebooth,instantid,consistentid,instantstyle,facechain,anytext,anydoor,largen,stylediff,smartbrush,imagebrush,smartedit,iedit,dragdiff,layerdiff,sdedit}, which focus on editing input images according to text prompts and preserving some identity such as person, scene, subject, or style. 
While the above models excel at generating image in one specific task or scenario, they have difficulty in extending to unseen tasks. 
To address the aforementioned challenges, some methods have been introduced to edit input images by following natural language instructions
~\citep{ip2p,cosxl,instructdiff,emuedit,ultraedit,seedx} which is more flexible to implement various tasks within a single model. 
However, a key bottleneck for these methods lies in the construction of high-quality instruction-paired datasets with annotated edits, which cause limited generalizability and suboptimal performance. 
In this paper, we focus on establishing a unified definition for multi-modal generation problems. Based on this definition, we aim to construct higher-quality, annotated data and instruction sets further to develop a unified foundational model for multimodal generation.

\section{Datasets Detail}\label{sec:datasets_detail}

\begin{figure*}[t]
	\scriptsize
	\centering
    \includegraphics[width=\linewidth]{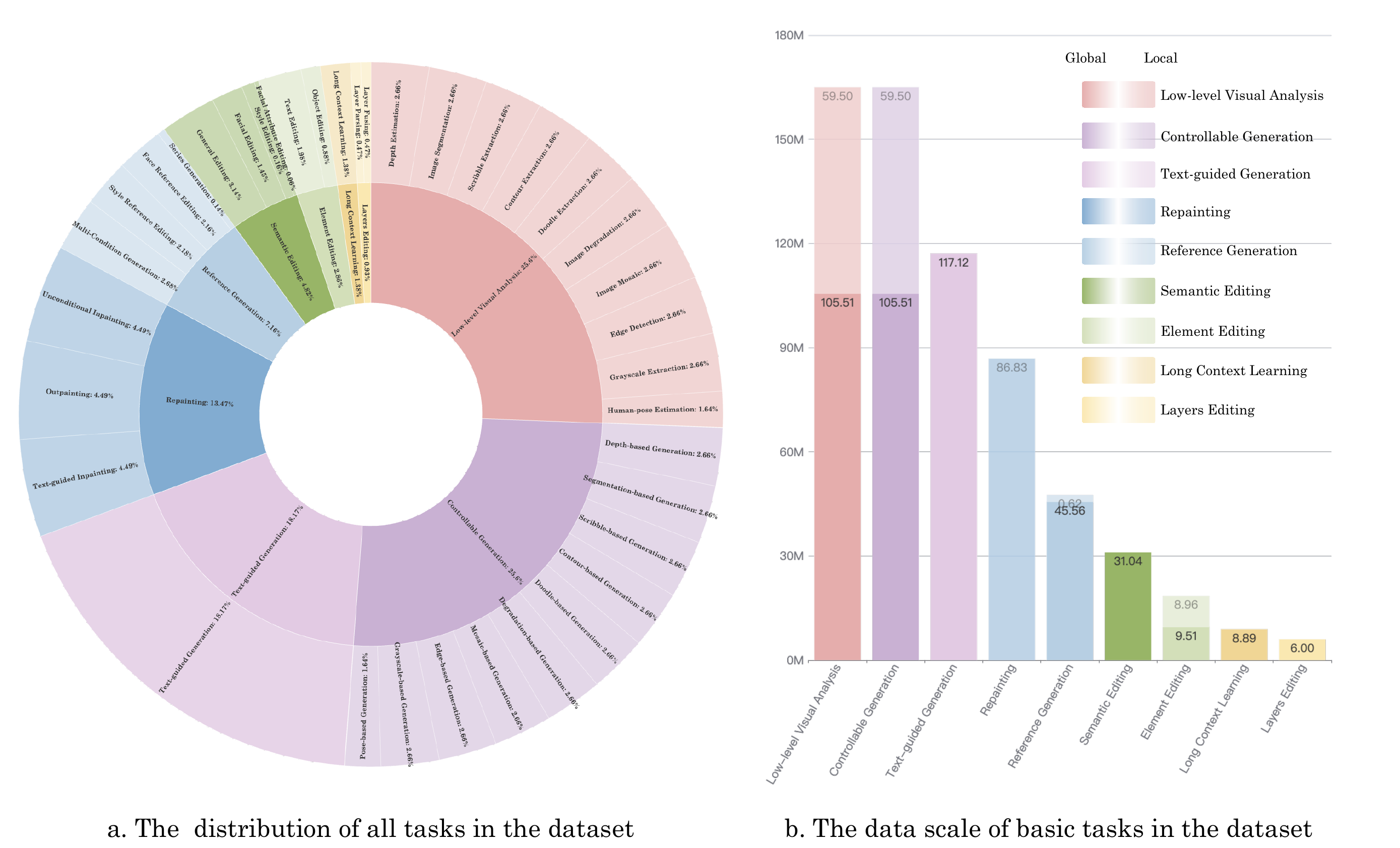}
        \
	\caption{\textbf{Statistics on the data scale for various tasks.} We collect 0.7 billion data pairs, which cover \textbf{8} basic types including \textbf{37} subtasks, multi-turn and long-context generation datasets.}
    \label{fig:train_dist}
\end{figure*}

We use an internal dataset of 0.7 billion data pairs to train a foundational model for generation and editing. 
The supported tasks include \textbf{8} basic types consisting of \textbf{37} subtasks, as well as a multi-turn and long-context generation task. 
These tasks use textual instructions along with zero or more reference images for generating or editing image.
The data distribution is depicted in ~\cref{fig:train_dist}a, and the absolute data scale is illustrated in ~\cref{fig:train_dist}b. 
In this section, we provide a detailed introduction to the data construction methods for various tasks.

\subsection{Text-guided Generation}

We collect approximately 117 million images and use MLLM model to supplement captions for images, creating pair data for text-to-image tasks. Additionally, this portion of the data serves as an intermediary bridge in various generation and editing tasks, allowing the combination of different task instructions to obtain pairs from original images to target images.

\subsection{Low-level Visual Analysis}

\begin{figure*}[t]
	\scriptsize
	\centering
        \includegraphics[width=\linewidth]{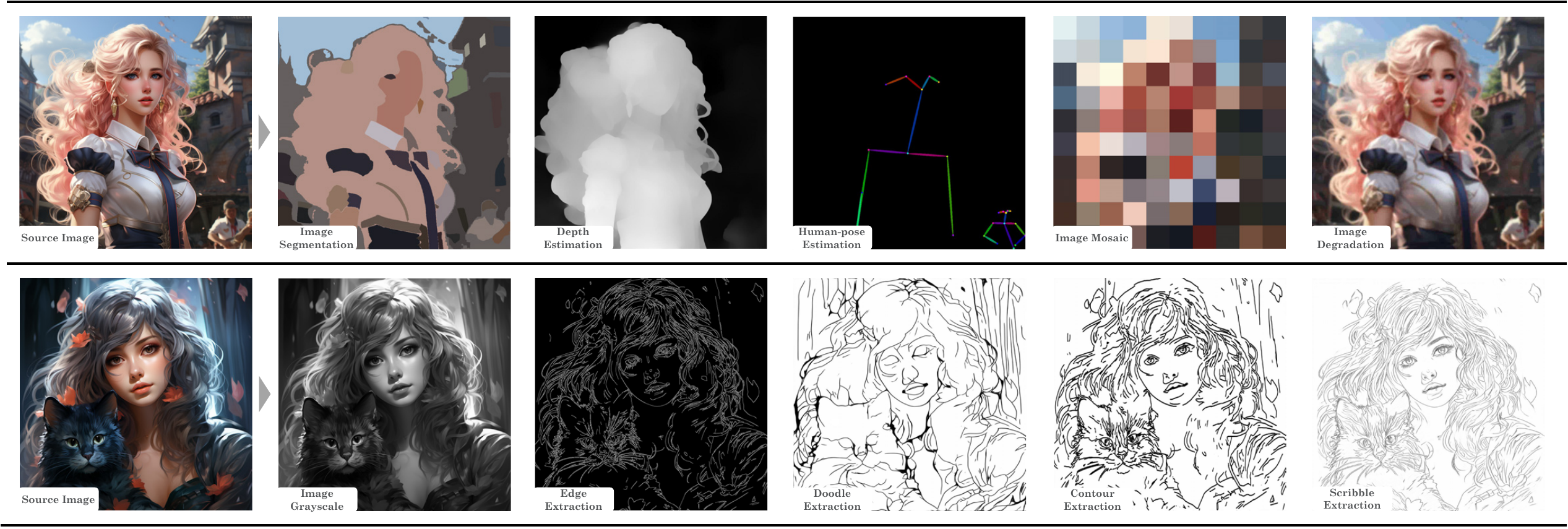}
	\caption{The visualization of low-level visual analysis preprocessing.}
    \label{fig:sample_lowlevel}
\end{figure*}

\begin{figure*}[t]
	\scriptsize
	\centering
        \includegraphics[width=\linewidth]{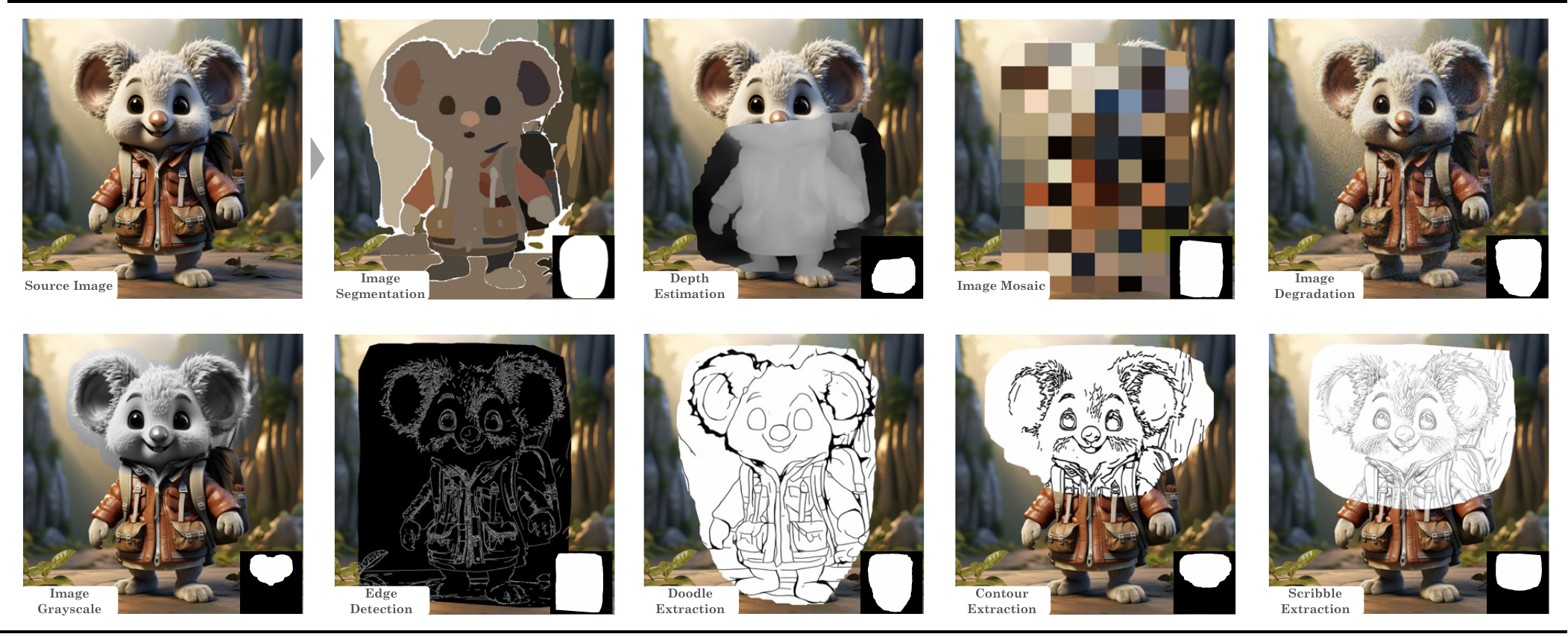}
	\caption{The visualization of regional low-level visual analysis preprocessing.}
    \label{fig:sample_lowlevel_regional}
\end{figure*}

Low-level Visual Analysis tasks involve analyzing and extracting various low-level visual features from a given image, like an edge map or segmentation map. These low-level visual features are typically employed as control signals in the controllable generation.
We select 10 commonly used low-level features in the controllable generation, including segmentation map, depth map, human pose, mosaic image, blurry image, gray image, edge map, doodle image, contour image, and scribble image. 
The visual features extracted at global and local levels are illustrated in ~\cref{fig:sample_lowlevel} and ~\cref{fig:sample_lowlevel_regional}, respectively. 

\begin{itemize}[leftmargin=10pt]
\setlength{\parsep}{0pt}
\setlength{\parskip}{0pt}
\item \textbf{Image Segmentation} involves extracting image spatial region information for different targets within an image. This is achieved by selecting and modifying specific areas for operations and editing in downstream tasks. We employ the Efficient SAM~\citep{efficientsam} tool for marking different target areas within an image. 
\item \textbf{Depth Estimation} indicates the relative distance information of different targets within an image. We use the Midas~\citep{midas} algorithm to extract depth information. 
\item \textbf{Human-pose Estimation} is employed for modeling the human body to obtain structured information about body posture. We make use of the RTMPose~\citep{rtmpose} algorithm to extract information from images containing human figures, and posture information visualization is done using OpenPose's 17-point~\citep{openpose} modeling method. 
\item \textbf{Image Mosaic} pixelates specific areas or the entire image to protect sensitive information. 
\item \textbf{Image Degradation} is used to degrade the quality of an image to simulate the phenomenon of image distortion found in the real world. Following the practice of super-resolution algorithms~\citep{realesrgan}, we add random noise to the input images. 
\item \textbf{Image Grayscale} is typically done to facilitate the editing of an image's original colors downstream. We do this conversion directly using OpenCV's Grayscale function. 
\item \textbf{Edge Detection} detects the edge information from the original image. We utilize the edge detection method named Canny~\citep{canny} implemented by OpenCV. 
\item \textbf{Doodle Extraction} is usually used to simulate relatively rough hand-drawn sketches by extracting the outline of objects and ignoring their details. We use the PIDNet~\citep{pidnet} and SketchNet~\citep{sketchnet} to extract this information. 
\item \textbf{Contour Extraction} is about delineating the outline of targets within an image, which simulates the drawing process of the image and is often used for secondary processing of images. We use the contour module from the informative drawing~\citep{infordraws} for this information extraction. 
\item \textbf{Scribble Extraction} involves retrieving the original line art information to capture the sketch-like form of the image. We utilize the anime-style module from informative drawings~\citep{infordraws} to extract the relevant information. 
\end{itemize}


\subsection{Controllable Generation} \label{sec:datasets_control}

In the realm of vision-based generative foundation models, the ability to generate corresponding content using any provided prompts is commonly present. 
To further control aspects such as spatial layout, structure, or color in the generated images, additional conditional information is often incorporated as inputs to the model. 
We integrate various controllable condition-to-image tasks within a unified framework to accommodate different visual conditions. The control conditions include the visual features mentioned in the low-level visual analysis section. For training data, we employ pairs constituted by the aforementioned control conditions in ~\cref{fig:sample_lowlevel} and regional control conditions in ~\cref{fig:sample_lowlevel_regional} obtained through low-level visual analysis, using the conditional part as inputs to the model to achieve pixel-precise image generation. For text guidance, we construct the instructions based on image captions with our proposed Instruction Captioner.

\subsection{Semantic Editing}  \label{sec:datasets_semantic Editing}

Semantic Editing aims to modify specific semantic attributes of an input image by providing detailed instructions. 
It involves facial editing, which aims to modify partial attributes of characters while preserving the overall identity, and style transforming, which aims to transform the image style to a specific artist theme guided by instruction while keeping content unchanged.
Additionally, any other semantic editing requests that do not fall into these two categories are classified as general editing, \textit{e.g.}, changing the background scene of an image, adjusting a subject's posture, and modifying the camera view.
We discuss the specifics according to the particular tasks.

\subsubsection{Facial Editing}  \label{sec:datasets_face}

Facial Editing encompasses both the transformation and preservation of facial attributes. Specifically, the facial attributes preservation task focuses on editing other elements of the image while maintaining the consistency of complex identity details in facial representations. The facial attributes transformation task is primarily concerned with altering specific attributes of the face without affecting other aspects of the image.

\begin{figure*}[t]
	\scriptsize
	\centering
        \includegraphics[width=\linewidth]{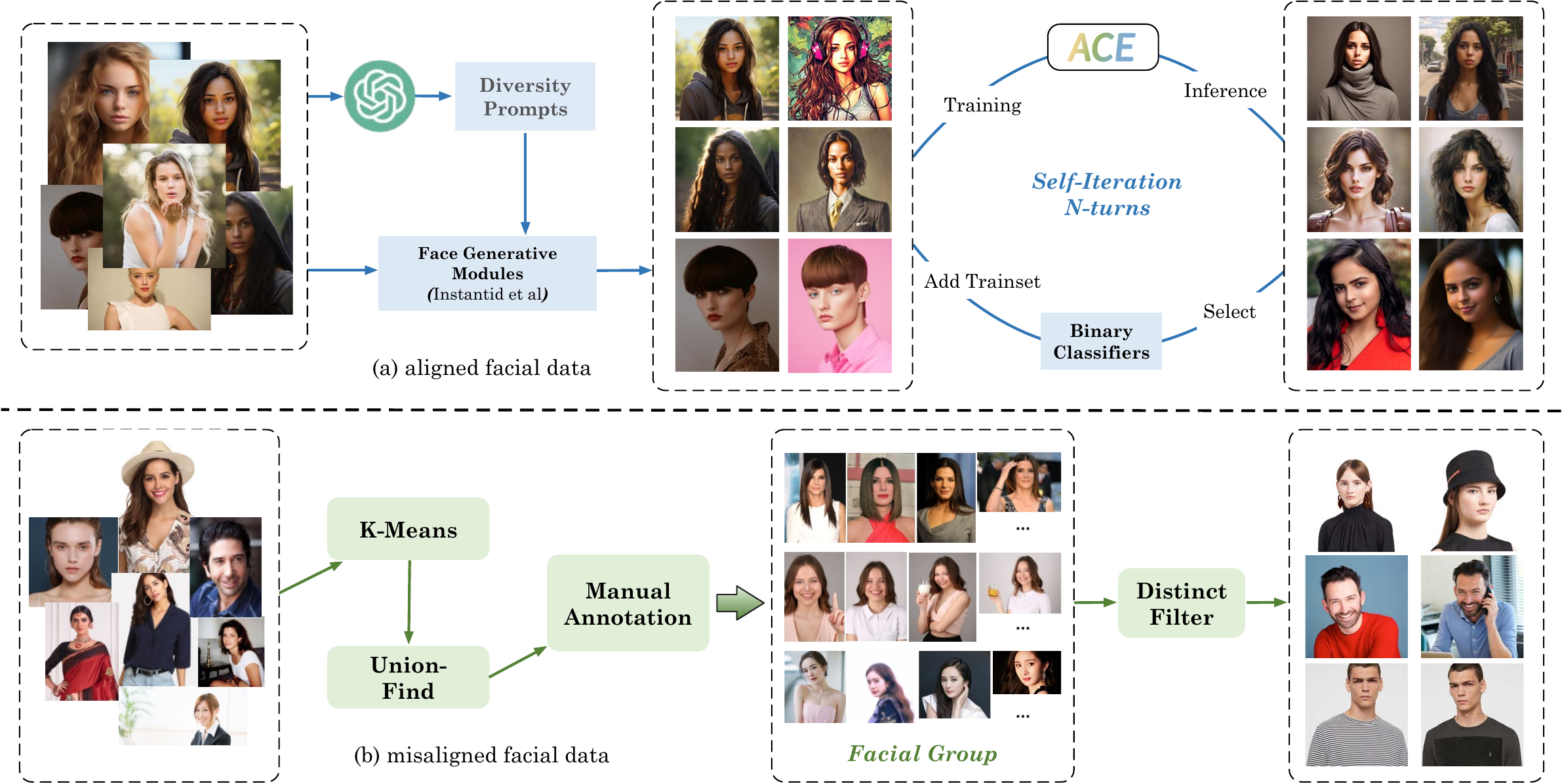}
	\caption{Illustration of facial editing data processing workflow. }
    \label{fig:pipe_face}
\end{figure*}

\begin{figure*}[t]
	\scriptsize
	\centering
    \includegraphics[width=\linewidth]{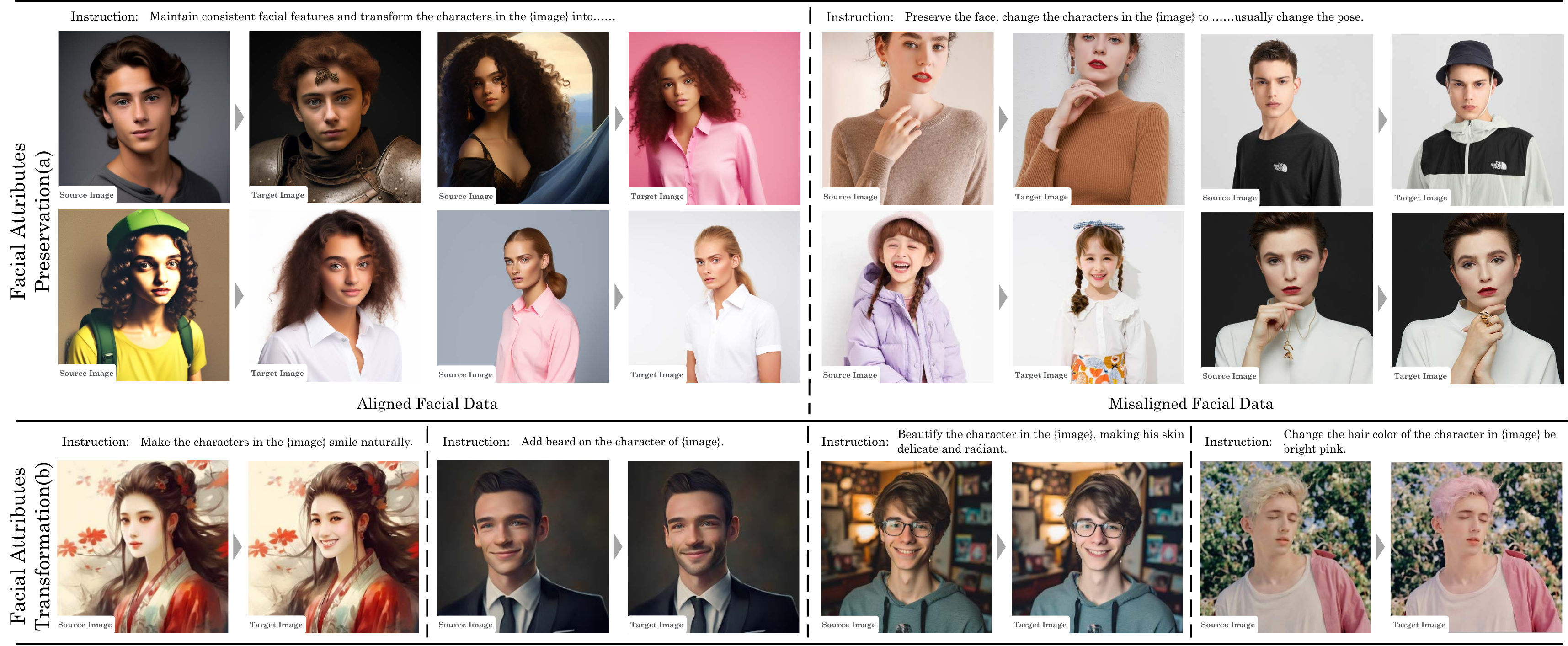}
	\caption{The dataset visualization of facial editing.}
    \label{fig:sample_face}
\end{figure*}

\textbf{Facial Attribute Preservation.}
The facial attribute preservation dataset is divided into two main parts: aligned and misaligned facial data as shown in \cref{fig:sample_face}a. There are two novel processing workflows as shown in \cref{fig:pipe_face}.
\textbf{(i) Aligned facial data.}
We generate pixel-aligned face data using generative models such as InstantID~\citep{instantid} and combine it with GPT models to produce diverse prompts. Subsequently, we train multiple lightweight binary classification models to clean the generated data based on 
image quality, PPI score, aesthetic scores, and other metrics. Additionally, we extract facial features using ArcFace~\citep{arcface} for similarity calculations, selecting high-matching data pairs with a similarity score exceeding 0.65. Once our model demonstrates the ability to maintain facial integrity, we initiate a self-iterative training process to generate higher quality data, as illustrated in \cref{fig:pipe_face}a.
\textbf{(ii) Misaligned facial data.} \label{sec:face_cluster}
We first employ a face detection algorithm~\citep{mtcnn} to filter images containing only one face. Subsequently, we utilized facial features to perform K-means clustering, resulting in 10,000 clusters.
Within each cluster, we conducted a second clustering using the union-find algorithm. Faces with a similarity score greater than 0.8 and less than 0.9 were linked to avoid grouping perfectly identical images.
Finally, manual annotation and deduplication were performed on the remaining clusters, yielding the final unaligned facial dataset as shown in \cref{fig:pipe_face}b.
Based on the general instruction construction process in \cref{sec:data_instr}, we design the instructions for facial editing to emphasize that the individuals in the image pairs being annotated are the same person. The instructions must reflect this and focus on the differences in personal details between the two images.

\textbf{Facial Attribute Transformation.}
We add four fine-grained facial attribute transformation tasks: smiling, beard, makeup, and hair dyeing. We obtained the relevant data in bulk by calling the Aliyun API and trained binary classifiers for each category to filter out data with indistinct changes. As a result, we acquired a total of 1.4 million high-quality pairs of data as shown in \cref{fig:sample_face}b. 
Equally, we strive to guide the generated captions to closely reflect the facial attributes, thereby enhancing the model's understanding of the similarities and differences in tasks related to facial attributes.

\subsubsection{Style Editing} \label{sec:datasets_style}
\begin{figure*}[t]
	\scriptsize
	\centering
    \includegraphics[width=\linewidth]{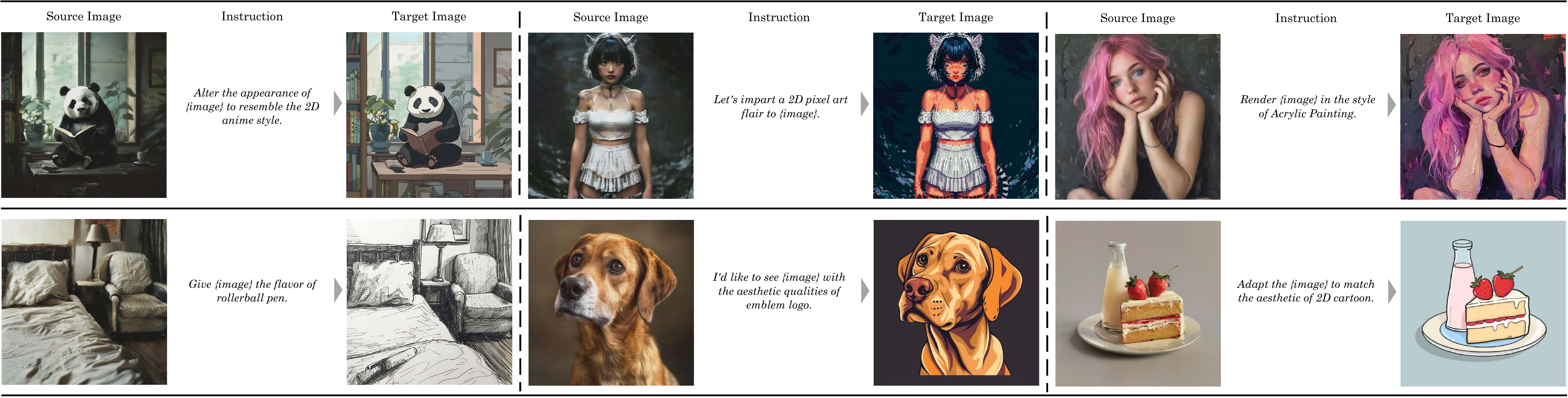}
        \
	\caption{The dataset visualization of style editing.}
    \label{fig:sample_style}
\end{figure*}
Following the similar image pair construction strategy from StyleBooth~\citep{stylebooth}, we prepare a larger training data that encompasses over 80 styles and 63000 image pairs. Besides, additional real-world and synthesized style images are collected as style editing target images, and their corresponding ``original" images are generated by transforming these collected images to different graphic styles such as cinematic, photography, \textit{etc}. In this way, we obtain around 70000 input and output image pairs of about 400 high-quality styles. 
We show samples of the final style editing data in ~\cref{fig:sample_style}.

We conduct different filter strategies to leverage the data quality: (i) Like StyleBooth, we use CLIP score as the metric to filter out the image pairs which have too minor or too great differences. 
(ii) To further filter out the faultful synthesized target images that are not particularly aligned with the provided prompt keywords in terms of style, we use CSR~\citep{csr} representations and implement style clustering within every style subgroup. Setting a threshold of 0.65, cosine similarities are calculated for union-find clustering. The largest cluster contains images in a similar visual style while other clusters are filtered out.

\subsubsection{General Editing}
\begin{figure*}[tbp]
  \scriptsize
  \centering
  \includegraphics[width=\linewidth]{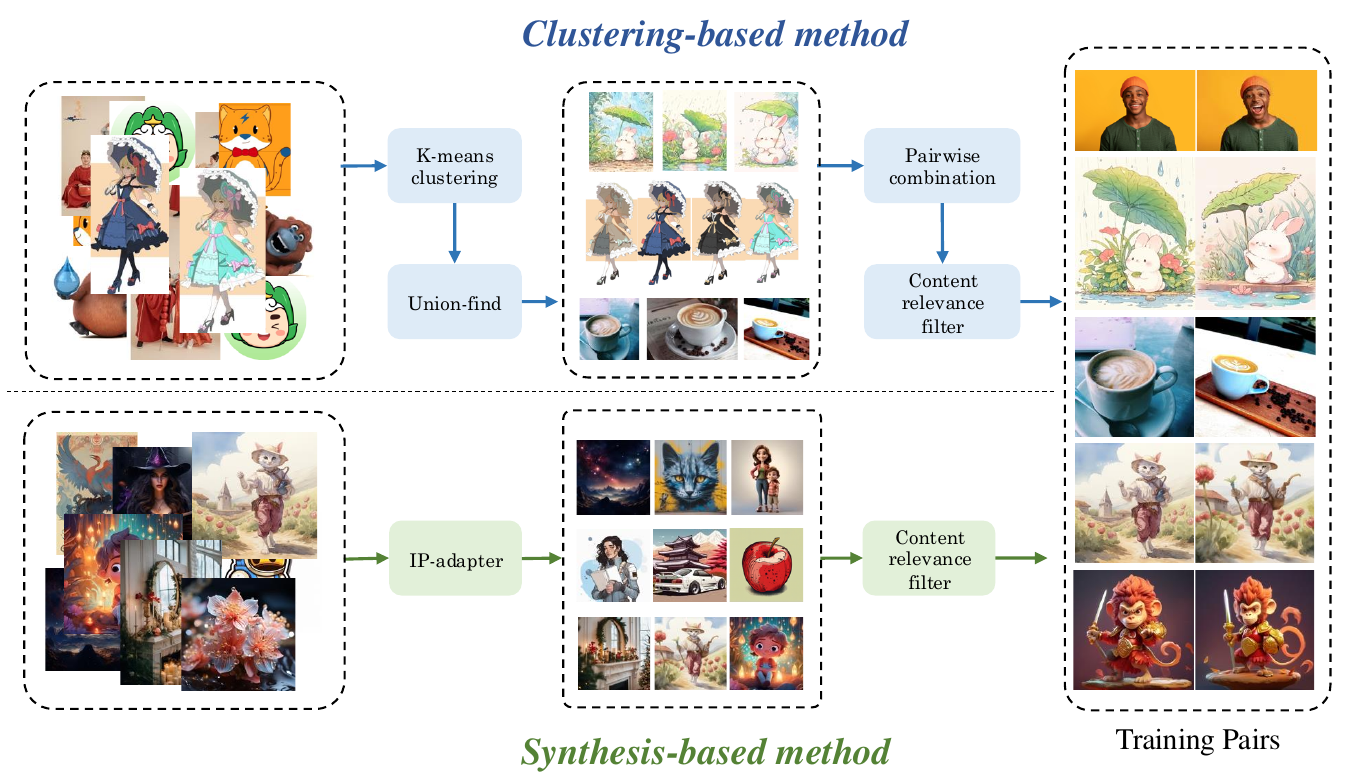}
  \caption{The dataset construction pipeline for general editing task.}
  \label{fig:pipe_general}
\end{figure*}

\begin{figure*}[h]
  \scriptsize
  \centering
  \includegraphics[width=\linewidth]{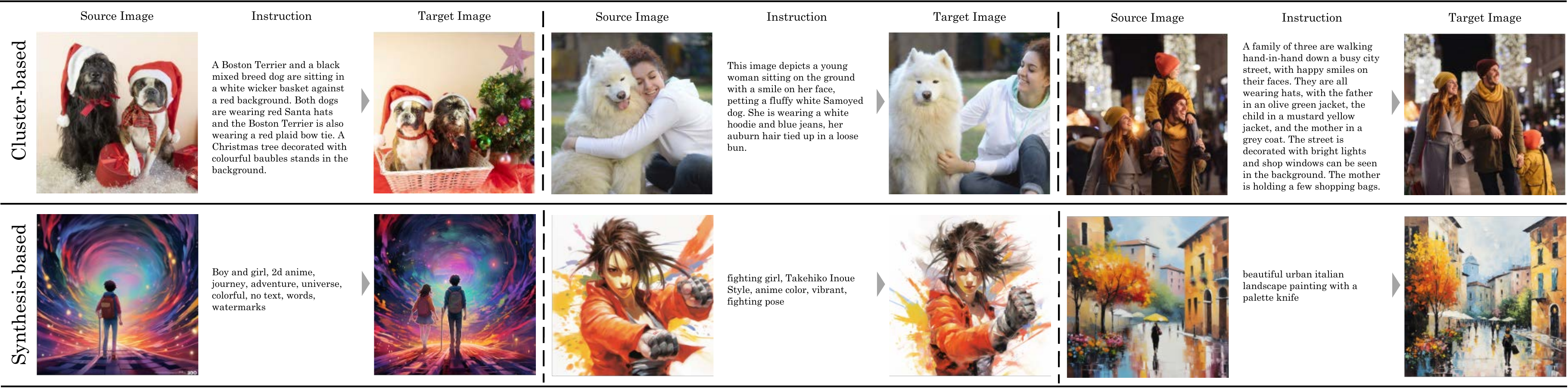}
  \caption{\textbf{General editing sample pairs generated by our dataset construction pipeline.} Image pairs in the first row are generated by cluster-based method, and those in the second row are generated from synthesis-based method.}
  \label{fig:sample_general}
\end{figure*}

The objective of general editing is to curate an image that seamlessly harmonizes with both textual and visual prompts for a variety of purposes. It involves two tasks, \textit{i.e.}, caption-guided image generation and instruction-guided image adaption. 
The former task receives one reference image and one caption as prompts to generate the image, and the latter task intends to adapt the source image by following the given instructions. The training data for these two tasks can be unified into the same format, which consists of a content-related image pair $(I_{\text{source}}, I_{\text{target}})$, and a text prompt indicates how to generate target image.
An essential goal of building such a training dataset is to acquire content-related image pairs, one of which serves as the source image and another serves as target image. 
The overall dataset construction pipeline is depicted in \cref{fig:pipe_general}. It includes two branches, \textit{i.e.}, \textbf{clustering-based} method, and \textbf{synthesis-based} method.

\noindent\textbf{Cluster-based method.}
We employ embedding-based clustering on the database to group content-similar images. Union-find technology is employed inside each cluster to achieve more fine-grained image pair aggregation.
We then collect all possible pairs from each disjoint set. Additionally, a binary classifier evaluates the content relevance of pairs, and those with low relevance are discarded.

\noindent\textbf{Synthesis-based method.}
We use IP-Adapter technology to synthesize images according to the reference images and text prompts, thus the content-related image pairs can be obtained. To ensure visual content is similar but not the same,  we set the image control strength $\lambda$ to 0.6, and a binary classifier is utilized to filter out content-unrelated pairs.
We depict some generated samples in \cref{fig:sample_general}.

For the text prompt of each image pair, we use the MLLM to generate both a caption that describes the visual content of the target image, and an instruction that indicates how to adapt the source image to the target image, as described in \cref{sec:data_instr}.

\begin{figure*}[t]
  \scriptsize
  \centering
  \includegraphics[width=\linewidth]{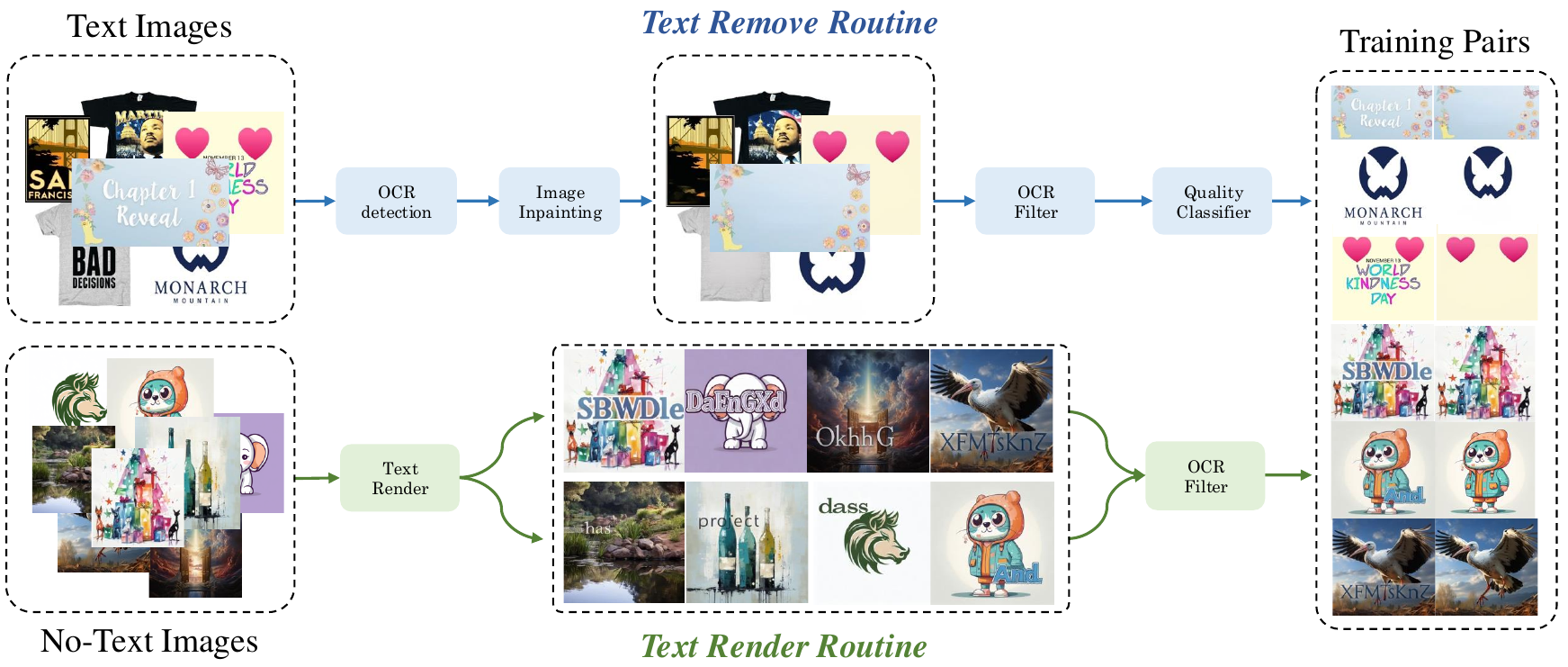}
  \caption{The pipeline of building training data of text editing.}
  \label{fig:pipe_text}
\end{figure*}

\begin{figure*}[!ht]
  \scriptsize
  \centering
  \includegraphics[width=\linewidth]{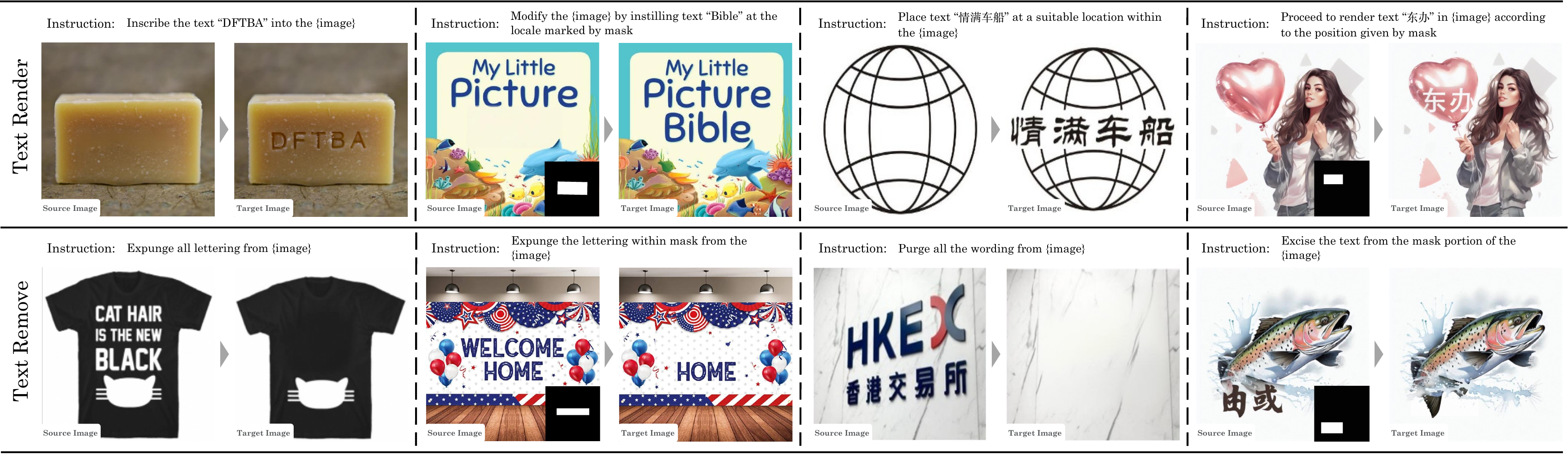}
  \caption{The dataset visualization of multi-lingual text editing.}
  \label{fig:sample_text}
\end{figure*}

\begin{figure*}[!ht]
  \scriptsize
  \centering
  \includegraphics[width=\linewidth]{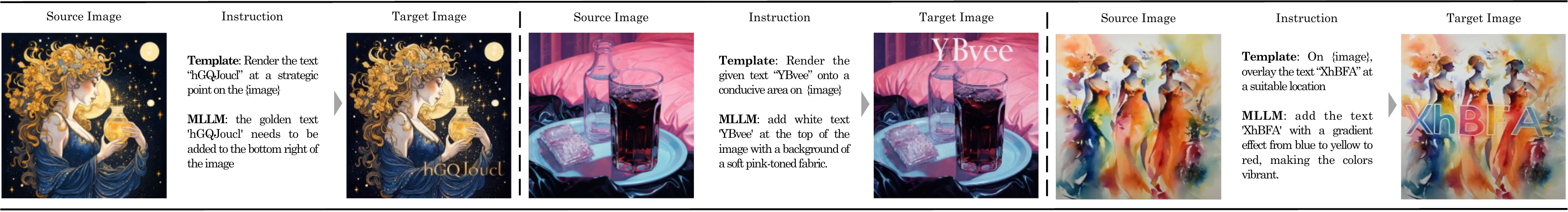}
  \caption{Template-based instructions and MLLM-based instructions on text editing.}
  \label{fig:text_inst}
\end{figure*}

\subsection{Element Editing}

Element editing focuses on the selective manipulation of specific subjects within an image. 
This process allows for the addition, deletion, or replacement of a particular subject while ensuring that the other elements within the image remain unchanged.
By doing so, the integrity of the overall composition is preserved, allowing users to make precise edits and achieve desired alterations without disrupting the context of the original scene.
We focus on two common elements: text and objects.

\subsubsection{Text Editing}
Text editing is an important task of element editing. 
Despite the progress gained in image generation, the capability of text rendering is still far from satisfying. 
Therefore, text editing is a necessary technology to revise the incorrect or deformed text rendered in image. 
Text editing involves text removing task, which is to erase text from image while preserving all other visual cues, and text rendering task, which is to render specific text at any location of an image. 
The goals of these two tasks are exactly the opposite, hence their training data can be shared to each other.
For instance, for any image pair $\{I_{a}, I_{b}\}$, suppose the text removing represents the generation direction from $I_{a}$ to $I_{b}$, on the contrary, the generation direction from $I_{b}$ to $I_{a}$ stands for text rendering.
Therefore, the objective of constructing the dataset thus becomes how to obtain a large number of image pairs, where one image contains the specified text and the other does not while keeping the non-text content unchanged.

We propose a two-branch data collection method to address this issue. The overall pipeline shown in \cref{fig:pipe_text} includes two paths:
(i) \textbf{Text remove path}. 
For images containing text data, which typically from publicly available text datasets such as AnyWord3M~\citep{anytext} and LaionGlyph-10M~\citep{laionglyph}, we first mask out all text regions. Then, we redraw the masked areas leveraging image inpainting method. To ensure the regenerated image does not contain any textual information, we employ OCR detection leveraging the open-sourcing OCR model (\textit{e.g.}, PP-OCR)~\citep{ppocr} and filter out all images that contain any texts. Finally, we adopt an image quality score predictor which is trained with small amounts of manually annotated data to score all text-removed images and pick high-quality samples according to the score.
(ii) \textbf{Text render path}. 
For any image dataset, We first employ OCR detection to ensure input images contain no text. 
Then random characters are rendered in random locations of these images by utilizing existing text editing methods (\textit{e.g.}, AnyText)~\citep{anytext}. 
We render text using Chinese or English characters to support multi-lingual text rendering capability. We depict some cases in \cref{fig:sample_text}. 
Finally, we implement OCR detection on the edited image to ensure all characters are rendered correctly.
When training, image pairs collected from both two paths are merged to form the total dataset.

We adopt template-based and MLLM-based methods to construct instructions that describe how to render or remove text from the input image. 
For MLLM-based method, besides the content of characters, we add extra color and position controls by specifying the text color and render position in the textual instruction. Given a text image, we utilize a pre-trained MLLM to describe the color, content, and position information of text in this image, thus a text editing instruction can be easily inferred based on these descriptions. Some cases of template-based and MLLM-based instructions are illustrated in \cref{fig:text_inst}.

\subsubsection{Object Editing}

\begin{figure*}[t]
	\scriptsize
	\centering
        \includegraphics[width=\linewidth]{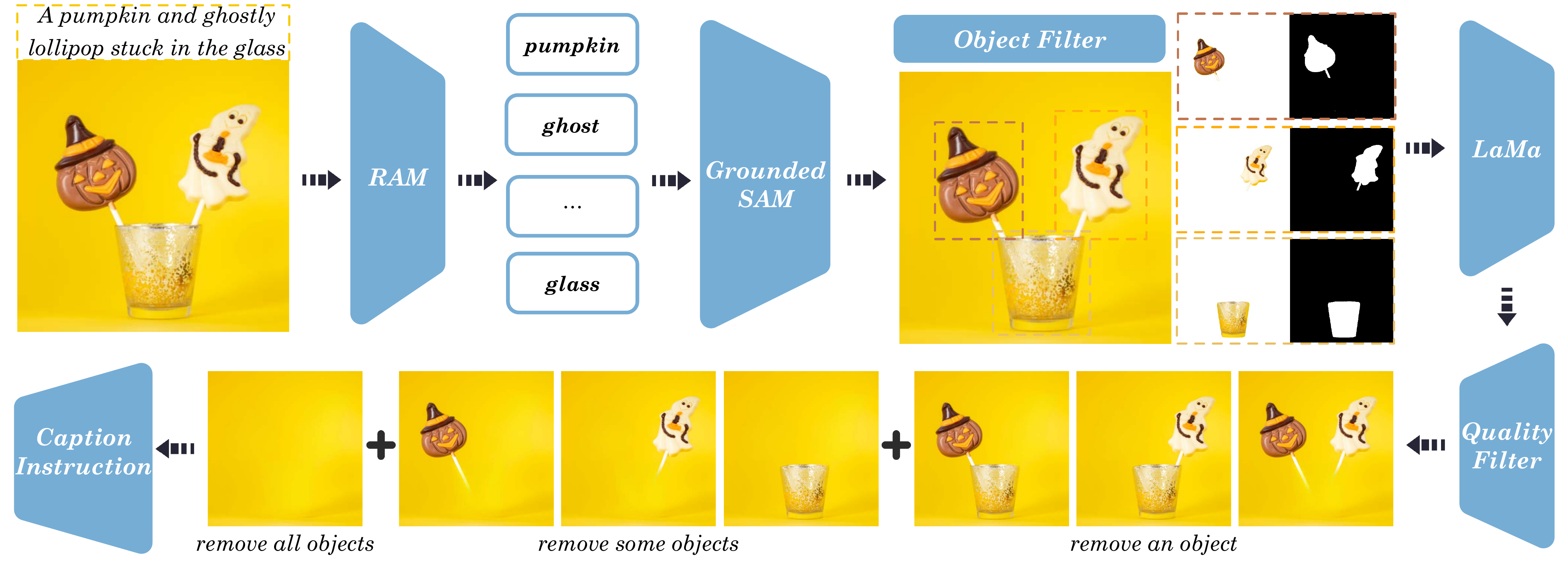}
        \caption{Illustration of data construction pipeline for object editing in element editing.}
    \label{fig:pipe_object}
\end{figure*}

\begin{figure*}[t]
	\scriptsize
	\centering
        \includegraphics[width=\linewidth]{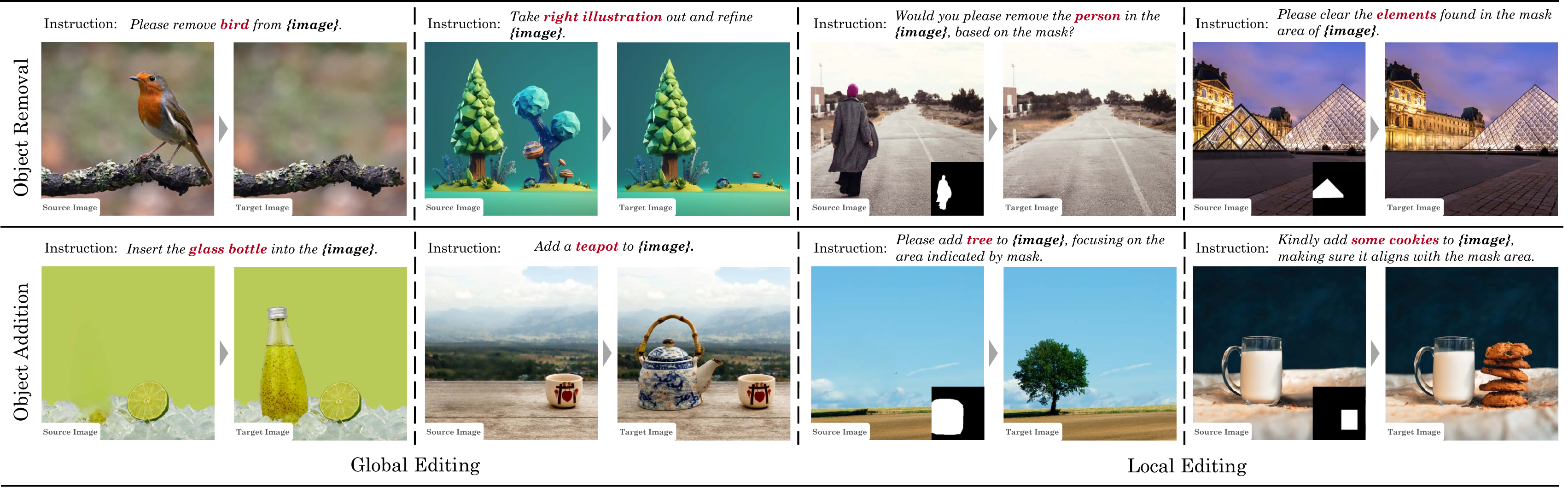}
	\caption{The dataset visualization of object editing in element editing.}
    \label{fig:sample_object}
\end{figure*}

Object-based image editing is one of the most commonly used techniques for creatively manipulating images. Its primary goal is to either remove or add objects in an image based on text instructions provided by the user, while ensuring a harmonious composition. To obtain training data for this task, we need to construct a pair of data to indicate the presence relationship of objects. Specifically, we focus on images that either contain a specific object or do not, ensuring that all other parts of the images remain as unchanged as possible, except for the area where the object is located.

We can see the entire dataset process from ~\cref{fig:pipe_object}.
We first utilize RAM~\citep{ram} for open-label tagging, obtaining semantic labels for different subjects in the image, and then applying Grounded-SAM~\citep{groundingdino, sam} to segment the input semantics. Next, we perform a preliminary screening of objects based on filtering criteria including the area of the masks and bounding boxes, as well as their effective ratios, removing any unreasonable subjects. We then use the LaMa~\citep{lama} method, combining the original image with the subject mask area for inpainting. This operation effectively removes local objects without significantly affecting other areas. Finally, we employ a pre-trained binary classification model to determine whether the inpainted image meets expectations, filtering out artifacts introduced by the inpainting algorithm. In terms of instruction formulation, we employ a template format that incorporates the \texttt{\{object\_name\}} tag, while also utilizing a common instruction based on image pairs.

Through the data construction pipeline, we can obtain the original image, the image with the object removed, the object mask, and the corresponding text instructions. This way, we can implement a forward pipeline for object removal and a reverse pipeline for object addition, while ensuring the integrity of the image and the accuracy of the text instructions, as in ~\cref{fig:sample_object}.

\subsection{Repainting}

\begin{figure*}[t]
	\scriptsize
	\centering
        \includegraphics[width=\linewidth]{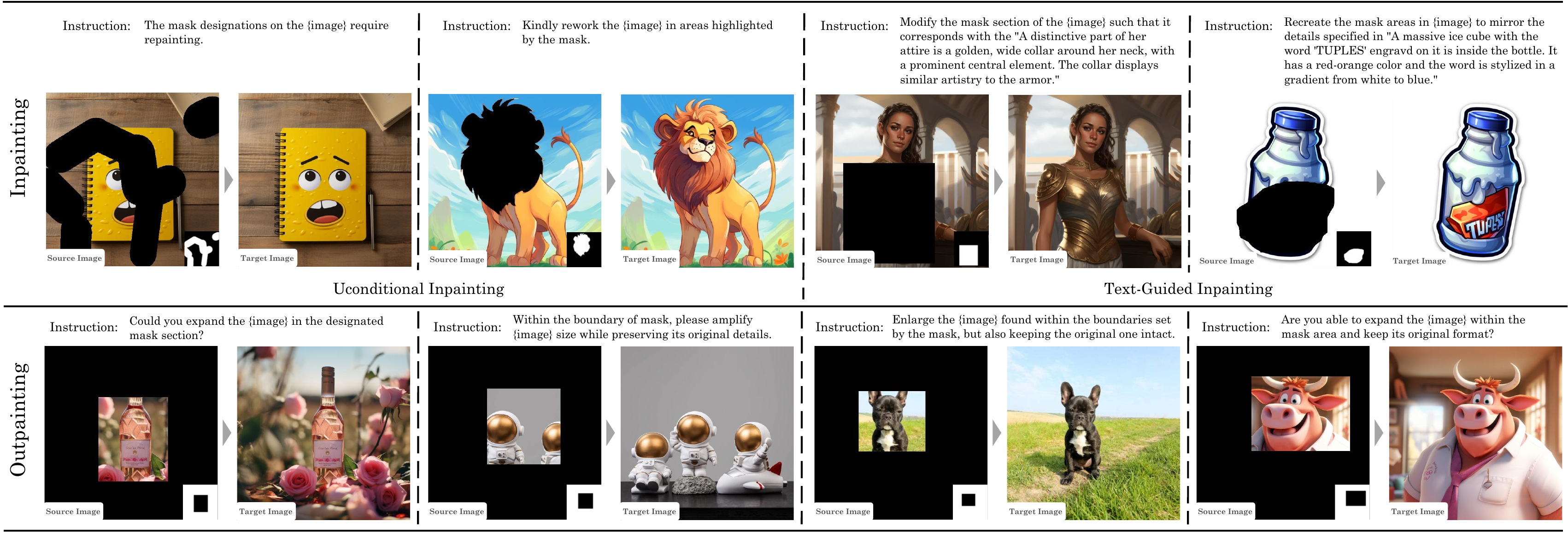}
	\caption{The dataset visualization of Repainting.}
    \label{fig:sample_repaint}
\end{figure*}

The repainting task can be defined as the process of reconstructing missing image information within specified masked regions. Depending on the location of the masked area and input conditions, this task can be categorized into three distinct types: unconditional inpainting, text-guided inpainting, and outpainting.
Some examples of training data are shown in ~\cref{fig:sample_repaint}.

\subsubsection{Unconditional Inpainting}

Unconditional image inpainting typically utilizes methods such as low-level textual information and Fourier Convolutions, combined with contextual information from the known areas of the image, to reconstruct the missing portions. 
This process usually requires an input consisting of an image to be inpainted and a mask indicating the regions that need to be filled, leading to an output image where the missing areas are completed. 
The task demands that the original information is preserved and that there is a high-quality seamless integration between the original and the filled-in areas. 
By employing LaMa's~\citep{lama} mask generation strategy, we randomly apply bbox or irregular-shaped masks to the images and vary the degree of this operation to enable the model to handle different types of missing regions as effectively as possible.

\subsubsection{Text-guided Inpainting}

Text-guided inpainting primarily aims to fill and restore missing parts of an image by utilizing text descriptions to guide the process. 
Unlike traditional unconditional inpainting, this method integrates textual information to guide the model, resulting in images that better meet the user's specific requirements. 
In constructing this dataset, we not only employ random masks paired with corresponding textual descriptions of the original images but also refine the process to focus on local regions. 
First, we obtain multiple object masks from the image, and then extract detailed textual descriptions for each object. Finally, we create triplets consisting of the original image, the local object mask, and the local object caption. 
This approach enables the generation of richer and more controllable details within local areas.

\subsubsection{Outpainting}

The outpainting task involves intelligently generating and completing the edge regions of an existing image so that the extended new image appears natural and continuous visually. 
The major challenge of this task is producing images that are rich in detail, diverse in content, and exhibit a certain level of associative ability. 
In terms of data processing, we employed commonly used techniques, applying random masks to the areas and directions that need to be expanded, in order to adapt to different scenarios of image completion.

\subsection{Layer Editing}

\begin{figure*}[t]
	\scriptsize
	\centering
        \includegraphics[width=\linewidth]{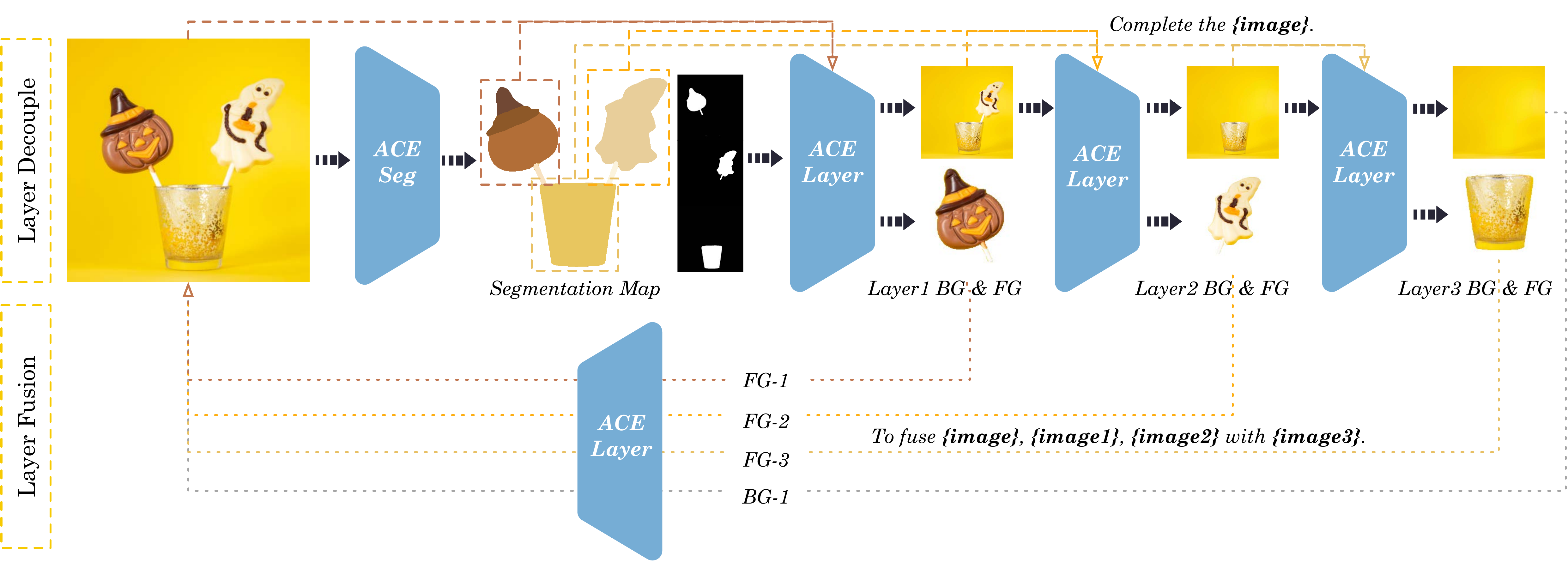}
        \caption{Illustration of inference pipeline layer decouple and layer fusion in layer editing.}
    \label{fig:pipe_layer}
\end{figure*}

\begin{figure*}[t]
	\scriptsize
	\centering
        \includegraphics[width=\linewidth]{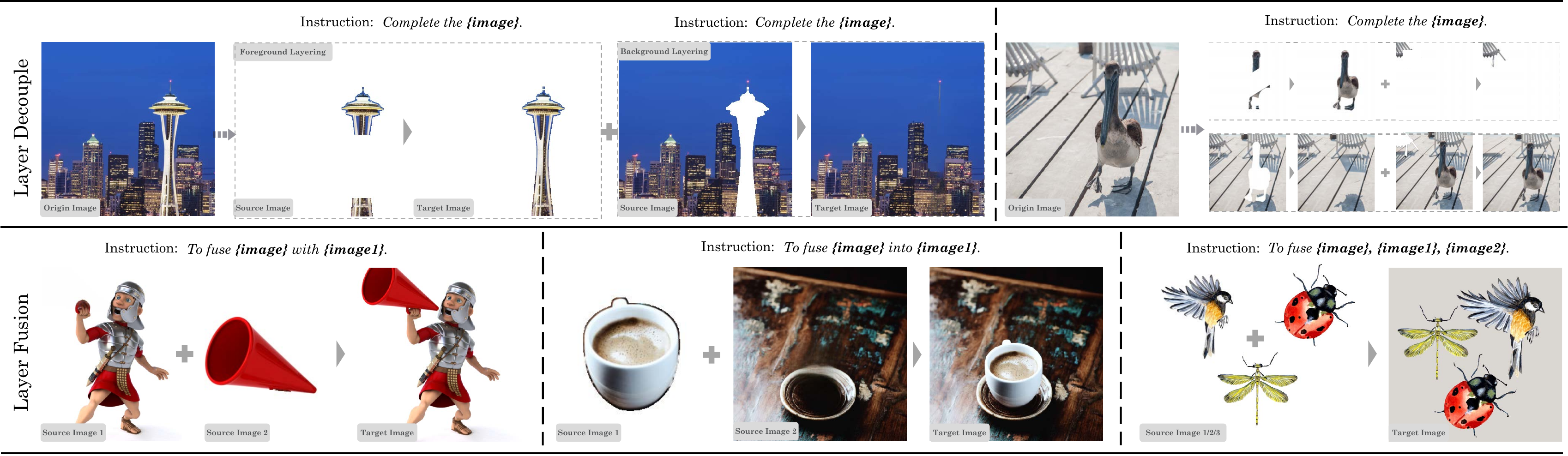}
	\caption{Sample data for training layer decouple and layer fusion in layer editing.}
    \label{fig:sample_layer}
\end{figure*}

Hierarchical layer editing operations on images involve two aspects: \textbf{(i) Layer decouple:} enables the separation of the main subject within a single image, resulting in a complete subject and a reconstructed background. The subject must be restored to its complete form, mitigating any gaps caused by occlusion or other reasons present in the original image. Meanwhile, the background is filled in for the blank areas left after the subject's separation, achieving a fully deconstructed fore/background. \textbf{(ii) Layer fusion:} allows for the incorporation of distinct independent subjects into a target image, facilitating high-quality image integration.
The inference pipeline can be seen in ~\cref{fig:pipe_layer}.

For the data construction, we follow the data workflow from the object editing task, focusing on slightly larger subjects and data containing multiple subjects within a single image. This approach creates compositions that allow for a lossless splitting of a single original image into multiple sub-images, and conversely establishes a correspondence for combining several images into one. 
Specifically, as shown in ~\cref{fig:sample_layer}, in layer decouple stage, we follow the instructions to transition from the original image to either a singular subject or a singular background. During training, the non-subject areas of the subject image and the incomplete portions of the background are filled with white color. Additionally, to simulate the scenario of subjects obscured in the image, we perform random masking on the extracted subject images. The output targets are the complete subject or background. In layer fusion stage, we employ a multi-reference image strategy, taking single or multiple subjects along with the background as inputs to guide the generation of the target image. Similarly, different subjects are supplemented with white color and placed on a randomly sized white canvas, with the training goal being to generate a harmonious and complete composite image.

\subsection{Reference Generation}

\begin{figure*}[!t]
	\scriptsize
	\centering
        \includegraphics[width=\linewidth]{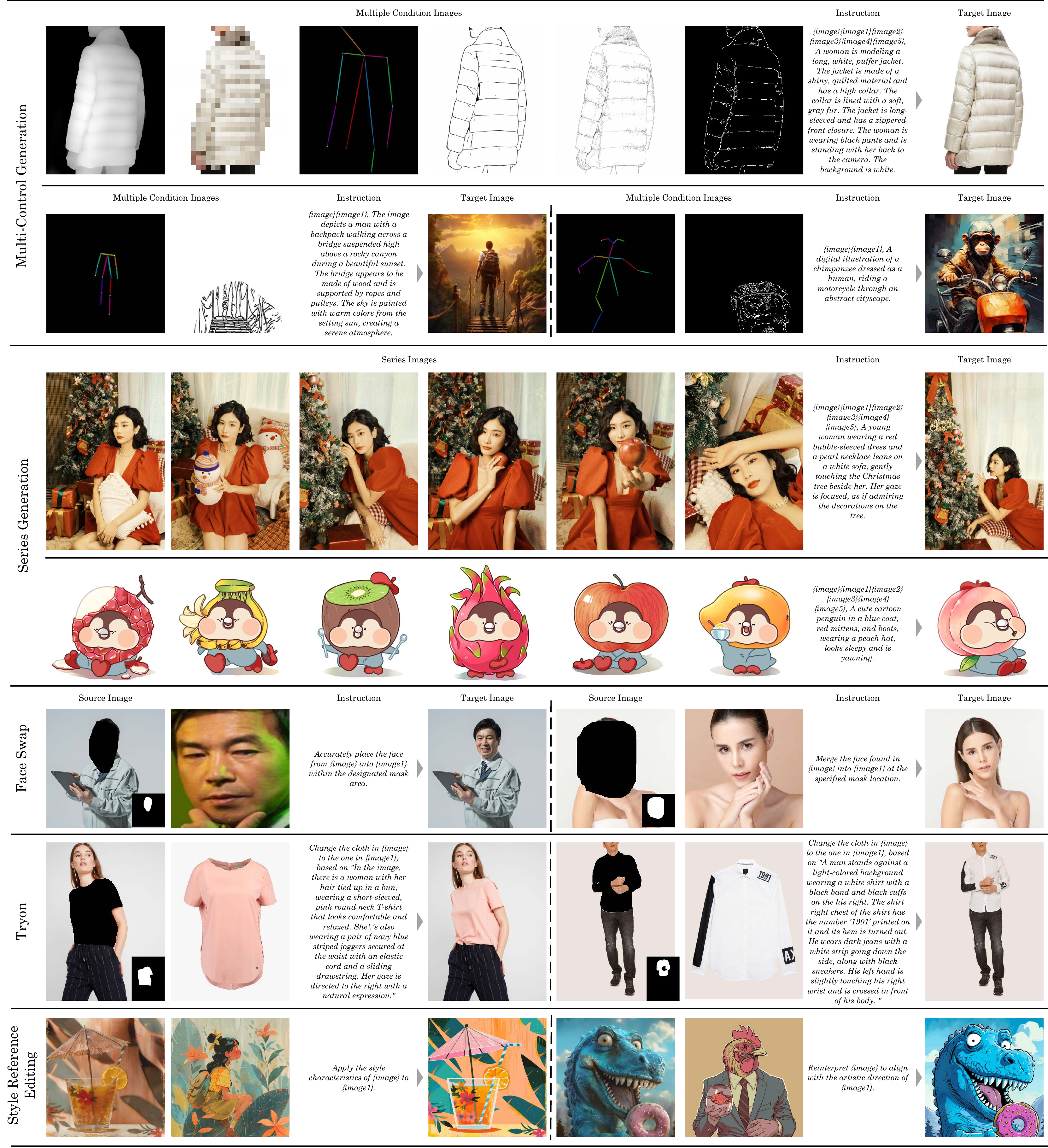}
	\caption{The dataset visualization of multi-reference.}
    \label{fig:sample_multiref}
    \vspace{-10pt}
\end{figure*}

Ordinary image generation and editing tasks require no more than one input image. 
Under certain circumstances, image generation needs multiple image inputs, such as multiple conditions in controllable generation, and a group of character design images for ID preservation. The same is true for editing tasks, one or more additional exemplar images are necessary to specify the expected visual elements in the editing area. For example, a reference image can be interpreted as the target image style appearance, face identity, \textit{etc}. 
Therefore, we prepare training data for multi-reference generation and reference-guided editing. Examples of training data are shown in \cref{fig:sample_multiref}.

\subsubsection{Multi-reference Generation}
\noindent
\textbf{Multi-condition generation.} 
In controllable generation, overlaying different types of conditions is usually necessary to control the different visual aspects of generated images.
Similar to the process in \cref{sec:datasets_control}, canny edge maps, depth maps, color maps, grayscale images, contours, scribbles, doodles, and pose keypoints are included for multi-condition generation. 
To make it possible to composite objects in different conditions, we use object segmentation to assign each area with a different condition.

\noindent
\textbf{Series Generation.} It has been widely studied how to generate images about one consistent visual element, like the portrait of a specific figure, pictures with the same styles, \textit{etc}. 
Usually, tuning a themed tuner (\textit{e.g.}, LoRA)~\citep{lora} with few images is the primary option. However, we are aiming to teach our model to understand and follow the rules lying behind image series. 
We collect image groups through image clustering. During the training phase, we randomly sample one image in the cluster as a target and 3 to 8 images as input images.

\subsubsection{Reference-guided Editing}
Style and face are two typical editing tasks benefiting from additional reference image inputs, providing supplementary visual information of the target images.

\noindent
\textbf{Style reference editing.} To construct the training data, we extend the data of style editing (\cref{sec:datasets_style}) by assigning an additional style reference image for each edit-target image pair. 
Reference images are randomly selected from other styled images within the same style category.

\noindent
\textbf{Face reference editing.} We use image pairs of misaligned facial data (\cref{sec:datasets_face}) for face reference editing. We pick one of the two images as reference image while another as target image. Therefore, the target and reference image are the same person but slightly different. The edit image is derived from target image by erasing the face area to avoid any spoilers.

\subsection{Multi-turn and Long-context Generation}

\begin{figure*}[t]
	\scriptsize
	\centering
        \includegraphics[width=\linewidth]{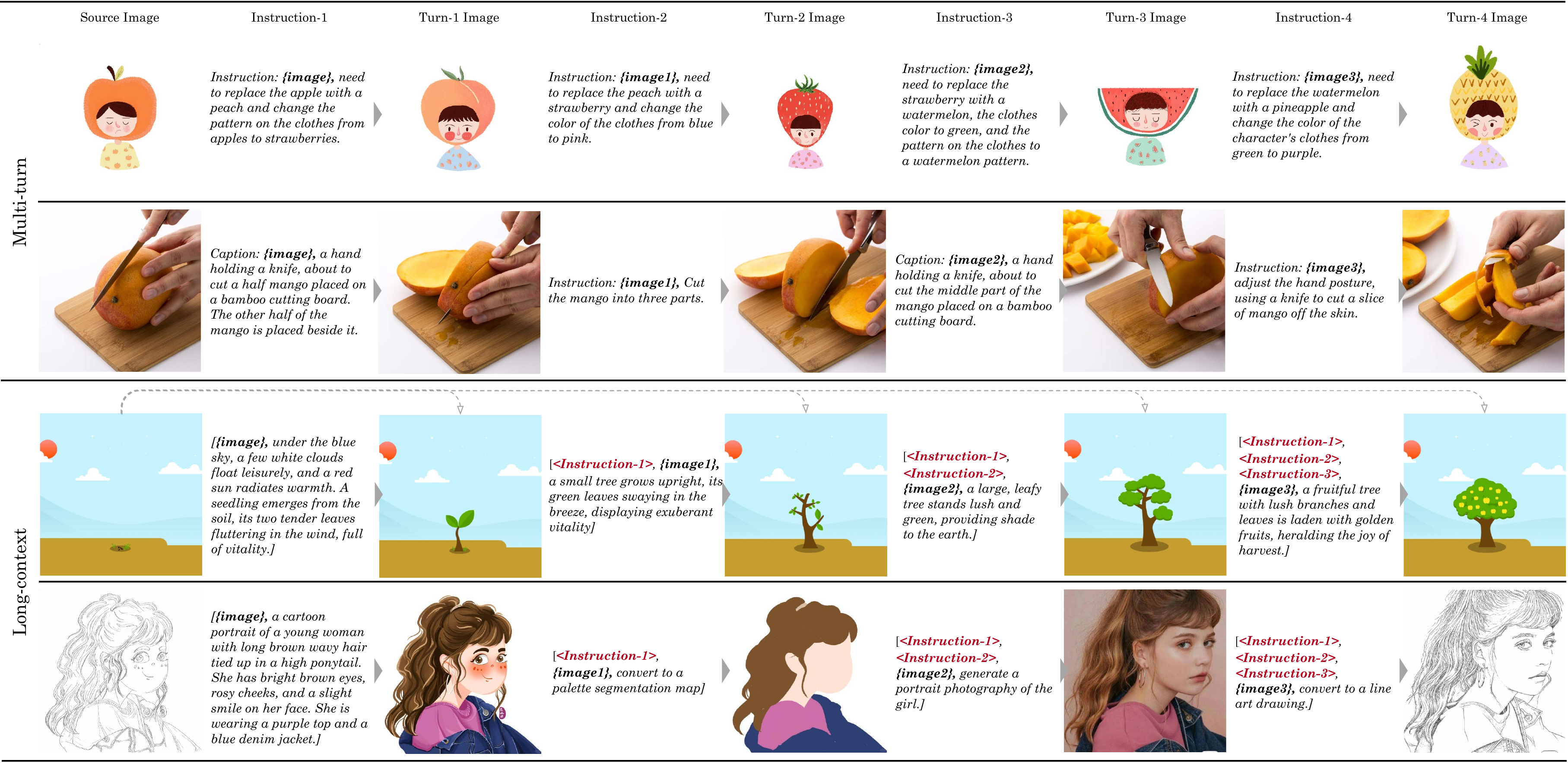}
	\caption{Sample data for training multi-turn and long-context generation task.}
    \label{fig:sample_multi}
    \vspace{-8pt}
\end{figure*}

Multi-turn editing refers to the process of obtaining the final image from an input image through multiple independent instruction-based editing, which poses significant challenges in both the model's precise understanding of instructions and the control over image quality in every round.
Further, the long-context generation process aims to leverage the contextual information provided in each round of interactions to construct a long sequence, thereby generating images that align with the intended directives. 
The generated images reference multiple images and their corresponding instruction information from previous interactions, capturing the user's genuine intent within the interaction framework, and allowing for more precise image editing.

The data construction consists of two parts: 
(i) Homogenous content-based condition unit: this involves employing a pair data collection strategy to obtain various clusters from a large-scale database, as shown in ~\cref{sec:data_pair}, where each cluster contains images paired with their respective captions and instruction generated in pairs. During training, we select one image from any chosen cluster as the starting point and build a multiple rounds data chain using its caption or instruction, predicting the final image as the endpoint of the chain. 
(ii) Task-based condition unit: we treat all the previously mentioned single-image tasks as individual turns within the task and randomly sample them to form a complete multiple precursor unit that guides the final image generation.

\section{Benchmark Details}\label{sec:bench}

\begin{figure*}[t]
	\scriptsize
	\centering
        \includegraphics[width=\linewidth]{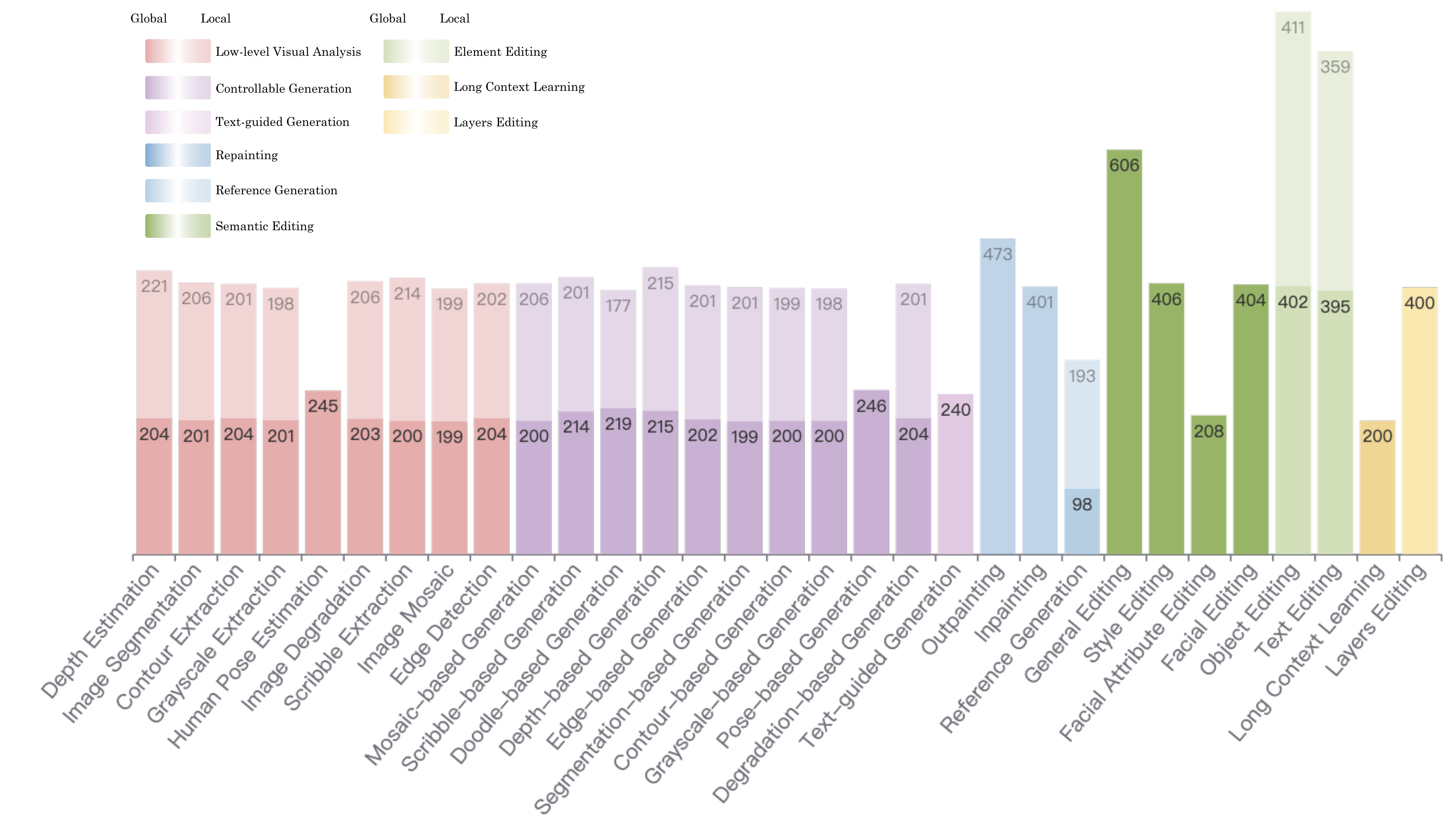}
	\caption{\textbf{The overview of benchmark distribution.} ``Global'' and ``Local'' refer to editing or generating based on the entire image, and editing or generating based on specific local areas of the image, respectively.}
    \label{fig:benchmark_dist}
\end{figure*}

\begin{table}[t]
    \centering
    \caption{The comparison between ACE benchmark and existing benchmarks.}
    \resizebox{0.98\textwidth}{!}{
    \begin{tabular}{ccccccc}
        \toprule
        Benchmark & Real Image? & Generated Image? & Multi-turn? & Regional? & Tasks & Data Scale \\
        \midrule
        MagicBrush & Y & N & Y & Y & - & 1588  \\
        Emu Edit & Y & N & N & N & 8 & 3589 \\
        \midrule
        ACE & Y & Y & Y  & Y & 31 & 12000 
        \\
        \bottomrule
    \end{tabular}
    }
    \label{table:benchmark}
\end{table}

Previous methods have proposed benchmarks to evaluate model performance for image editing, with notable examples including MagicBrush~\citep{magicbrush} and Emu Edit~\citep{emuedit}. 
MagicBrush has 1,588 samples, which includes 1,053 single-turn and 535 multi-turn instances, and primarily comes from MS-COCO~\citep{coco}. 
Emu Edit first defines 8 different categories of potential image editing operations and constructs the instructions by human annotators. 
The main issues with the above methods are insufficient coverage of tasks and generally poor data quality.
\method builds a benchmark comprising 12,000 samples, covering more than 31 tasks while accommodating 5900 real images and 6100 generated images.
In addition, \method benchmark supports both regional editing and multi-turn editing tasks.
The specific statistics are shown in the ~\cref{fig:benchmark_dist}, and the comparison with other benchmarks is presented in the ~\cref{table:benchmark}. 

\section{Implementation Details}\label{sec:impl}

We employ the T5 language model as the text encoder and DiT-XL/2~\citep{dit} as the base network architecture. The model capacity is nearly 0.6B and the parameters are partly initialized by PixArt-$\alpha$~\citep{pixart}. 
The maximum length of the text token sequence is set to 120.
We freeze VAE and T5 modules, utilizing AdamW~\citep{adamw} optimizer to train the DiT module with a weight decay of 5e-4 and a learning rate of 2e-5. All experiments are conducted in A800.

A multi-stage training strategy is employed to progressively enhance the aesthetic quality and increase the generalizability of model. The training details are presented in \cref{tab:train_detail}.
First, we train the instruction-following capability on single-image tasks using 0.7 billion data points, with the number of single image tokens limited to 1024. Next, we expand the tasks to include multiple-image scenarios. After learning the instruction alignment, we utilize high-aesthetic data to enhance the model's aesthetics and extend the max image token number to 4096 for generating higher-resolution image.

\begin{table}[t]
    \centering
    \footnotesize
    \caption{The multi-stage training details for ACE.}
    \label{tab:train_detail}
    \begin{tabular}{ccccccc}
    \toprule
    \multirow{2}{*}{\makecell[c]{Stage}} & \multirow{2}{*}{\makecell[c]{Model\\Capacity}} & \multirow{2}{*}{\makecell[c]{Train Data\\Scale}} & \multirow{2}{*}{\makecell[c]{Visual Sequence\\Length}} & \multirow{2}{*}{\makecell[c]{Max Image\\Number}} & \multirow{2}{*}{\makecell[c]{Training\\Steps}} & \multirow{2}{*}{\makecell[c]{Batch\\Size}}  \\ 
    & & & & & & \\
    \midrule
    Instruction Align & 0.6B    & 0.7 Billion & 1024 & 1 & 900K & 800  \\ 
    Instruction Align & 0.6B    & 0.7 Billion & 1024 & 9 & 100K & 400  \\ 
    Aesthetic Improvement & 0.6B    & 50 million & 1024 & 9 & 500K & 400  \\ 
    Aesthetic Improvement & 0.6B    & 50 million & 4096 & 9 & 100K & 960 \\ 
    \bottomrule
    \end{tabular}
\end{table}

\section{More Experiments}\label{sec:exp_more}

\begin{figure*}[t]
    \scriptsize
    \centering
    \includegraphics[width=\linewidth]{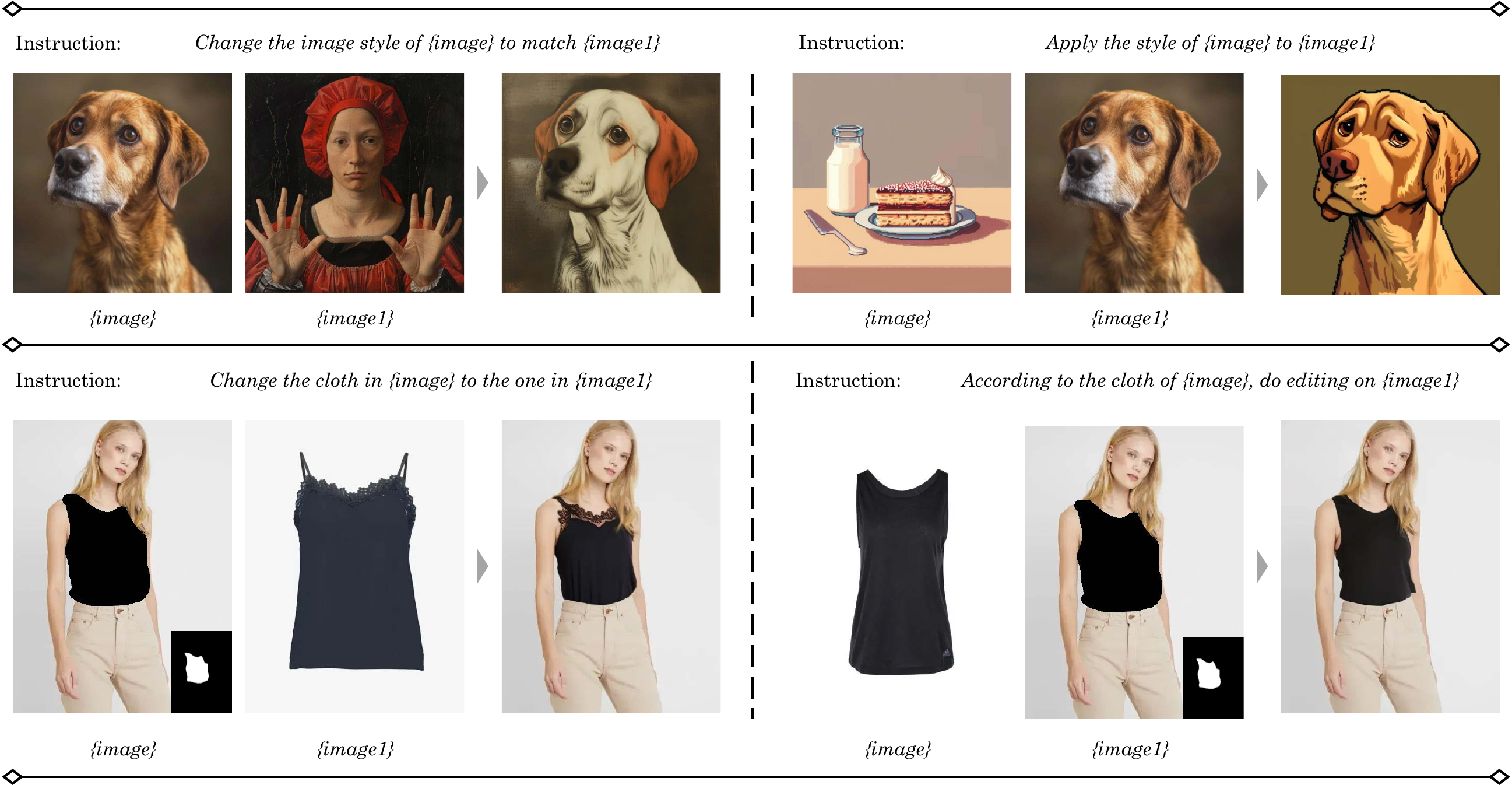}
    \caption{\textbf{The effectiveness of Image Indicator Embeddings.} Model follows the image indicators in instructions to distinguish the source and reference images.}
    \vspace{-10pt}
    \label{fig:image_ind}
\end{figure*}


\noindent\textbf{Design of Image Indicator Embeddings.}
We test the effectiveness of Image Indicator Embeddings by adjusting the order of input images. As we can see in \cref{fig:image_ind}, the model always understands which image is the source image and which is the reference following the image indicators in the instruction. This means textual instructions and images are implicitly associated via the design of Image Indicator Embeddings.

\noindent\textbf{Emu Edit Benchmark.}
We also conduct a comparison on Emu Edit benchmark~\citep{emuedit}. It includes 3,589 examples of 8 tasks: background alteration, comprehensive image changes, style alteration, object removal, object addition, localized modifications, color/texture alterations, and text editing.
This benchmark measures the similarity between output and input images and the provided captions. 
We calculate the L1 distance, CLIP similarity, and DINO similarity between the generated image and input image, together with the CLIP text-image direction similarity measuring agreement between the change in captions and the change in images, and CLIP similarity between the generated image and output caption.
We use the code adapted from the MagicBrush evaluation code and models of CLIP ViT-B/32 and DINO ViT-S/16.
As shown in \cref{tab:assess_emu}, \method achieves comparable performance to its baselines.


\begin{table}[t]
\small
\centering
\setlength{\tabcolsep}{8.5pt}
\caption{Results on Emu Edit benchmark. \method shows comparable performance to its baselines.}
\label{tab:assess_emu}
{
\begin{tabular}{lccccc}
\toprule
\textbf{Method} & CLIPdir$\uparrow$ & CLIPout$\uparrow$ & L1$\downarrow$ & CLIPimg$\uparrow$ & DINO$\uparrow$ \\ 
\midrule
InstructPix2Pix~\citep{ip2p}       & 0.0739 & 0.2681 & 0.1240 & 0.8508 & 0.7647 \\
MagicBrush~\citep{magicbrush}      & 0.0831 & 0.2701 & 0.0995 & 0.8664 & 0.7927 \\
    Emu Edit~\citep{emuedit}       & \bf 0.1073 & \bf 0.2791 & 0.0893 & 0.8743 & 0.8398 \\
UltraEdit~\citep{ultraedit}        & 0.0888 & \underline{0.2783} & \bf 0.0532 &  \underline{0.8814} & \underline{0.8524} \\
CosXL~\citep{cosxl}                 & \underline{0.0901} & 0.2775 & 0.0940 & 0.8686 & 0.8340 \\
\midrule
\rowcolor{tabhighlight} \textbf{\method} (Ours) & 0.0855 & 0.2746 & \underline{0.0761} & \bf 0.8952 & \bf 0.8620 \\ 
\bottomrule
\end{tabular}
}
\end{table}

\noindent\textbf{Facial Editing}.
When evaluating the face identity preservation ability, we designed a Face Similarity (FS) metric to measure the consistency of faces between generated images and original images. We first detect the face region using MTCNN~\citep{mtcnn}, then extract face embeddings with the ArcFace~\citep{arcface} algorithm. The cosine similarity between normalized embeddings is calculated as the face similarity score. The images generated by MagicBrush and InstructPix2Pix exhibit excessive similarity to the original images, thus metrics for these two methods are not computed.
We observed a non-linear growth in the Face Similarity score. To analyze this, we extracted facial features from over 5 million data points in MS1M\_V3~\citep{ms1mv3} and grouped them into clusters based on their face\_id. We then calculated the pairwise similarity within each cluster, resulting in a mean of $mean=0.5258$ and a standard deviation of $std=0.1765$. Due to the large standard deviation, we further analyzed the percentage of samples with scores above 0.3493($mean-std$) that met the instructions to evaluate the Effective Score(ES) of facial ID persistence.


\begin{table}[t]
    \centering\small
    \caption{Quantitative results for Facial Editing tasks on ACE benchmark. $\dagger$ indicates that InstanID requires an additional landmark as a condition, while other methods do not.}
    \begin{tabular}{lcc}
        \toprule
        Method & Face Similarity & Effective Score \\
        \midrule
        InstantID$^\dagger$ ~\cite{instantid} & 84.08 & 0.96 \\
        \midrule
        CosXL ~\cite{cosxl} & 66.49  & 0.37 \\
        UltraEdit ~\cite{ultraedit} & 62.91 & 0.16 \\
        IP-Adapter ~\cite{ipadapter} & \underline{66.51} & 0.31 \\
        FaceChain ~\cite{facechain} & 65.46 & \underline{0.42} \\
        \rowcolor{tabhighlight} \textbf{\method} (Ours) & \textbf{70.07} & \textbf{0.67} \\
        \bottomrule
    \end{tabular}
    \label{tab:face_benchmark}
\end{table}



\begin{table}[!t]
    \centering\small
    \caption{Quantitative results for Local Text Render tasks on \method benchmark.}
    \begin{tabular}{lcc}
    \toprule
    Method & Edit Distance & Sentence Accuracy \\
    \midrule
    UDiffText~\citep{udifftext}      & \underline{0.6827}      & \underline{0.4110}    \\
    AnyText~\citep{anytext}          & 0.6035      & 0.3313    \\
    \rowcolor{tabhighlight} \textbf{\method} (Ours)                       & \textbf{0.8211}      & \textbf{0.5767}    \\
    \bottomrule
    \end{tabular}
    \label{tab:text_metric}
\end{table}

Our model significantly outperforms other methods in the absence of facial landmark information, with improvements of \textbf{3.56\%} and \textbf{25\%} in the FS and ES metrics as shown in \cref{tab:face_benchmark}. 
Although our model (0.6B) demonstrates inferior performance on metrics compared to InstantID (2.6B), it is important to highlight that InstantID utilizes an additional facial landmark as a conditioning factor. Moreover, as indicated by the results of prompt-following and image quality assessments in \cref{tab:user_study}, our model shows a highly competitive performance overall.

\noindent\textbf{Local Text Render}.
To adequately evaluate the performance of text editing, we provide the quantitative analysis of our method with two SOTA text render methods, \textit{i.e.}, UDiffText, and Anytext, on the local text render task of \method benchmark.
Each generated text line is cropped according to the specific position and fed into an OCR model to obtain predicted results.
As described in Anytext, we calculate the Sentence Accuracy and the Normalized Edit Distance for each method. The former metric evaluates the sentence-level accuracy and the latter metric evaluates the char-level precision.
From \cref{tab:text_metric} we can observe that our ACE outperforms the other two methods, achieving performance gains of \textbf{14\%} and \textbf{16\%} in terms of Normalized Edit Distance and Sentence Accuracy, respectively. This demonstrates our superior text rendering capability.

\begin{figure*}[t]
  \scriptsize
  \centering
  \includegraphics[width=0.95\linewidth]{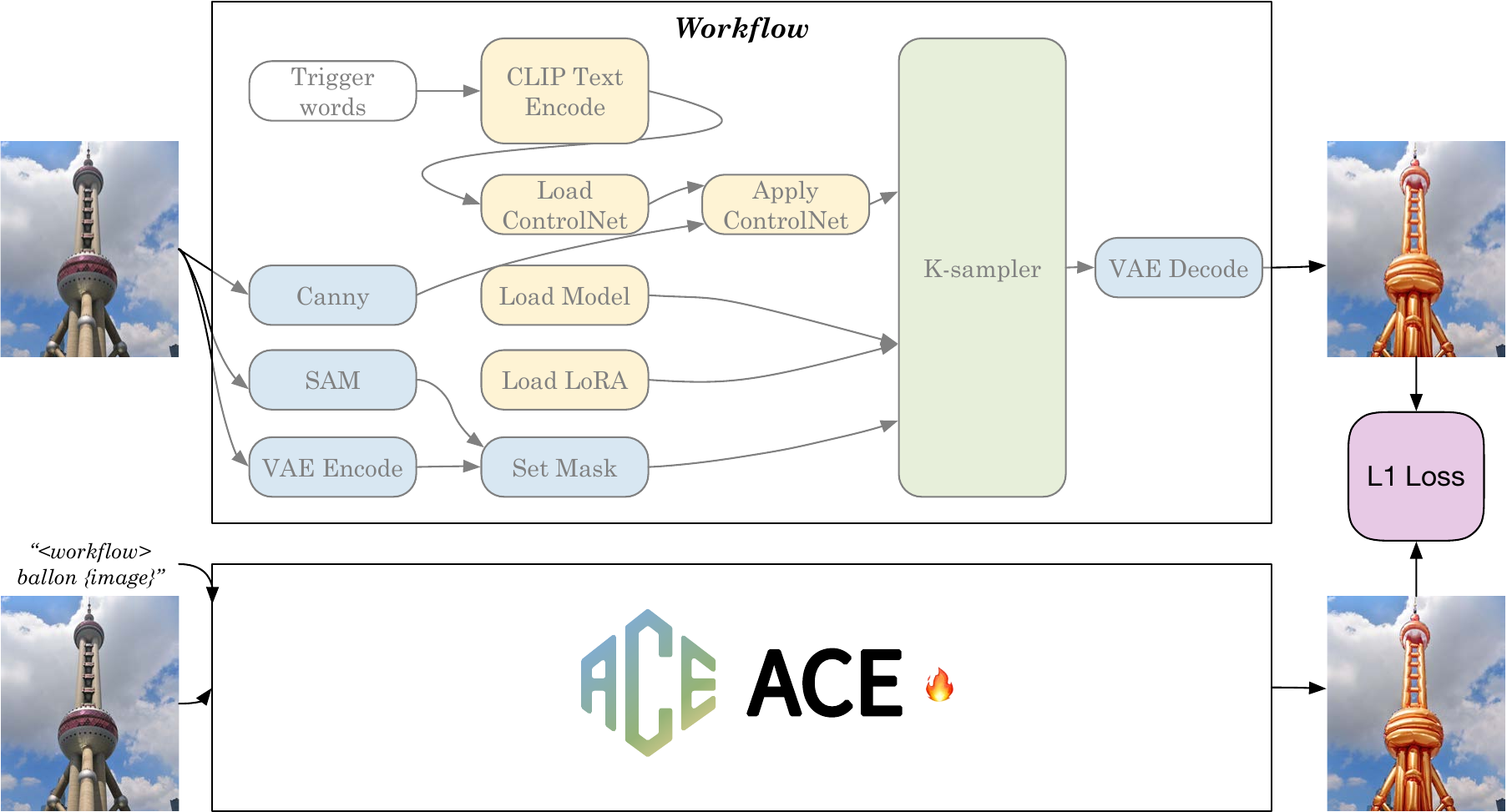}
  \caption{The pipeline of workflow distillation.}
  \label{fig:workflow}
\end{figure*}

\begin{figure*}[t]
  \scriptsize
  \centering
  \includegraphics[width=0.95\linewidth]{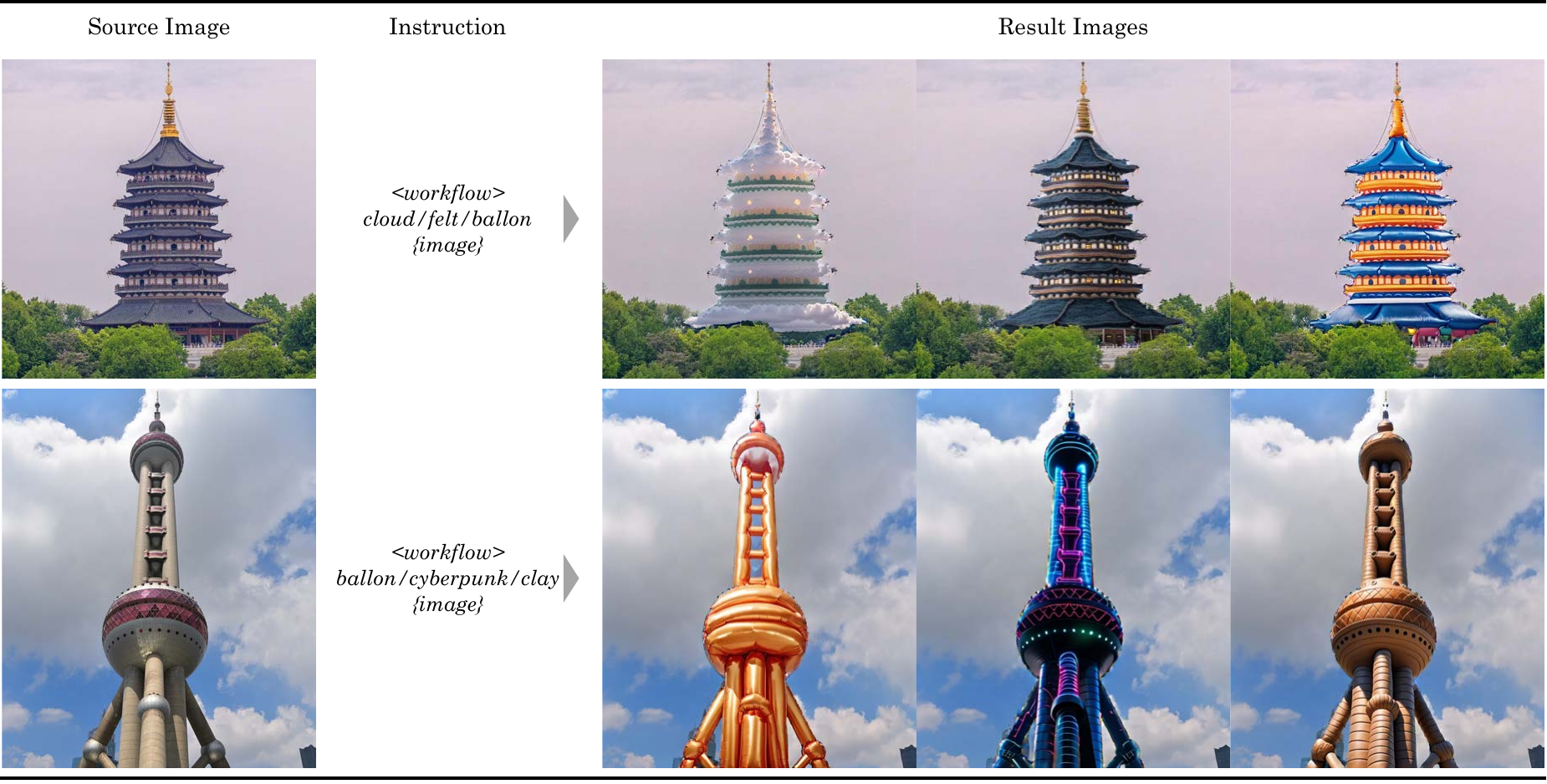}
  \caption{The results visualization of workflow distillation.}
  \label{fig:workflow_case}
\end{figure*}

\section{Application}\label{sec:app}

\begin{figure*}[t]
  \scriptsize
  \centering
  \includegraphics[width=0.95\linewidth]{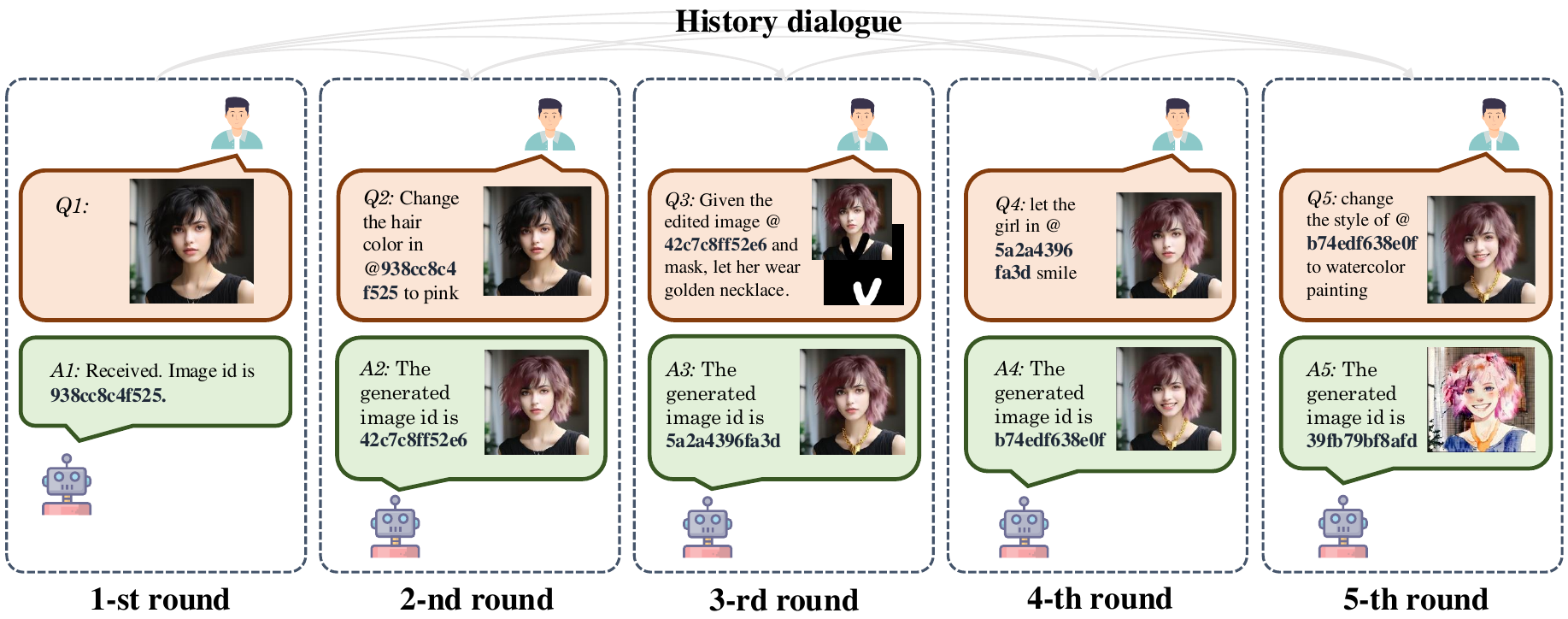}
  \caption{The multi-turn conversation pipeline of our chat bot application.}
  \label{fig:chatbot}
\end{figure*}

\begin{figure*}[t]
  \scriptsize
  \centering
  \includegraphics[width=0.95\linewidth]{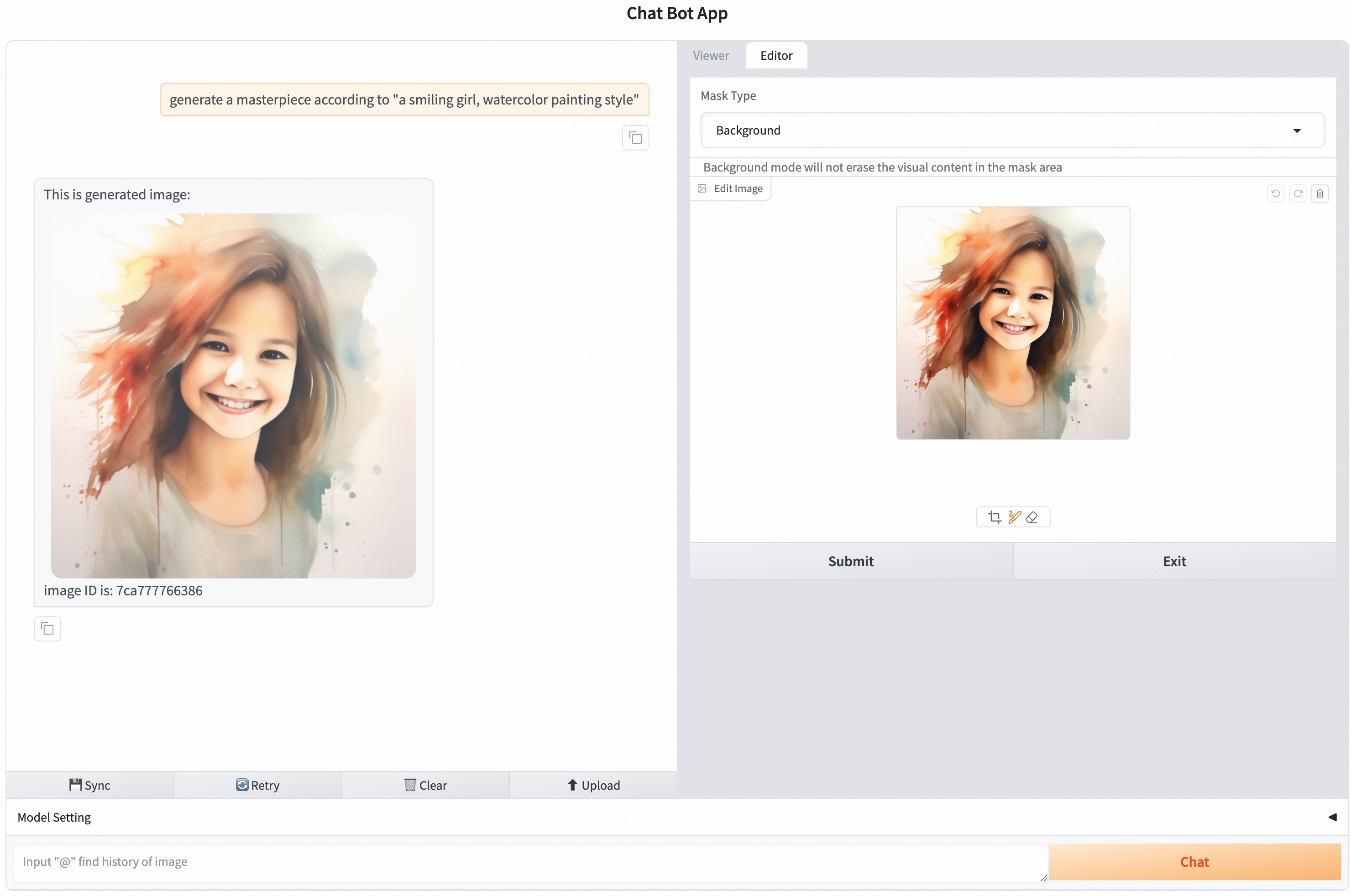}
  \caption{The user interface of the chat bot application built with Gradio.}
  \label{fig:chatbot_ui}
  \vspace{-10pt}
\end{figure*}

\subsection{Workflow Distillation}

There are many excellent workflows assembling LoRAs, ControlNets, and T2I models on open-source platforms, which enable users to achieve certain results. 
To show the capability and compatibility of \method, we collect several outstanding workflows to obtain their result images for distillation. We train \method with the inputs and corresponding outputs of these workflows, as well as a fixed special trigger instruction, as illustrated at \cref{fig:workflow}. 
Our model acquires similar abilities of these workflows, as shown in \cref{fig:workflow_case}, which demonstrates the great potential of \method.

\subsection{Chat Bot}

Leveraging our diffusion model, we build a chat bot application to achieve chat-based image generation and editing. Rather than a cumbersome visual agent pipeline, our chat bot supports all image creation requests with only one model serving as the backend, hence achieving significant efficiency improvement compared with visual agents.
We depict a multi-turn conversation sample in \ref{fig:chatbot} and illustrate the user interface in \cref{fig:chatbot_ui}. We could command the model to create any desired image by chatting with it using natural language. The overall system can be formulated as 
\begin{equation}
    A_{j} = ChatBot(H_{<j}, Q_{j}),
\end{equation}
where $A_{j}$ denotes the $j$-th round output of chat bot, $Q_{j}$ represents the $j$-th round user request, and $H_{<j}=\{(Q_{1}, A_{1}), (Q_{2}, A_{2}), ..., (Q_{j-1}, A_{j-1})\}$ represents the history of dialogue before $j$-th round.
By introducing the history dialogue information into the current conversation, our model excels at understanding complex user requests, therefore achieving better prompt following ability.

\section{More Visualization} \label{sec:vis_more}
\begin{figure*}[ht]
    \scriptsize
    \centering
    \includegraphics[width=\linewidth]{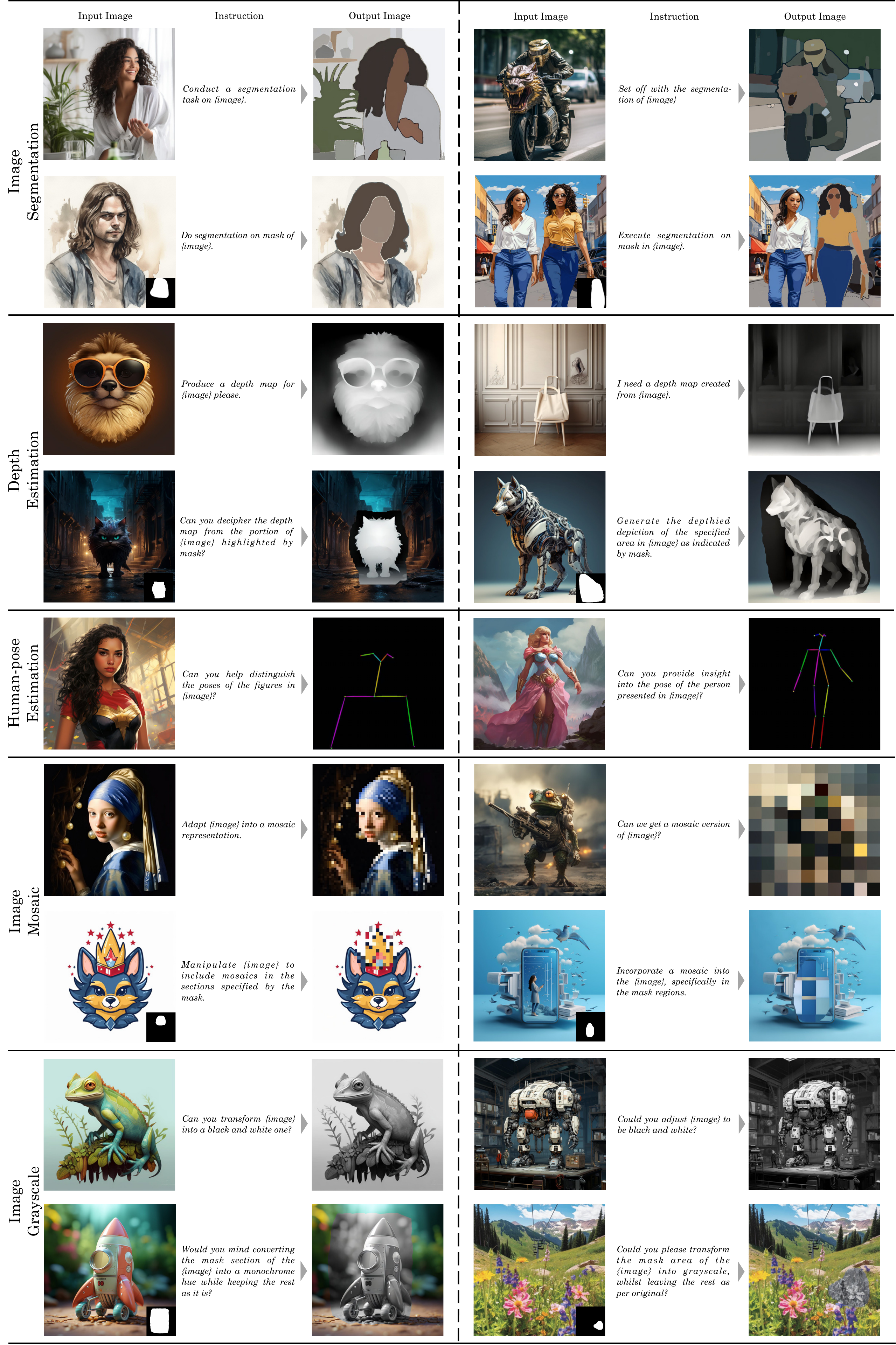}
    \caption{The \method's generated visualization of image segmentation, depth estimation, human-pose estimation, image mosaic, and image grayscale in low-level visual analysis.}
    \label{fig:low_sub1}
\end{figure*}

\begin{figure*}[ht]
    \scriptsize
    \centering
    \includegraphics[width=\linewidth]{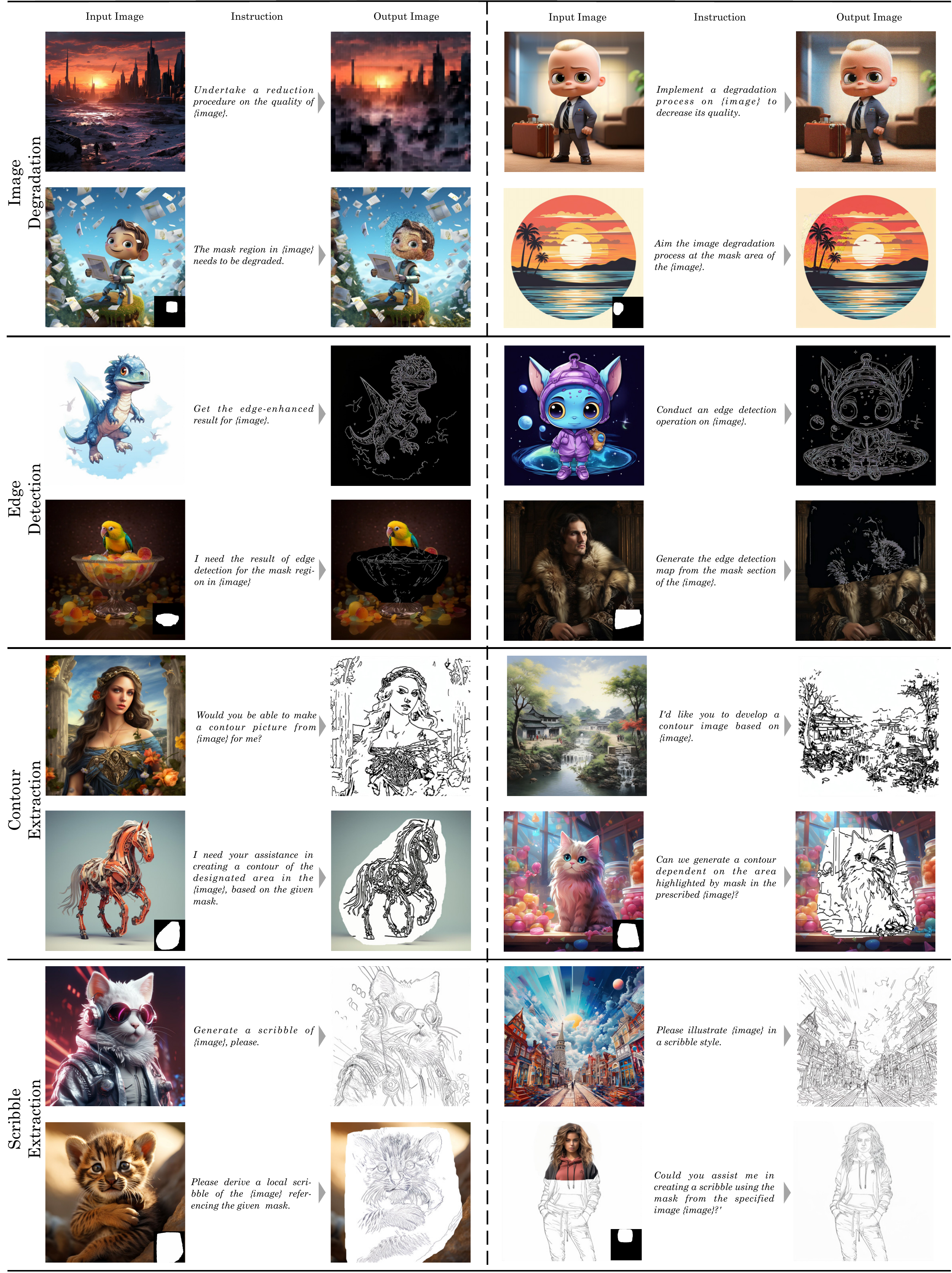}
\caption{The \method's generated visualization of image degradation, edge extraction, contour extraction, and scribble extraction in low-level visual analysis.}
    \label{fig:low_sub2}
    \vspace{-10pt}
\end{figure*}

\begin{figure*}[ht]
    \scriptsize
    \centering
    \includegraphics[width=\linewidth]{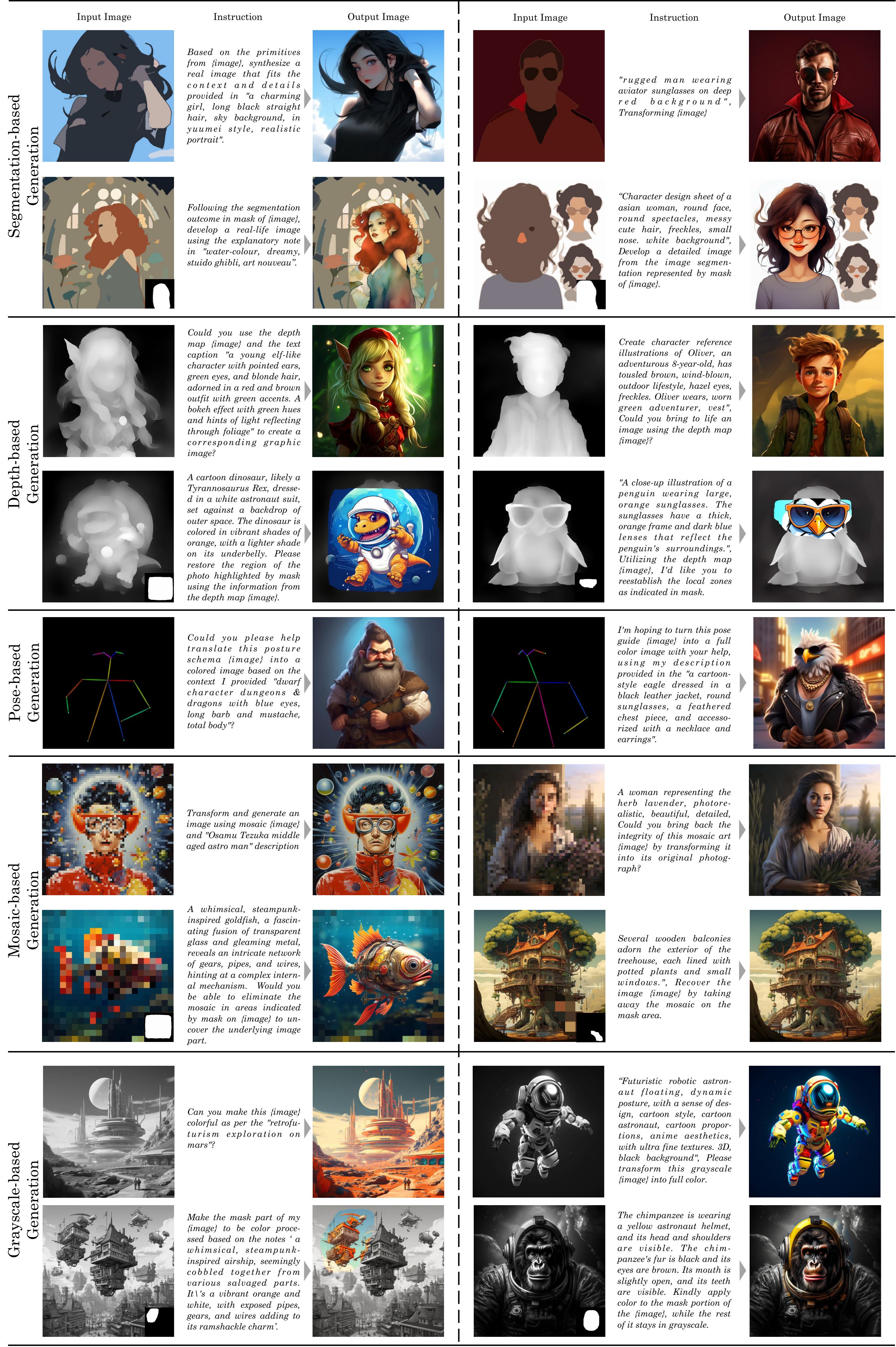}
\caption{The \method's generated visualization of segmentation-based, depth-based, pose-based, mosaic-based, and grayscale-based generation in controllable generation.}
    \label{fig:control_sub1}
\end{figure*}

\begin{figure*}[ht]
    \scriptsize
    \centering
    \includegraphics[width=\linewidth]{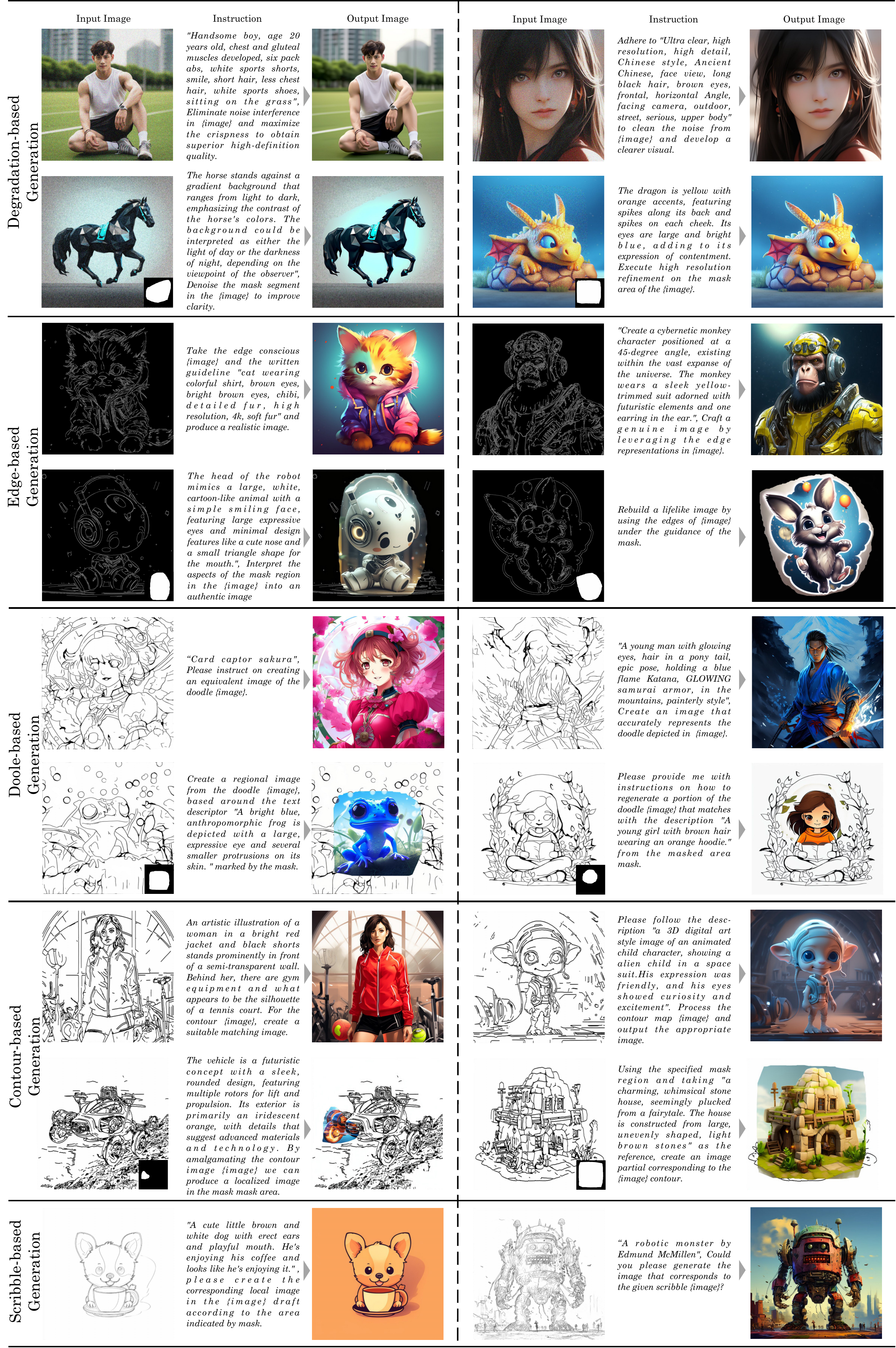}
\caption{The \method's generated visualization of degradation-based, edge-based, doodle-based, contour-based, and scribble-based generation in controllable generation.}
    \label{fig:control_sub2}
\end{figure*}

\begin{figure*}[ht]
    \scriptsize
    \centering
    \includegraphics[width=\linewidth]{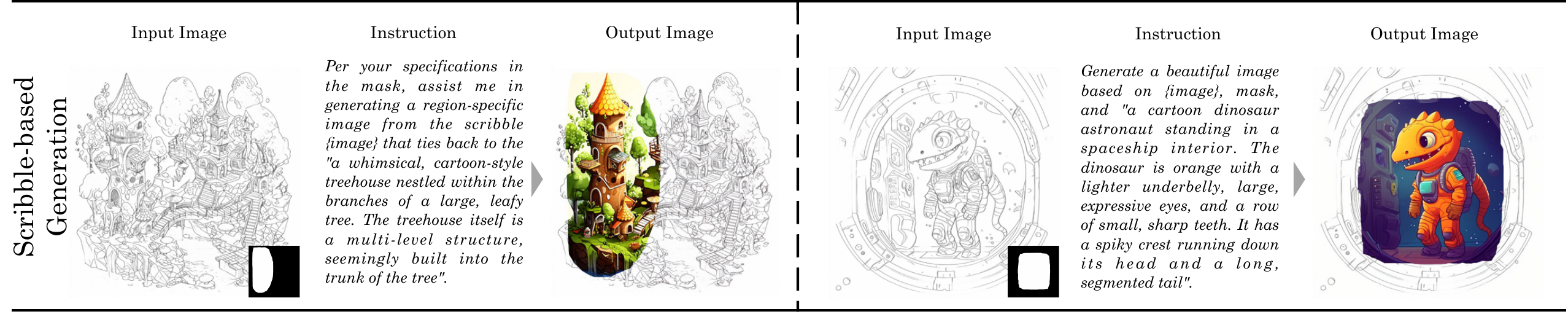}
    \captionsetup{aboveskip=5pt, belowskip=5pt}
\caption{The \method's generated visualization in scribble-based controllable generation.}
    \label{fig:control_sub3}
\end{figure*}

\begin{figure*}[hbtp]
    \scriptsize
    \centering
    \includegraphics[width=\linewidth]{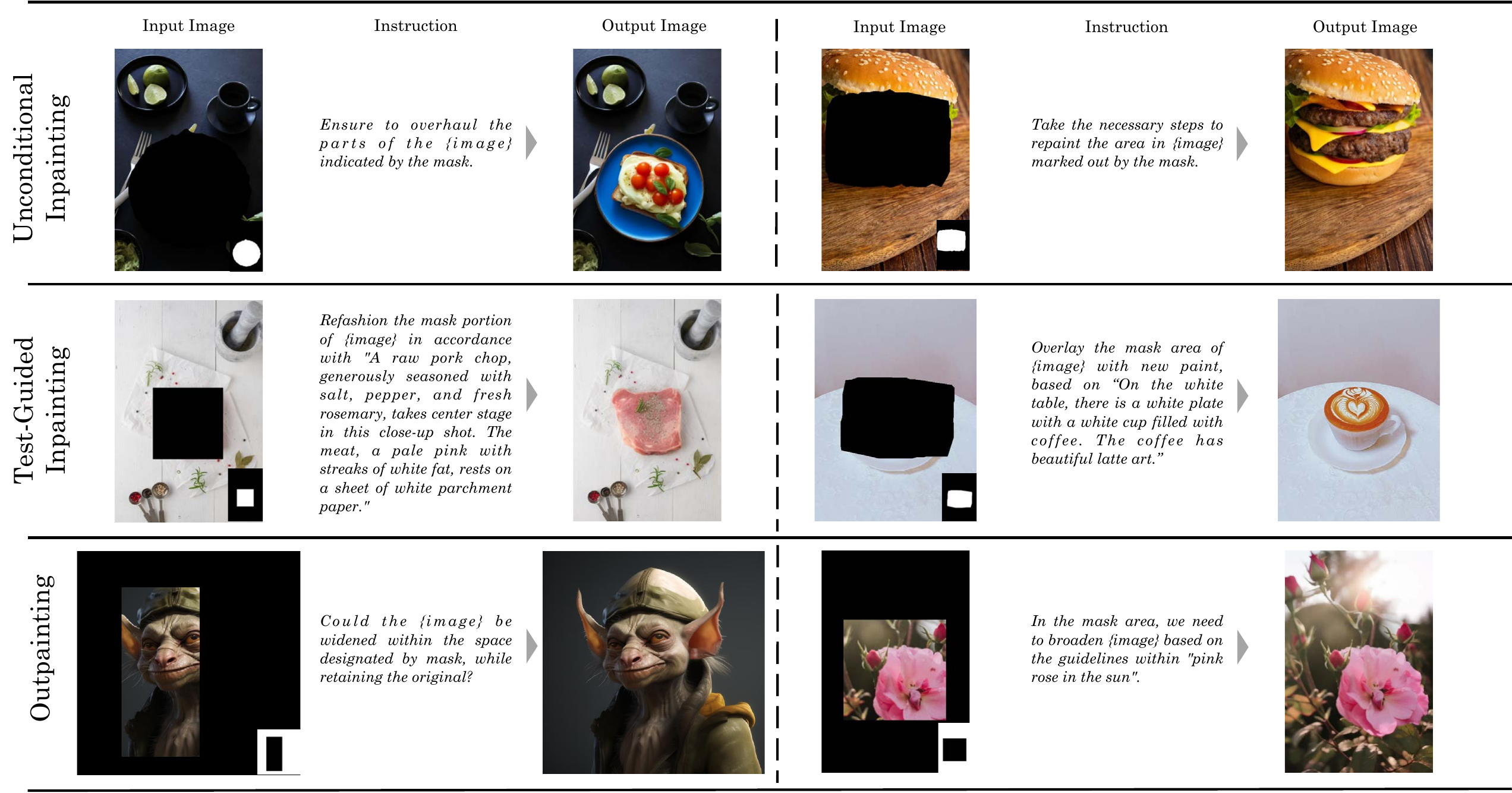}
    \captionsetup{aboveskip=5pt, belowskip=5pt}
    \caption{The \method's generated visualization of repainting.}
    \label{fig:vis_repaint_more}
\end{figure*}

\begin{figure*}[hbtp]
    \scriptsize
    \centering
    \includegraphics[width=\linewidth]{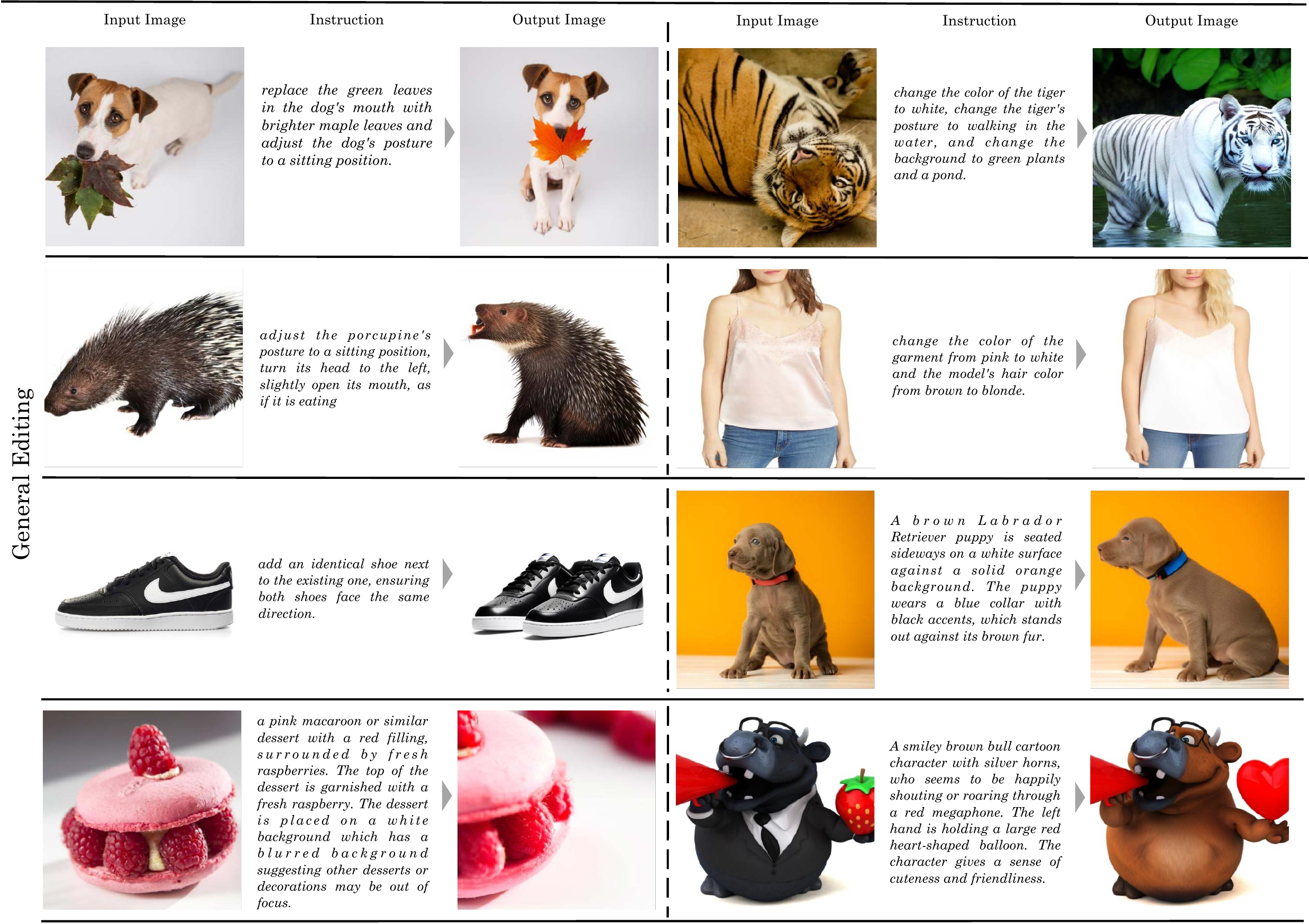}
    \captionsetup{aboveskip=5pt, belowskip=5pt}
    \caption{The \method's generated visualization of general editing in semantic editing.}
    \label{fig:vis_general_more}
\end{figure*}

\begin{figure*}[hbtp]
    \scriptsize
    \centering
    \includegraphics[width=\linewidth]{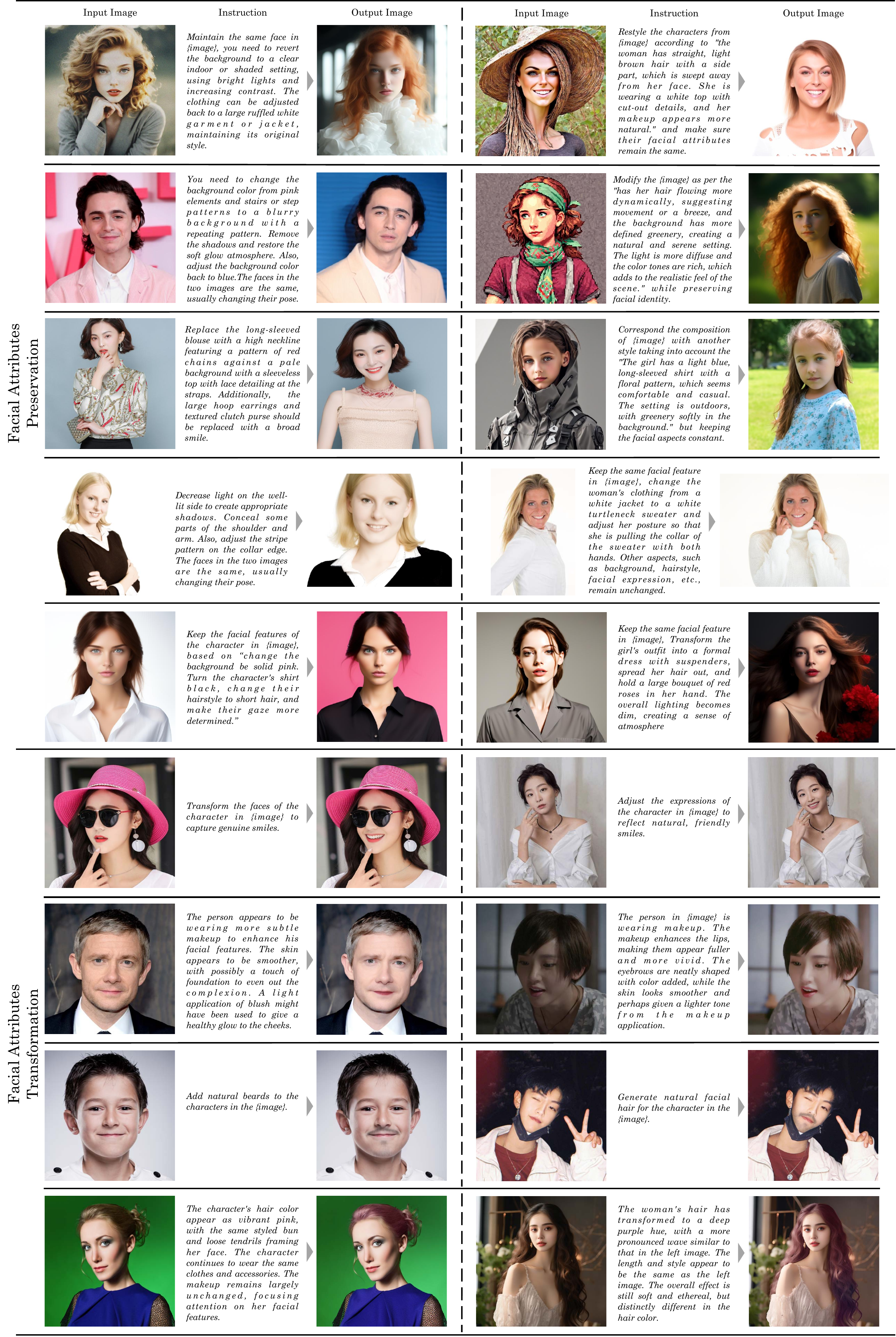}
    \vspace{-3mm}
    \caption{The \method's generated visualization of facial editing in semantic editing.}
    \label{fig:vis_face_more}
\end{figure*}

\begin{figure*}[htbp]
    \scriptsize
    \centering
    \includegraphics[width=\linewidth]{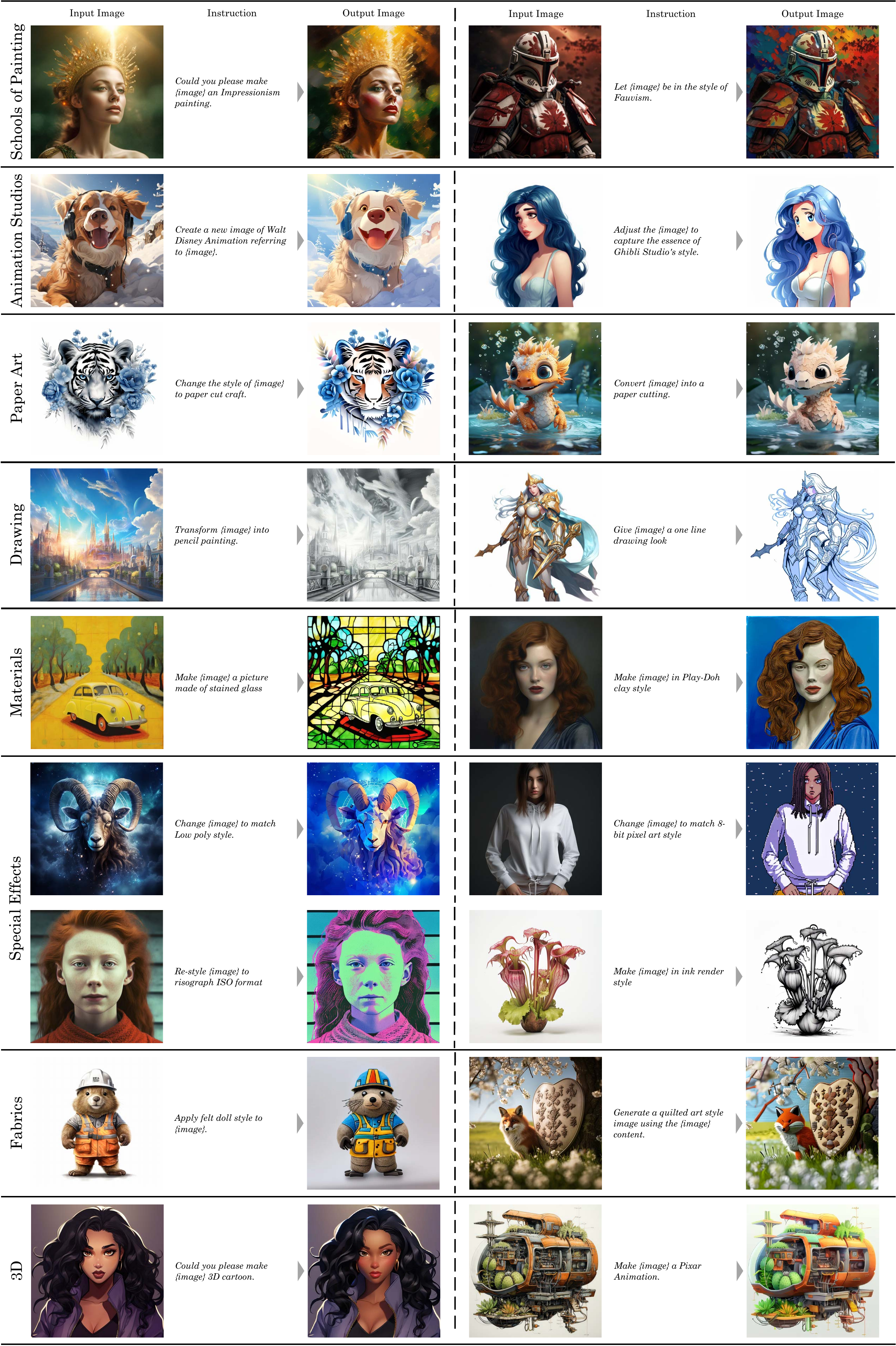}
    \caption{The \method's generated visualization of style editing  in semantic editing.}
    \label{fig:vis_style}
\end{figure*}

\begin{figure*}[hbtp]
    \scriptsize
    \centering
    \includegraphics[width=\linewidth]{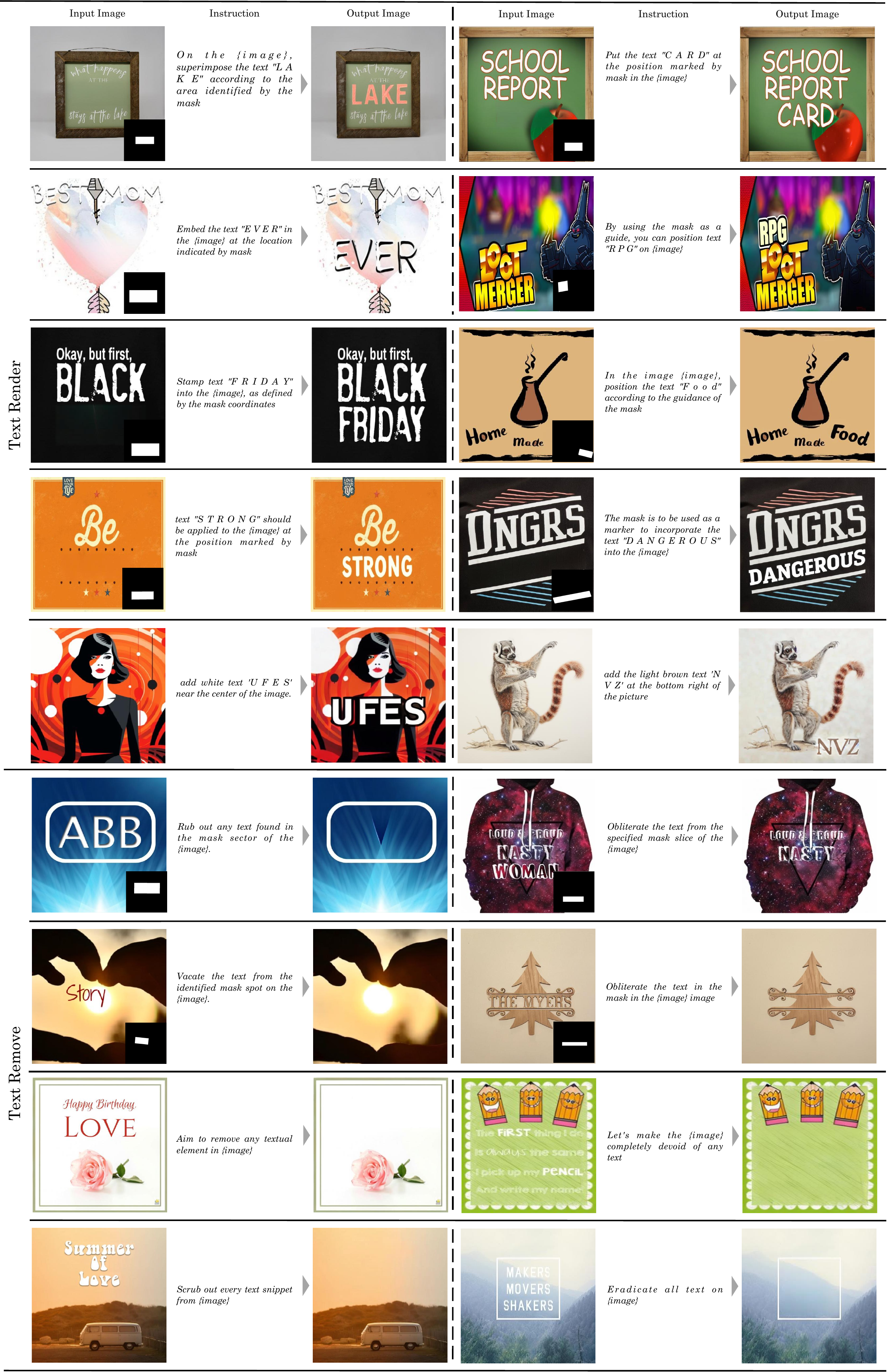}
    \vspace{-3mm}
    \caption{The \method's generated visualization of text editing in element editing.}
    \label{fig:vis_text_more}
\end{figure*}

\begin{figure*}[ht]
    \scriptsize
    \centering
    \includegraphics[width=\linewidth]{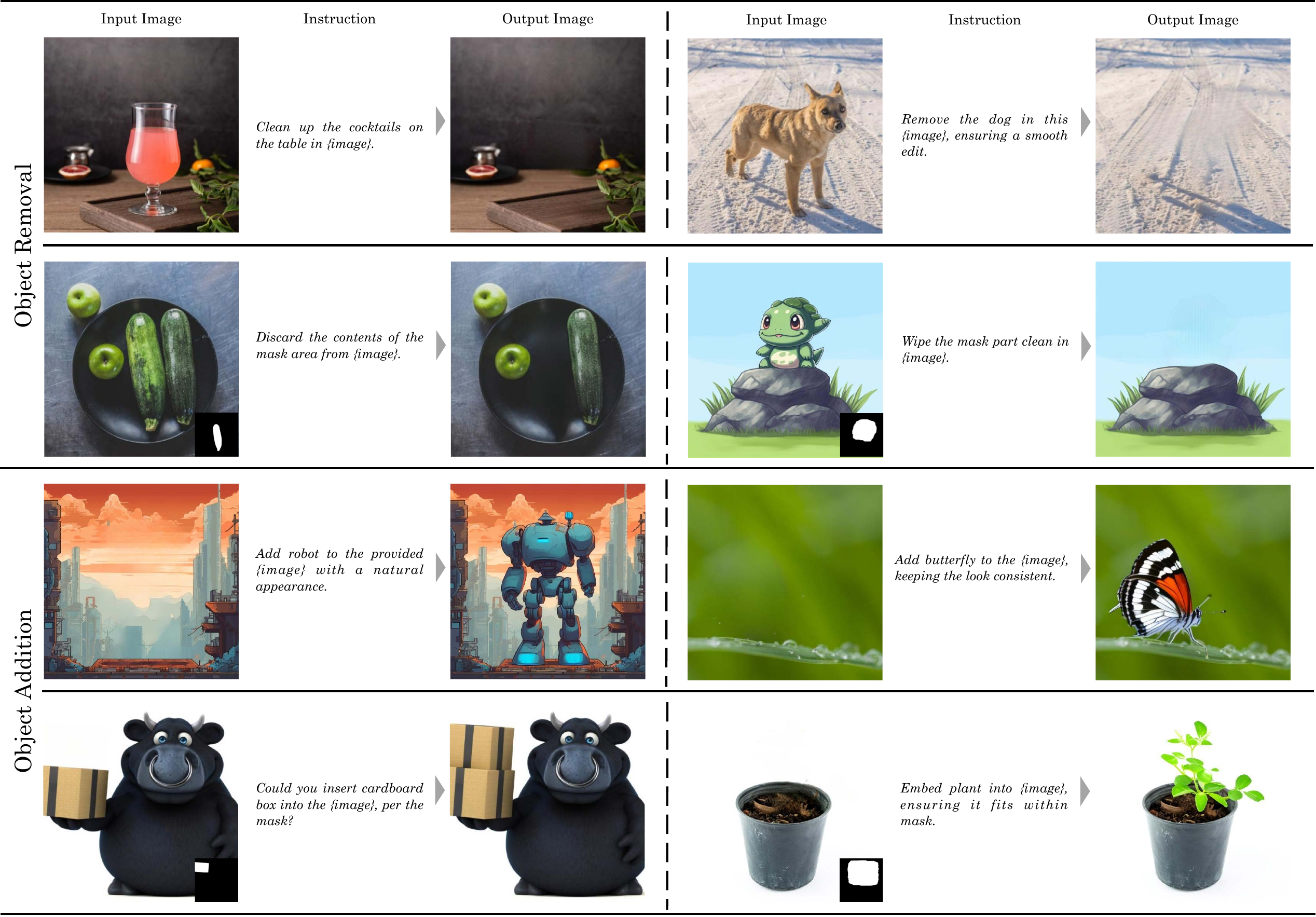}
    \caption{The \method's generated visualization of object editing in element editing.}
    \label{fig:vis_object}
\end{figure*}

\begin{figure*}[ht]
    \scriptsize
    \centering
    \includegraphics[width=\linewidth]{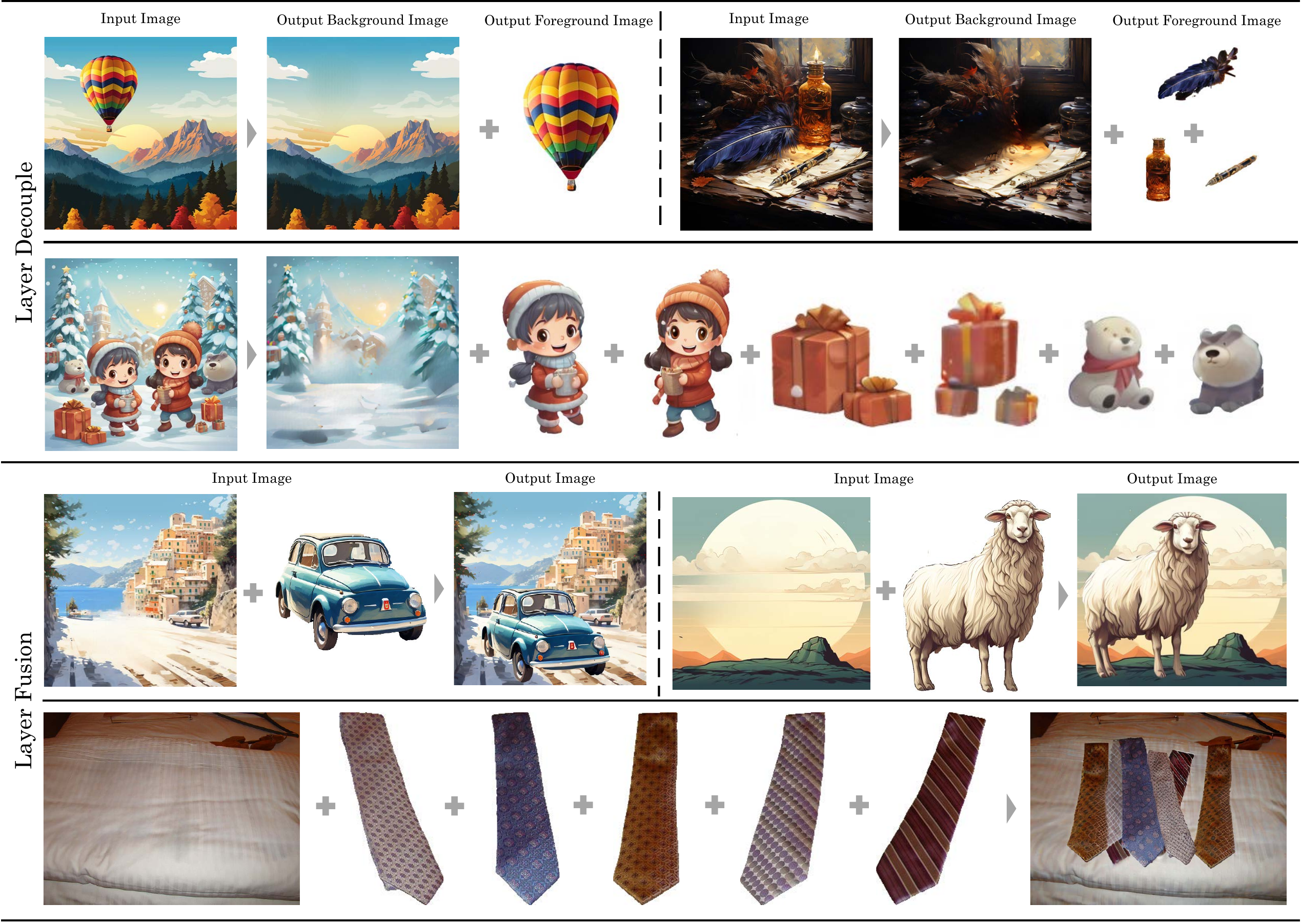}
    \caption{The \method's generated visualization of layer decouple and layer fusion in layer editing.}
    \label{fig:vis_layer}
\end{figure*}

\begin{figure*}[ht]
    \scriptsize
    \centering
    \includegraphics[width=\linewidth]{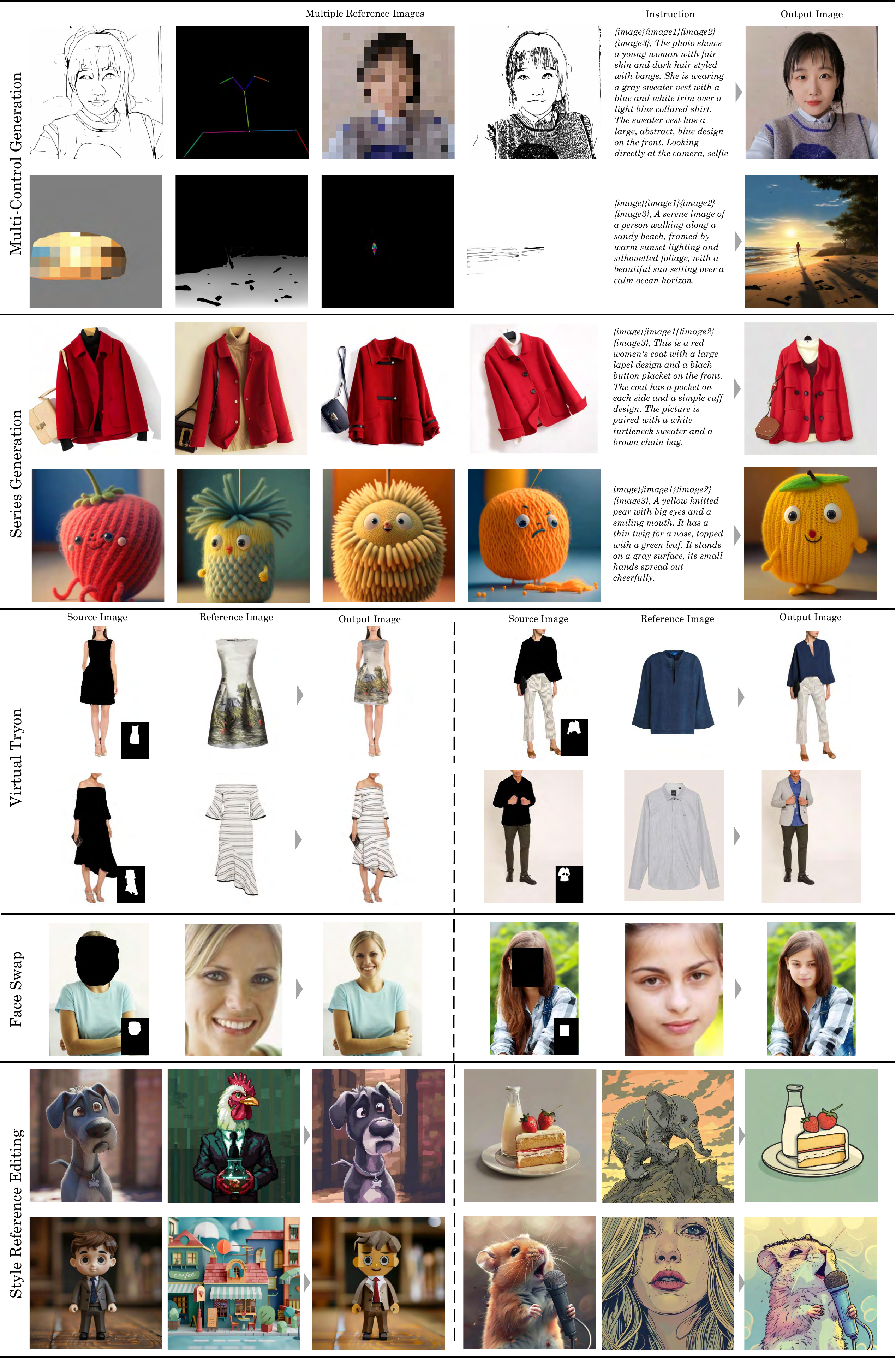}
    \caption{The \method's generated visualization of multi-reference generation and reference-guided editing.}
    \label{fig:vis_ref}
\end{figure*}

\begin{figure*}[ht]
    \scriptsize
    \centering
    \includegraphics[width=\linewidth]{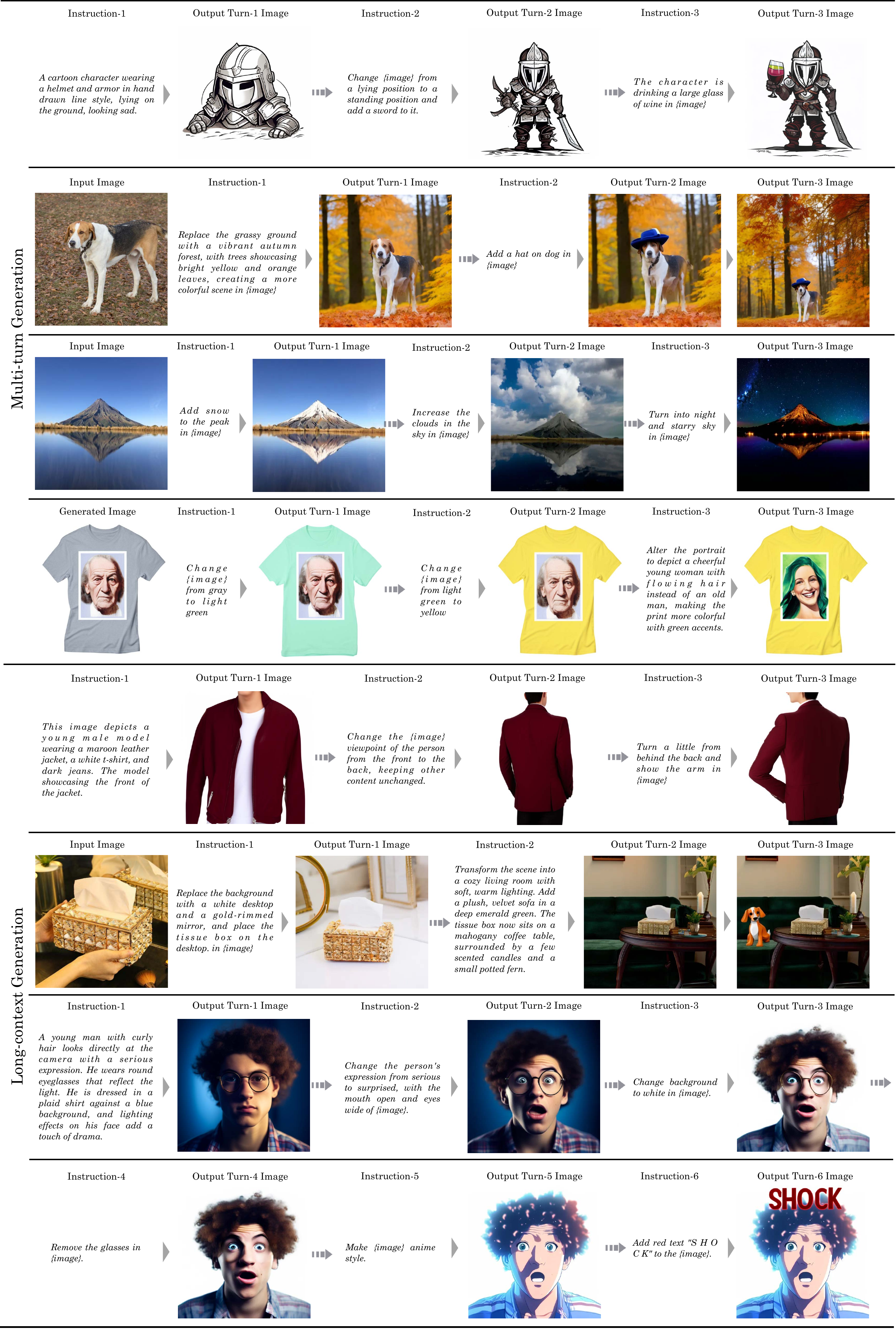}
    \caption{The \method's generated visualization of multi-turn and long-context generation.}
    \label{fig:vis_multi}
\end{figure*}

In ~\cref{fig:low_sub1}, and ~\cref{fig:low_sub2}, we present the visualization results of \method in low-level visual analysis. 
The ~\cref{fig:control_sub1}, ~\cref{fig:control_sub2}, and ~\cref{fig:control_sub3} are the visualization of controllable generation. 
The visualization results of repainting are depicted in ~\cref{fig:vis_repaint_more}. 
Semantic editing tasks such as general editing, facial editing, and style editing are illustrated in ~\cref{fig:vis_general_more}, ~\cref{fig:vis_face_more}, ~\cref{fig:vis_style}. 
The visualization results of elements editing including text editing, and object editing are shown in ~\cref{fig:vis_text_more}, and ~\cref{fig:vis_object}. 
In ~\cref{fig:vis_layer}, we present the visualization results of layer decouple and layer fusion.
The visualization of reference generation can be found at ~\cref{fig:vis_ref} and that of multi-turn and long-context generation are present in ~\cref{fig:vis_multi}.
\method demonstrates proficient instruction following, high-quality generation, and versatility across different tasks.

\section{Discussion}\label{sec:discussion}

\noindent 
\textbf{Societal Impacts.} \label{sec:soci}

From a positive perspective, the intelligent generation and editing of images can provide artists and designers with innovative tools to inspire new concepts, enhance creativity and artistic expression in images, lower the barriers to artistic creation, and reduce the labor-intensive manual processes involved.
Additionally, the method can serve various industries. 
In the field of education and training, they can be used to create supplementary teaching materials, such as illustrations for picture books, enhancing students' learning experiences and improving communication and understanding in lessons. 
In business environments, companies can utilize the method to generate marketing materials and product designs, thereby increasing production efficiency and creative output. 
The positive impacts of these technologies offer new possibilities for creativity, educational quality, and business efficiency, making it worthwhile for us to actively explore and apply them.

New technologies not only bring new opportunities but also come with challenges.
Firstly, issues related to copyright and authorship are prominent, potentially infringing upon the rights of original works and leading to legal disputes. 
Secondly, false information generated by models may exacerbate the spread of rumors, undermining public trust in information. 
Lastly, the inherent biases and stereotypes present in generated content can challenge societal values and moral standards. 
Therefore, while we acknowledge the conveniences and innovations offered by these technologies, it is imperative to carefully consider and effectively manage these negative impacts to ensure the sustainability of technological development and uphold social responsibility.

\noindent
\textbf{Limitations.} \label{sec:limit}

First, our approach offers a unified framework for existing editing tasks. For specific tasks such as text-to-image generation, the aesthetic quality of our generated results lags behind that of state-of-the-art generative models like Midjourney and FLUX. These models have achieved breakthroughs by focusing on a single task of generating images from text prompts. In contrast, our model supports a broader range of input types and handles a wider variety of tasks, such as performing diverse edits under open-ended instructions. Additionally, training on higher-quality data and using a larger-scale model could help bridge this gap.

Second, the model for instruction editing needs to accurately capture the user's actual intent. In our framework, we utilize a fixed encoder-decoder language model to encode text instructions. However, as user instructions become more complex and diverse, the difficulty of interpreting these instructions also increases. Furthermore, the current model is unable to handle multiple intents or tasks from a single instruction simultaneously and requires intent decomposition.

Third, we support the input of multiple images and multiple instructions to construct long contextual information for generation. On one hand, it is inevitable that, due to limited hardware resources, training and inference with multiple images become increasingly challenging as the number of tokens increases. On the other hand, excessively long contextual inputs pose a significant challenge for the model, as the forgetting of historical information during the process may lead to biases in the final generated results.

\noindent
\textbf{Future Work.} \label{sec:future}

We try to illustrate some existing constraints of the model in limitations, which can serve as directions for our future work. 
Firstly, the phenomenon of scaling laws has been demonstrated in the NLP field, indicating that further exploration of scaling laws in complex generation tasks is warranted. We will focus on two main approaches: on one hand, we will work on expanding high-quality data, which includes improving data quality, incorporating more complex tasks, and enhancing the precision of instructional data; on the other hand, we will directly increase the model architecture's scale to enhance its general generative capabilities.
Secondly, we aim to introduce LLMs or MLLMs to accurately capture the intentions of users at the instruction level, leveraging their robust general understanding of language and images. This involves two specific objectives: firstly, to enhance the model's ability to generalize from input text instructions to match single-task or multi-task contexts; and secondly, to improve the understanding of input images that are to be edited, thereby assisting subsequent instruction operations and ensuring the accuracy and diversity of the generated content.
Finally, it is essential to explore the long-sequence modeling of multi-modal data comprising multiple rounds of image and text interactions. We will engage in continuous contemplation regarding how to ensure that the historical context of image-text pairs is truly beneficial, similar to the functionality of chatGPT-like language models.

\newpage


\newpage
\newpage


\end{document}